\documentclass{article}

\usepackage{amsmath,amsfonts,bm}

\usepackage{microtype}
\usepackage{graphicx,float}
\usepackage{subfigure}
\usepackage{booktabs} 
\usepackage{multibib}

\usepackage{hyperref}

\usepackage[accepted]{icml2021}

\icmltitlerunning{Multiscale Invertible Generative Networks for High-Dimensional Bayesian Inference}

\begin{document}

\setlength{\abovedisplayskip}{4pt}
\setlength{\belowdisplayskip}{4pt}

\twocolumn[
\icmltitle{Multiscale Invertible Generative Networks \\ for High-Dimensional Bayesian Inference}



\icmlsetsymbol{equal}{*}

\begin{icmlauthorlist}
\icmlauthor{Shumao Zhang}{Caltech}
\icmlauthor{Pengchuan Zhang}{Microsoft}
\icmlauthor{Thomas Y. Hou}{Caltech}
\end{icmlauthorlist}

\icmlaffiliation{Caltech}{Department of Computational \& Mathematical Sciences, Caltech, Pasadena, California, USA}
\icmlaffiliation{Microsoft}{MSR AI Lab, Redmond, Washington, USA}

\icmlcorrespondingauthor{Shumao Zhang}{shumaoz@caltech.edu}

\icmlkeywords{Machine Learning, ICML}

\vskip 0.3in
]



\printAffiliationsAndNotice{}  

\newcommand{\R}{\mathbb{R}}
\newcommand{\E}{\mathbb{E}}
\newcommand{\Cov}{\mathrm{Cov}}
\newcommand{\rd}{\mathrm{d}}
\newcommand{\id}{\mathrm{id}}
\newcommand{\KL}{D_{\mathrm{KL}}}
\newcommand{\Jef}{D_{\mathrm{J}}}
\newcommand{\eps}{\varepsilon}
\newcommand{\F}{\mathcal{F}}
\newcommand{\N}{\mathcal{N}}
\newcommand{\A}{A}
\newcommand{\B}{B}
\renewcommand{\L}{\mathcal{L}}
\newcommand{\PC}{PC}
\newcommand{\J}{\mathrm{J}}
\newtheorem{theorem}{Theorem}[section]
\newtheorem{lemma}{Lemma}[section]
\newtheorem{assumption}{Assumption}[section]

\newcommand{\RN}[1]{%
  \textup{\uppercase\expandafter{\romannumeral#1}}%
}

\newenvironment{theorem_repeat}[1]
  {\renewcommand{\thetheorem}{\ref{#1}}%
   \addtocounter{theorem}{-1}%
   \begin{theorem}}
  {\end{theorem}}

\newcommand{\rank}{\textrm{rank}}

\begin{abstract}
We propose a Multiscale Invertible Generative Network (MsIGN) and associated training algorithm that leverages multiscale structure to solve high-dimensional Bayesian inference. To address the curse of dimensionality, MsIGN exploits the low-dimensional nature of the posterior, and generates samples from coarse to fine scale (low to high dimension) by iteratively upsampling and refining samples. MsIGN is trained in a multi-stage manner to minimize the Jeffreys divergence, which avoids mode dropping in high-dimensional cases. On two high-dimensional Bayesian inverse problems, we show superior performance of MsIGN over previous approaches in posterior approximation and multiple mode capture. On the natural image synthesis task, MsIGN achieves superior performance in bits-per-dimension over baseline models and yields great interpret-ability of its neurons in intermediate layers.
\end{abstract}

\vspace{-5mm}
\section{Introduction}
\label{sec: intro}

To infer about hidden system states $x\in\R^d$ from observed system data $y\in\R^{s}$, Bayesian inference blends some prior knowledge, given as a distribution $\rho$, with data $y$ into a powerful posterior. Since direct measurement of $x$ can be inaccessible, the data $y$ is generated through $y=\F(x)+\eps$, where $\F$ is a forward map that can be highly nonlinear and complicated, $\eps\in\R^{s}$ is random noise modelled by some distribution. For illustration simplicity, we assume a Gaussian $\N (0, \Gamma)$ for $\eps$. The posterior is characterized as
\begin{align}\label{eq: bayes}
    q(x|y)=\frac{1}{Z}\rho(x)\L(y|x)\,,
\end{align}
where $\L$ is the likelihood given as
\begin{align}\label{eq: likelihood}
    \L(y|x) = \N ( y-\F(x) ; 0, \Gamma)\,,
\end{align}
which is the density of $\eps=y-\F(x)$, and $Z$ is some normalizing constant that is usually intractable in practice. For simplicity reason, in the following context we abbreviate $q(x|y)$ in (\ref{eq: bayes}) as $q(x)$ , because the data $y$ {\it only} plays the role of defining the target distribution $q(x)$ in our framework.

A key and long-standing challenge in Bayesian inference is to approximate, or draw samples from the posterior $q$, especially in high-dimensional (high-$d$) cases. An arbitrary distribution can concentrate its density anywhere in the space, and these concentrations (also called ``modes'') become less connected as $d$ increases. As a result, detecting these modes requires computational cost that grows exponentially with $d$. This intrinsic difficulty of mode collapse is a consequence of the curse of dimensionality, which all existing Bayesian inference methods suffer from, e.g., MCMC-based methods \cite{neal2011mcmc,welling2011bayesian,cui2016dimension}, SVGD-type methods \cite{liu2016stein,chen2018unified,chen2019projected}, and generative modeling \cite{morzfeld2012random,parno2016multiscale,Hou2018}.

In this paper, we exploit the multiscale structure to deal with the high-dimensional Bayesian inference problems. The multiscale structure means that the forward map $\F$ depends mostly on some low-$d$ structure of $x$, referred as coarse scale, instead of high-$d$ ones, referred as fine scale. For example, the terrain shape $x$, given as the discretization of 2-D elevation map on a 2-D lattice grid, is a quantity with dimension equal to the number of grid points. Simulating the 2-D precipitation distribution $y$ using terrain shape $x$ at the scale of kilometer is a reasonable approximation to itself at the scale of meter. The former one is a coarse-scale version of the latter, and has $10^6$-times fewer problem dimension (grid points). Such multiscale structure is very common in high-$d$ problems, especially when $x$ is some spatial or temporal quantity. The coarse-scale approximation to the original fine-scale problem is low-$d$ and computationally attractive, and can help divide-and-conquer the high-$d$ challenge. The multiscale property is discussed in detail in Section \ref{sec: theory}.

We approximate the target $q$ by a parametric family of distribution $p_\theta$, and look for an optimal choice of $\theta$. The working distribution $p_\theta$ is the density of $T(z;\theta)$, where  $z$ is random seed which we assume to be Gaussian noise here, $T$ is a transport map parameterized by $\theta$ that drives $z$ to the sample of $p_\theta$. The optimality of $\theta$ is determined by the match of $p_\theta$ to $q$, measured by the Jeffreys divergence $\Jef(p_\theta\Vert q)$.

\begin{figure}[tbp]
\begin{center}
\centerline{\includegraphics[width=\columnwidth]{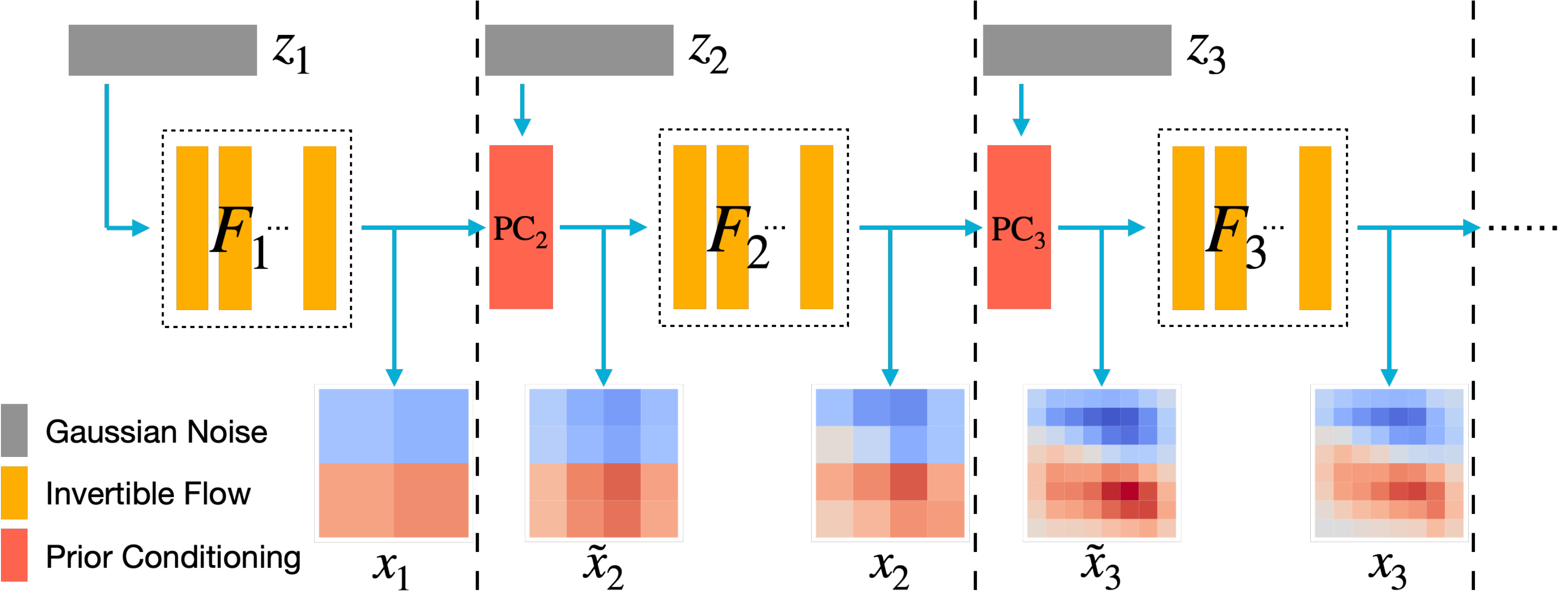}}
\caption{MsIGN generates samples from coarse to fine scale, as depicted by (\ref{eq: MsIGN}). Each scale, separated by dash lines, takes in $x_{l-1}$ from the coarser scale with random seed $z_l$, and outputs a sample $x_l$ of the finer scale. MsIGN iteratively upsamples (by $\PC_l$) and refines (by $F_l$) samples to the target scale. \iffalse Samples are here taken from the example in Section \ref{sec: exp_elp_BIP}.\fi}
\label{fig: network}
\end{center}
\vspace{-6mm}
\end{figure}

We propose a Multiscale Invertible Generative Network (MsIGN) as the map $T$, with a novel training strategy to minimize the Jeffreys divergence. Specifically, $T$ maps $z$ to the sample $x=x_L$ of $p_\theta$ in a coarse-to-fine manner:
\begin{align}\label{eq: MsIGN}
\begin{split}
    &\quad x_1 = F_1(z_1)\,, \\
    \tilde{x}_l = \PC_l(x_{l-1}, z_l)\,, &\quad x_l = F_l(\tilde{x}_l)\,, \quad 2 \le l \le L\,.
\end{split}
\end{align}
Here we split $z$ into $(z_1, z_2, \ldots, z_L)$. At scale $l$, the prior conditioning layer $\PC_{l}$ upsamples the coarse-scale $x_{l-1}\in\R^{d_{l-1}}$ to a finer scale $\tilde{x}_{l}\in\R^{d_{l}}$, which is the ``best guess'' of $x_l$ given its coarse scale version $x_{l-1}$ and the prior $\rho$. The invertible flow $F_l$ then modifies $\tilde{x}_l$ to $x_l\in\R^{d_{l}}$, which again can be considered as a coarse scale version of $x_{l+1}$. The final sample $x=x_L$ is constructed iteratively, as the dimension $d_1<d_2<\ldots<d_L=d$ grows up, see Figure \ref{fig: network}. The overall map $T$ is invertible from $z$ to $x$.

We train MsIGN by minimizing the Jeffreys divergence $\Jef(p_\theta\Vert q)$, defined by \citep{harold1973scientific} as
\begin{align}\label{eq: jeffreys}
\begin{split}
    \Jef ( p_\theta \Vert q ) &= \KL ( p_\theta \Vert q ) + \KL ( q \Vert p_\theta )\\
    &=\E_{p_\theta}\left[\log\left(p_\theta/q\right)\right]+\E_{q}\left[\log\left(q/p_\theta\right)\right]\,.
\end{split}
\end{align}
Jeffreys divergence removes bad local minima of single-sided Kullback-Leibler (KL) divergence to avoid mode missing. We build its unbiased estimation by importance sampling, with the output of the prior conditioning layer as proposal distribution. Furthermore, MsIGN is trained in a multi-stage manner, from coarse to fine scale. At stage $l$, we train $\{F_{l^\prime}:l^\prime\leq l\}$ so that $x_l$ approximates the posterior at its scale, while $\PC_l$ are pre-computed and fixed. Each stage provides a good proposal distribution for the importance sampling at the next stage thanks to the multiscale property.

{\textbf{Contribution\hspace{3mm}} We claim four contributions in this work. First, we propose a Multiscale Invertible Generative Network (MsIGN) with a novel prior conditioning layer that can generate samples from a coarse-to-fine manner. Second, MsIGN allows multi-stage training to minimize the Jeffreys divergence, which helps avoid mode collapse in high-$d$ problems. Third, when applied to two Bayesian inverse problems, MsIGN clearly captures multiple modes in the high-$d$ posterior and approximates the posterior accurately, demonstrating its superior performance over previous methods. Fourth, we also apply MsIGN to image synthesis tasks, where it achieves superior performance in bits-per-dimension among baseline models. MsIGN also yields great interpret-ability of its neurons in intermediate layers.

We introduce the theoretical motivation in Section \ref{sec: theory}, and give detailed introduction of the network structure of our MsIGN in Section \ref{sec: network}, while its training strategy is described in Section \ref{sec: training}. Then we review related work, and provide numerical studies in Section \ref{sec: related} and \ref{sec: exp} respectively.
\vspace{-0.2cm}

\section{Theoretical Motivation}
\label{sec: theory}


Let $\A\in\R^{d_c\times d}$ be a linear operator that downsamples $x$ to its coarse-scale low-$d$ version $x_c=\A x\in\R^{d_c}$ with $d_c<d$. For example, $\A$ can be the average pooling operator with kernel size $2$ and stride $2$ which downsamples $x$ to $1/4$ of its original dimensions.

\textbf{Multiscale structure\hspace{3mm}} In many high-$d$ Bayesian inference problems, the observation $y$ relies more on global, coarse-scale structure than local, fine-scale structure of $x$. This multiscale structure can be described as 
\begin{align}\label{eq: mutliscale}
\F(x)\approx\F(\B\A x)\,,\quad\forall x\in\R^d\,,
\end{align}
where $\A\in\R^{d_c\times d}$ is the downsample operator that compress $x$ to a coarse-scale version $\A x$, and $\B\in\R^{d\times d_c}$ transforms the coarse-scale low-$d$ $\A x$ to a valid system input for $\F$. For example, $\B$ can be the nearest-neighbor upsample operator such that $\B\A x$ has the same size as $x$, but only contains its coarse-scale information. The relation (\ref{eq: mutliscale}) arises frequently when $x$ has some spatial or temporal structure, see an example in Figure \ref{fig: multiscale}.

\begin{figure}[tbp]
\begin{center}
\centerline{\includegraphics[width=\columnwidth]{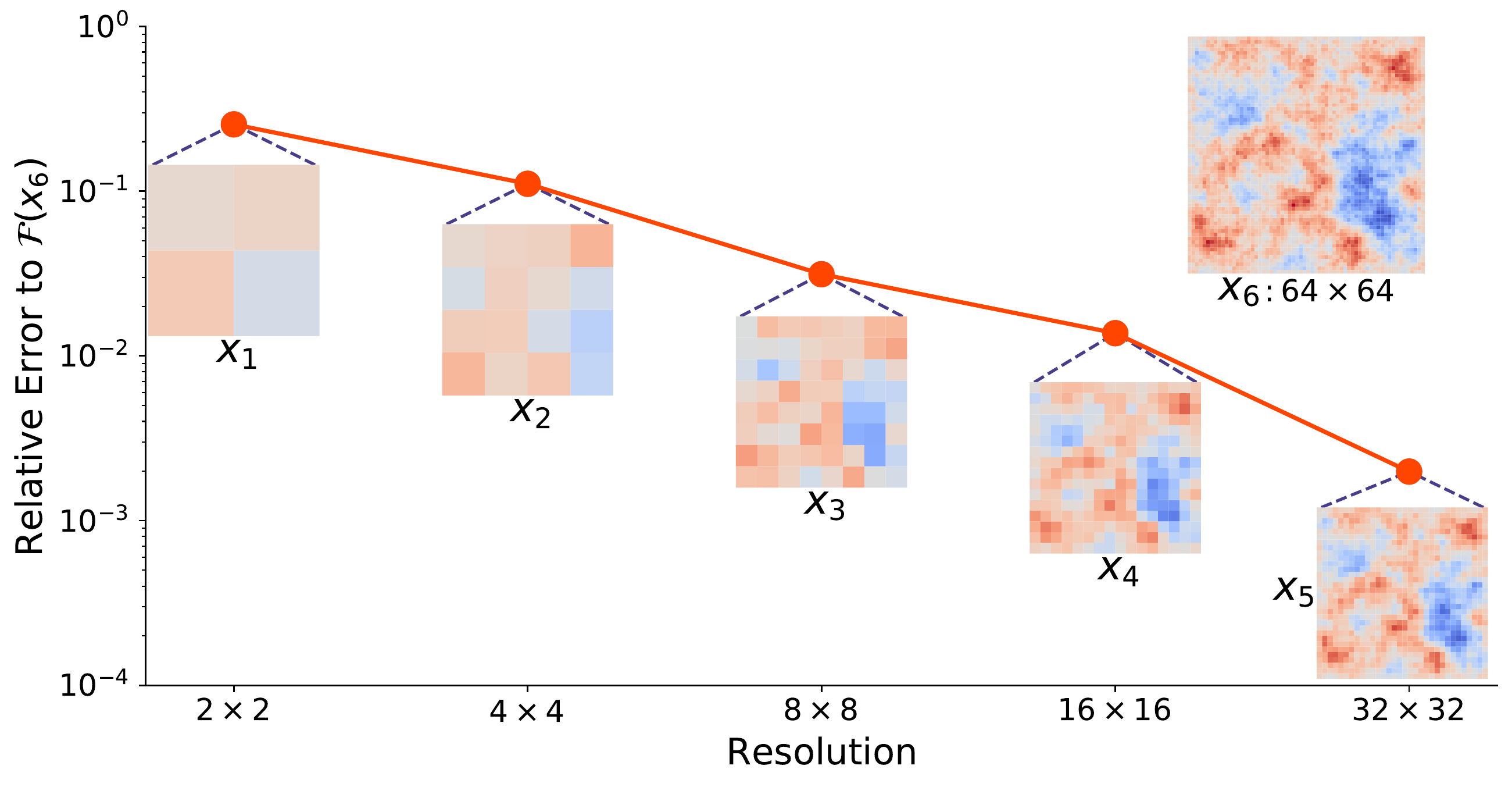}}
\vspace{-0.4cm}
\caption{An example of multiscale problem from Section \ref{sec: exp_elp_BIP}. The coarse-scale $x_1, \ldots, x_5$ are  downsampling of the original $x_6$ from resolution $64\times64$. As the resolution gets refined, the relative error to $\F(x_6)$ significantly drops. Typically, it suggests a good approximation in (\ref{eq: mutliscale}) when, for example, setting $Ax=x_5$.}
\label{fig: multiscale}
\end{center}
\vspace{-10mm}
\end{figure}

\textbf{Scale decoupling\hspace{3mm}} Let $x_c=\A x$ be the coarse-scale variable. Like in (\ref{eq: likelihood}), the coarse-scale likelihood is defined as
\begin{align}\label{eq: coarse_likelihood}
    \L_c(y|x_c) = \N ( y-\F(\B x_c) ; 0, \Gamma)\,,
\end{align}
and we expect $\L_c(y|\A x)\approx\L(y|x)$ due to (\ref{eq: mutliscale}). On the other hand, let $\rho_c$ be the probability density of $x_c=\A x$ when $x\sim\rho$, which is the coarse-scale prior, the conditional probability rule suggests that $\rho(x|x_c)=\rho(x|\A x=x_c)=\rho(x)/\rho_c(x_c)$, which is equivalent to $\rho(x)=\rho_c(x_c)\rho(x|x_c)$.

With the likelihood approximation and the prior decoupling, the posterior $q$ admits the following scale decoupling:
\begin{align}\label{eq: posterior_decoupling}
    q(x)&=\frac{1}{Z}\rho(x)\L(y|x)\approx\frac{1}{Z}\rho(x)\L_c(y|x_c)\nonumber\\
    &=\frac{1}{Z}\rho_c(x_c)\rho(x|x_c)\L_c(y|x_c)\\
    &=\frac{Z_c}{Z}\rho(x|x_c)q_c(x_c)\propto\frac{1}{\tilde{Z}}\rho(x|x_c)q_c(x_c):=\tilde{q}(x)\,,\nonumber
\end{align}
where $q_c(x_c):=\frac{1}{Z_c}\rho_c(x_c)\L_c(y|x_c)$ is the coarse-scale posterior analog to (\ref{eq: bayes}), and $\tilde{q}(x):=\frac{1}{\tilde{Z}}\rho(x|x_c)q_c(x_c)$ is a distribution to approximate $q$, with normalizing constants $Z_c, \tilde{Z}$.

\section{Network Architecture}
\label{sec: network}

The key observation (\ref{eq: posterior_decoupling}) is essentially
\begin{align}\label{eq: key_obs}
    \underbrace{q(x) \approx \tilde{q}(x)}_{(iii)}=\underbrace{q_c(x_c)}_{(i)}\underbrace{\rho(x|x_c)}_{(ii)} \,,
\end{align}
where $\approx$ and $=$ are up to some multiplicative constant. It suggests a three-step way to sample from $q$:
\begin{enumerate}\vspace{-0.2cm}
    \itemsep-0.3em 
    \item[$(i)$] generate a sample $x_c$ from $q_c$;
    \item[$(ii)$] sample $\tilde{x}$ from $\rho(\cdot|x_c)$;
    \item[$(iii)$] further modify $\tilde{x}$ to $x$ to better approximate $q$.
\end{enumerate}\vspace{-0.1cm}
We design a prior conditioning layer $\PC$ to sample $\tilde{x}$ from $\rho(\cdot|x_c)$ for $(ii)$, and an invertible flow $F$ that modifies $\tilde{x}$ for $(iii)$. To obtain $x_c$ from $q_c$ in $(i)$, we apply the above procedure recursively until the dimension of the coarsest scale is small enough so that $q_c$ can be easily sampled by a standard method. As an example of this three-step sampling strategy, in the image synthesis task, a high-resolution image $x$ can be {\it approximated} by $\tilde{x}$, the upsampled image of its low-resolution version $x_c$ superimposed with random noise according to the prior $\rho$, which will be specified in Section \ref{sec: exp_Image} for this task. Needless to say, $\tilde{x}$ needs further modification to achieve good quality in high resolution.

\textbf{Prior conditioning\hspace{3mm}} We feed the coarse-scale sample $x_c$ together with some random seed $z\in\R^{d-d_c}$ to the prior conditioning layer $\PC$ to sample from the conditional distribution $\rho(\cdot|x_c)$: $x=\PC(x_c, z)$. The conditional sample $x$ should satisfy the constrain $\A x=x_c$. We further require the layer $\PC$ to be invertible between $x$ and $(x_c, z)$ to maintain the invertiblity of our overall network. Since $\PC$ depends only on the prior distribution $\rho$ and downsampling operator $\A$, it can be pre-computed {\it regardless} of the likelihood $\L$. In fact, when the prior is a Gaussian, the prior conditional distribution is still a Gaussian and the prior conditioning layer $\PC$ admits a closed form:
\vspace{-0.2cm}

\begin{theorem}\label{thm: gaussian_condition}
Suppose that $\rho$ is a Gaussian with density $\N(x; 0, \Sigma)$ where the covariance $\Sigma$ is positive definite, then with $U^c := \Sigma \A^T(\A\Sigma\A^T)^{-1}\in\R^{d\times d_c}$ and $\Sigma^c := \Sigma-\Sigma \A^T(\A\Sigma\A^T)^{-1}\A\Sigma\in\R^{d\times d}$, we have
$$\rho(x|\A x=x_{c}) = \N(x; U^c x_{c}, \Sigma^c)\,.$$
Furthermore, there exists a matrix $W\in\R^{d\times(d-d_c)}$ such that $\Sigma^c=WW^T$, and the prior conditioning layer $\PC$ can be given as, with $z\in\R^{d-d_c}$ being standard Gaussian
$$x=\PC(x_c, z)=U^cx_c+Wz\,,$$
and $\PC$ is invertible between $x$ and $(x_c, z)$.
\end{theorem}

\vspace{-0.2cm}
We leave the proof in Appendix \ref{ap: proof_gaussian_condition}. When the prior is non-Gaussian, the prior conditioning layer $\PC$ still exists with invertibility guarantee, but it is now {\it nonlinear}. In this case, we can pre-train an invertible network to approximate the conditional sampling process. Once $\PC$ is pre-computed, its parameters are fixed in the training stage.

\textbf{Invertible flow\hspace{3mm}} The invertible flow $F$ is a parametric invertible map that modifies the sample $\tilde{x}$ from the prior conditioning layer to a sample of the target $q$, in other words, it modifies the distribution $\tilde{q}$ in (\ref{eq: key_obs}) to the target $q$. In our experiments in Section \ref{sec: exp}, we utilize the invertible block of Glow \cite{kingma2018glow}, which consists of actnorm, invertible $1\times1$ convolution, and affine coupling layer, and stack several such blocks as the inverse flow $F$ in MsIGN. The approximation (\ref{eq: key_obs}) also suggests that $F$ be initialized as an identity map in training, see Section \ref{sec: training}.

\textbf{Recursive design\hspace{3mm}} To initialize our sampling strategy (\ref{eq: key_obs}) with a sample $x_c$ from the coarse-scale posterior $q_c$, we recursively apply our strategy until the dimension of the coarsest-scale is small enough. Let $L$ be the number of recursion, also called scales in the following context. Let $x_l \in \R^{d_l}$ be the variable at scale $l$ $(1\leq l\leq L)$, whose distribution is the $l$-th scale posterior $q_l$ analog to the $q_c$ in Section \ref{sec: theory} and $q_L=q$. The problem dimension keeps increasing as $l$ goes up: $d_1<d_2<\ldots<d_L=d$. Details of constructions at scale $l$ can be found in Appendix \ref{ap: multiscale_property}.

Our network structure is shown in (\ref{eq: MsIGN}) and Figure \ref{fig: network}, with $z_1\in\R^{d_1}$ and $z_l\in\R^{d_l-d_{l-1}} (2\leq l\leq L$) be the random seed drawn from standard Gaussian at each scale. At scale $l$ $(2\leq l\leq L)$, a prior conditioning layer $\PC_l$ randomly upsamples $x_{l-1}\in\R^{d_{l-1}}$, taken from $q_{l-1}$ approximately, to $\tilde{x}_{l}\in\R^{d_l}$, and an invertible flow $F_l$ modifies $\tilde{x}_{l}$ to $x_l$ to approximate $q_l$. At scale $l=1$, we directly learn an invertible flow $F_1$ that transports $z_1\sim\N(0, I)$ to $x_1\sim q_1$ since the problem dimension is small enough to allow efficient application of standard methods.

Write the overall random seed $z\in\R^d$ as a concatenation of $(z_1, z_2, \ldots, z_L)$, and write $\theta$ as the parameters in MsIGN. The overall network of MsIGN parameterizes a map $T(\cdot;\theta)$ such that samples are generated by $x=T(z;\theta)$. Let $p_z, p_\theta$ be the density of $z, x$ respectively. Our design also allows the invertible mapping $z=T^{-1}(x;\theta)$, so by the change-of-variable formula the density of $p_\theta$ is given by
\begin{align}\label{eq: density}
    p_\theta(x)=p_z(T^{-1}(x;\theta))|\det\J_x T^{-1}(x;\theta)|\,,
\end{align}
where $\J_xT^{-1}$ is the Jacobian of $T^{-1}$ with respect to $x$.

We also remark that when certain bound needs to be enforced on the output, we can append element-wise output activations at the end of MsIGN. For example, image synthesis can use the sigmoid function so that pixel values lie in $[0,1]$. Such activations should be bijective to keep the invertible relation between random seed $z$ to the sample $x$.
\vspace{-0.2cm}

\section{Training Strategy}
\label{sec: training}

We learn network parameter $\theta$ by solving the optimization $\min_\theta \Jef(p_\theta\Vert q)$. Since prior conditioning layers $\PC_l$, for $2\leq l\leq L$, are pre-computed and fixed, trainable parameter $\theta$ only comes from the invertible flows $F_l$, for $1\leq l\leq L$.

\textbf{Jeffreys divergence\hspace{3mm}} While the KL divergence is widely used as the training objective for its easiness to compute, its landscape could admit local minima that don't favor the optimization. In fact, \cite{nielsen2009sided} suggests that $\KL(p_\theta\Vert q)$ is zero-forcing, meaning that it enforces $p_\theta$ be small whenever $q$ is small. As a consequence, mode missing can still be a local minimum, see Appendix \ref{ap: Jef_KL_comparison}. Therefore, we turn to the Jeffreys divergence (\ref{eq: jeffreys}) which significantly penalizes mode missing and can remove such local minima.

Estimating the Jeffreys divergence requires computing an expectation with respect to the target $q$, which is normally prohibited. Since MsIGN constructs a good approximation $\tilde{q}$ to $q$, we do importance sampling with $\tilde{q}$ as the proposal distribution for the Jeffreys divergence and its derivative:
\vspace{-0.2cm}

\begin{theorem}\label{thm: jeffreys}
The Jeffreys divergence and its derivative to $\theta$ admit the following formulation which can be estimated by the Monte Carlo method  {\it without} samples from $q$,
\vspace{-0.1cm}
\begin{align}
    \Jef (p_\theta \Vert q)=&~\E_{p_\theta}\left[ \log\frac{p_\theta}{q} \right]+\E_{\tilde{q}} \left[ \frac{q}{\tilde{q}} \log\frac{q}{p_\theta} \right]\,.\label{eq: compute_jeffreys} \\
    \begin{split}\label{eq: compute_grad_jeffreys}
        \frac{\partial}{\partial\theta}\Jef ( p_\theta \Vert q )=&~\E_{p_\theta}\left[\left(1+\log\frac{p_\theta}{q}\right)\frac{\partial \log p_\theta}{\partial\theta}\right]\\
        &~-\E_{\tilde{q}}\left[\frac{q}{\tilde{q}}\frac{\partial \log p_\theta}{\partial\theta}\right]\,.
    \end{split}
\end{align}
Furthermore, the Monte Carlo estimation doesn't need the normalizing constant $Z$ in (\ref{eq: bayes}) as it can cancel itself.
\end{theorem}
\vspace{-0.2cm}

Detailed derivation is left in Appendix \ref{ap: proof_jeffreys}. With the derivative given above, we optimize the Jeffreys divergence by stochastic gradient descent. We remark that $\partial \log p_\theta/\partial\theta$ is available by the backward propagation of MsIGN, and $\tilde{q}$ comes from coarser scale model in multi-stage training. 

\textbf{Multi-stage training\hspace{3mm}} The multiscale design of MsIGN enables a coarse-to-fine multi-stage training. At stage $l$, we target at capturing the posterior $q_l$ at scale $l$, and only train invertible flows before or at this scale: $F_{l^\prime}$, with $l^\prime\leq l$.

Additionally, at stage $l$, we initialize $F_l$ as the identity map, and $F_{l^\prime}$, with $l^\prime<l$, as the trained model at stage $l-1$. The reason is implied by (\ref{eq: key_obs}), where now $q$, $q_c$ represents $q_l$, $q_{l-1}$ respectively. The stage $l-1$ model provides good approximation to $q_{l-1}$, and together with $\PC_l$ it provides a good approximation $\tilde{q}_l$ to $q_l$. Thus, setting $F_l$ as the identity map will give a good initialization to MsIGN in training. Our experiment shows such multi-stage strategy significantly stabilizes training and improves final performance.

We conclude the training of MsIGN in Algorithm \ref{alg: training}.

\vspace{-3mm}
\begin{algorithm}[H]
    \caption{Train MsIGN by optimizing the Jeffreys divergence in a multi-stage manner}
    \label{alg: training}
\begin{algorithmic}[1]
    \OUTPUT $\theta=(\theta_1, \ldots, \theta_L)$, $\theta_l$ are parameters in $F_l$.
    \STATE Pre-compute and fix all prior conditioning layers $\PC_l$.
    \STATE Learn $\theta_1$ by standard methods such that sample $x_1=F_1(z_1)$ approximates $q_1$.
    \FOR{$l=2$ {\bfseries to} $L$}
    \STATE Initialize $\theta_l$ so that $F_l$ is an identity map.
    \STATE Concatenate last-stage model with $\PC_l$ as $\tilde{q}_l$.
    \STATE With $q_l$ as the target $q$, $\tilde{q}_l$ as the proposal $\tilde{q}$ in (\ref{eq: compute_grad_jeffreys}), compute the gradient using Monte Carlo.
    \STATE Learn $\theta_l$ by stochastic gradient descent.
    \ENDFOR
\end{algorithmic}
\end{algorithm}
\vspace{-5mm}

\section{Related Work}
\label{sec: related}

Invertible generative models \citep{deco1995nonlinear} are powerful exact likelihood models with efficient sampling and inference. They have achieved great success in natural image synthesis, see, e.g., \citep{dinh2016density,kingma2018glow}, and variational inference in providing a tight evidence lower bound, see, e.g, \citep{rezende2015variational}. In this paper, our proposed MsIGN utilizes the invertible block in Glow \citep{kingma2018glow} as building piece for the invertible flow at each scale. The Glow block can be replaced by any other invertible blocks, without any algorithmic changes. Different from Glow, MsIGN adopts a novel multiscale structure such that different scales can be trained separately, making training much more stable. Besides, the multiscale idea enables better explain-ability of its hidden neurons. Invertible generative models like \citep{dinh2016density,kingma2018glow,ardizzone2019guided} adopted a similar multiscale idea, but their multiscale strategy is not in a ``spatial'' sense: the intermediate neurons are not semantically interpret-able as shown in Figure \ref{fig: img_hierarchy}. The multiscale idea is also used in generative adversarial networks (GANs), as in \citep{denton2015deep,odena2017conditional,  karras2017progressive, xu2018attngan}. But lack of invertibility in these models makes it difficult for them to apply to Bayesian inference problems.

Different from the image synthesis task where large amount of samples from target distribution are available, in Bayesian inference problems only an unnormalized density is available and i.i.d. samples from the posterior are the target. This  main goal of this paper is to train MsIGN to approximate certain high-$d$ Bayesian posteriors. Various kinds of parametric distributions have been proposed to approximate posteriors before, such as polynomials \citep{el2012bayesian,parno2016multiscale,matthies2016inverse,spantini2018inference}, non-invertible generative networks \citep{feng2017learning,Hou2018}, invertible networks \citep{rezende2015variational,ardizzone2018analyzing,kruse2019benchmarking} and certain implicit maps \citep{chorin2009implicit,morzfeld2012random}. Generative modeling approach has the advantage that i.i.d. samples can be efficiently obtained by evaluating the model in the inference stage. However, due to the tricky non-convex optimization problem, this approach for both invertible \citep{chorin2009implicit,kruse2019benchmarking} and non-invertible \citep{Hou2018} generative models becomes increasingly challenging as the dimension grows. To overcome this difficulty, we propose to minimize the Jeffreys divergence, which has fewer local minima and better landscape compared with the commonly-used KL divergence, and to train MsIGN in a coarse-to-fine manner.

Other than the generative modeling, various Markov Chain Monte Carlo (MCMC) methods have been the most popular in Bayesian inference, see, e.g., \citep{beskos2008mcmc,neal2011mcmc,welling2011bayesian,chen2014stochastic,chen2015convergence,cui2016dimension}. Particle-optimization-based sampling is a recently developed effective sampling technique with Stein variational gradient descent (SVGD) \citep{liu2016stein}) and many related works, e.g., \citep{liu2017stein,liu2018riemannian,chen2018unified,chen2019projected,chen2020projected}. The intrinsic difficulty of Bayesian inference displays itself as highly correlated samples, leading to undesired low sample efficiency, especially in high-$d$ cases. The multiscale structure and multi-stage strategy proposed in this paper can also benefit these particle-based methods, as we can observe that they benefit the amortized-SVGD \citep{feng2017learning,Hou2018} in Section \ref{sec: exp_BIP_ablation}. We leave more discussion about the related work in Appendix \ref{ap: related_work}.

\vspace{-2mm}

\section{Experiment}
\label{sec: exp}

We study two high-$d$ Bayesian inverse problems (BIPs) in Section \ref{sec: exp_BIP} as test beds for distribution approximation and multi-mode capture. We also apply MsIGN to the image synthesis task to benchmark with flow-based generative models and demonstrate its interpret-ability in Section \ref{sec: exp_Image}.

\begin{figure*}[tbp]
\centering
\includegraphics[width=0.36\textwidth]{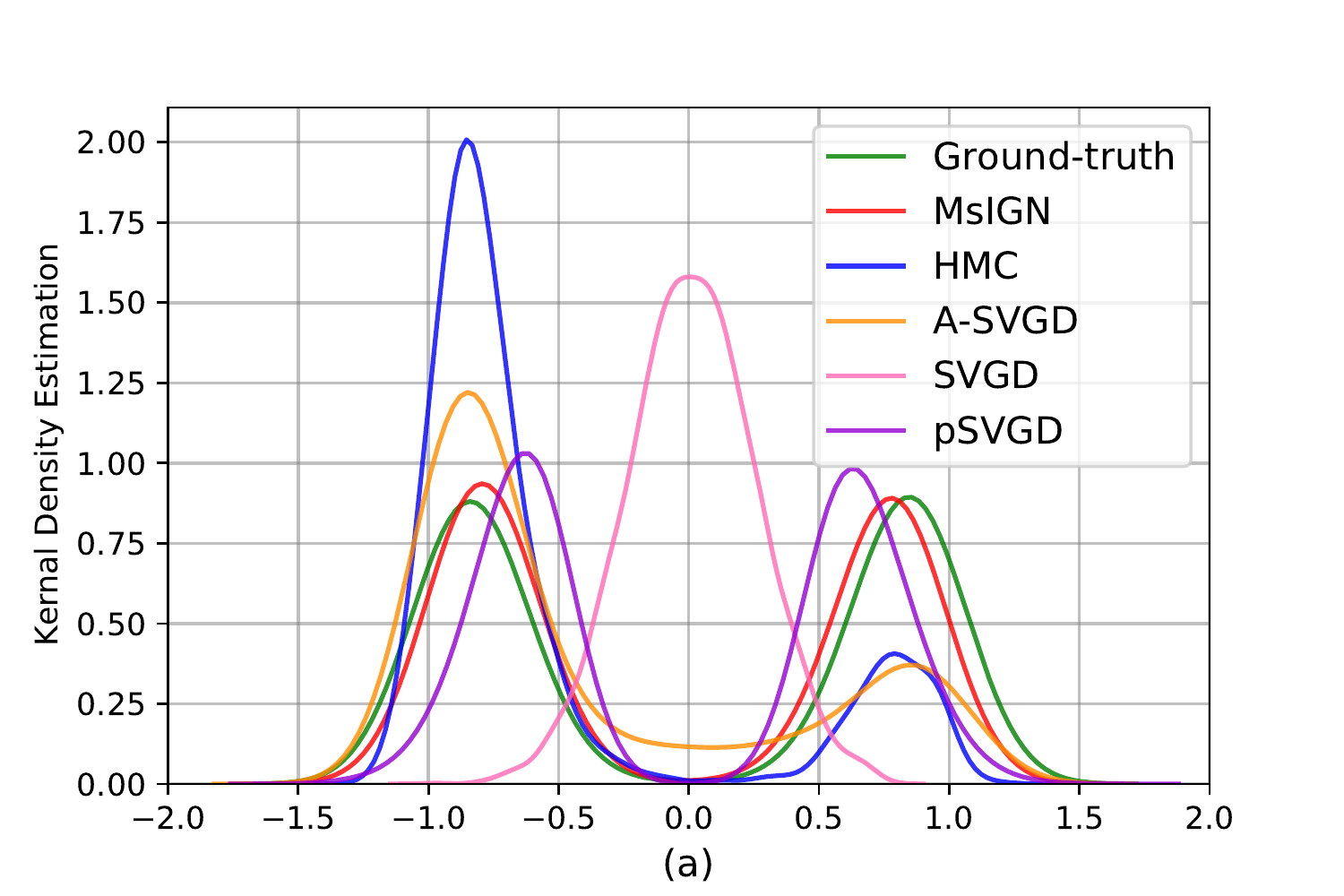}\includegraphics[width=0.6\textwidth]{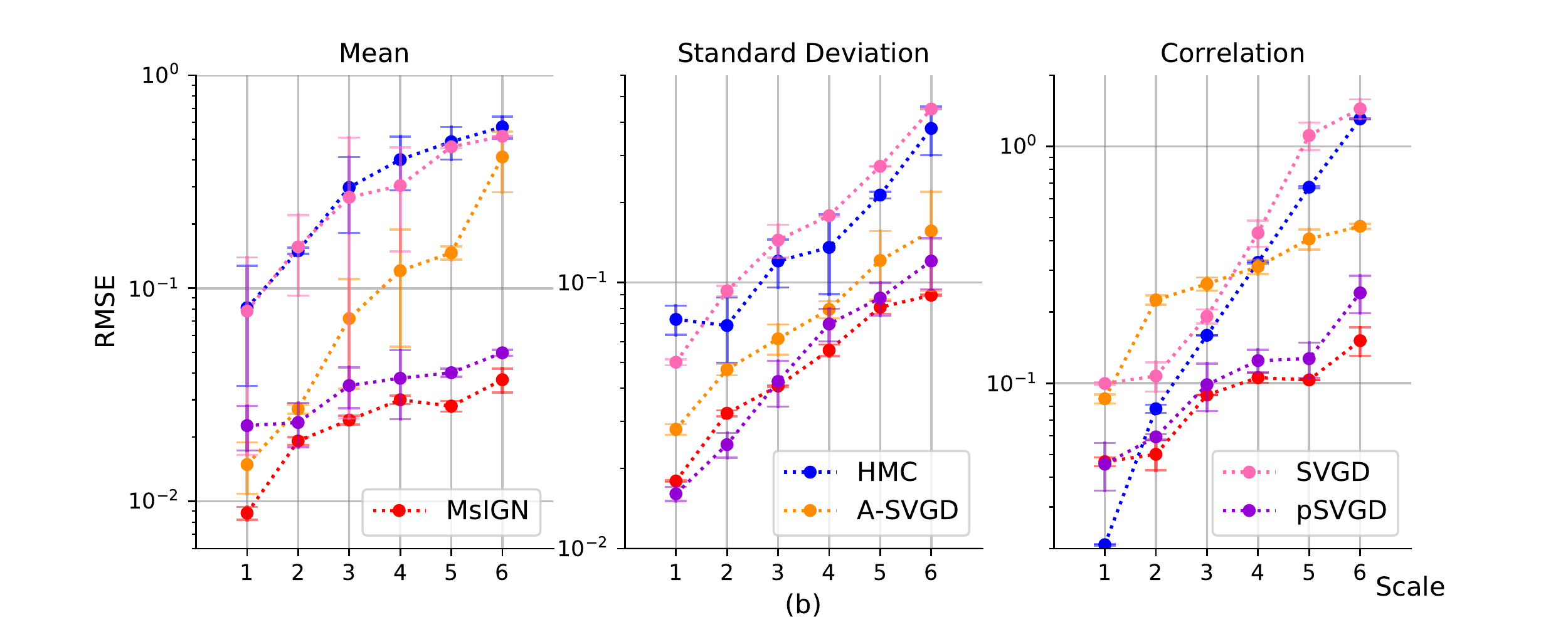}
\vspace{-4mm}
\caption{Results in the synthetic BIP. (a): Sample marginal distribution along the critical direction $w_{k^*}$. MsIGN is more robust in capturing both modes and close to ground-truth. (b): Root mean square error (RMSE) and its 95\% confidence interval of three independent experiments. MsIGN is more accurate in distribution approximation, especially at finer scale when the problem dimension is high. \iffalse The margin is statistical significant as shown by the confidence interval.\fi}
\label{fig: toy_BIP}
\vspace{-4mm}
\end{figure*}

\begin{figure*}[tbp]
\centering
\includegraphics[width=0.36\textwidth]{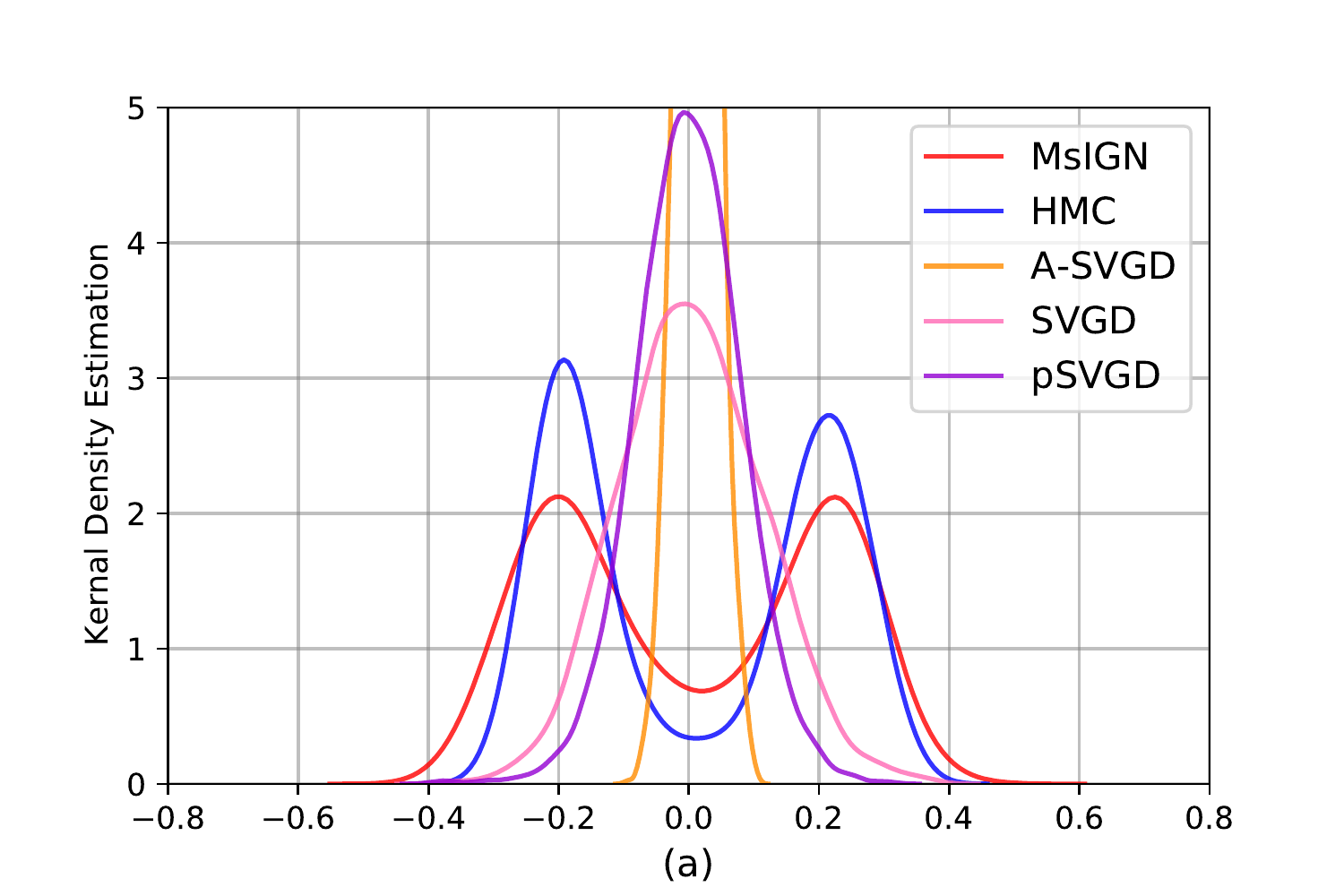}\includegraphics[width=0.6\textwidth]{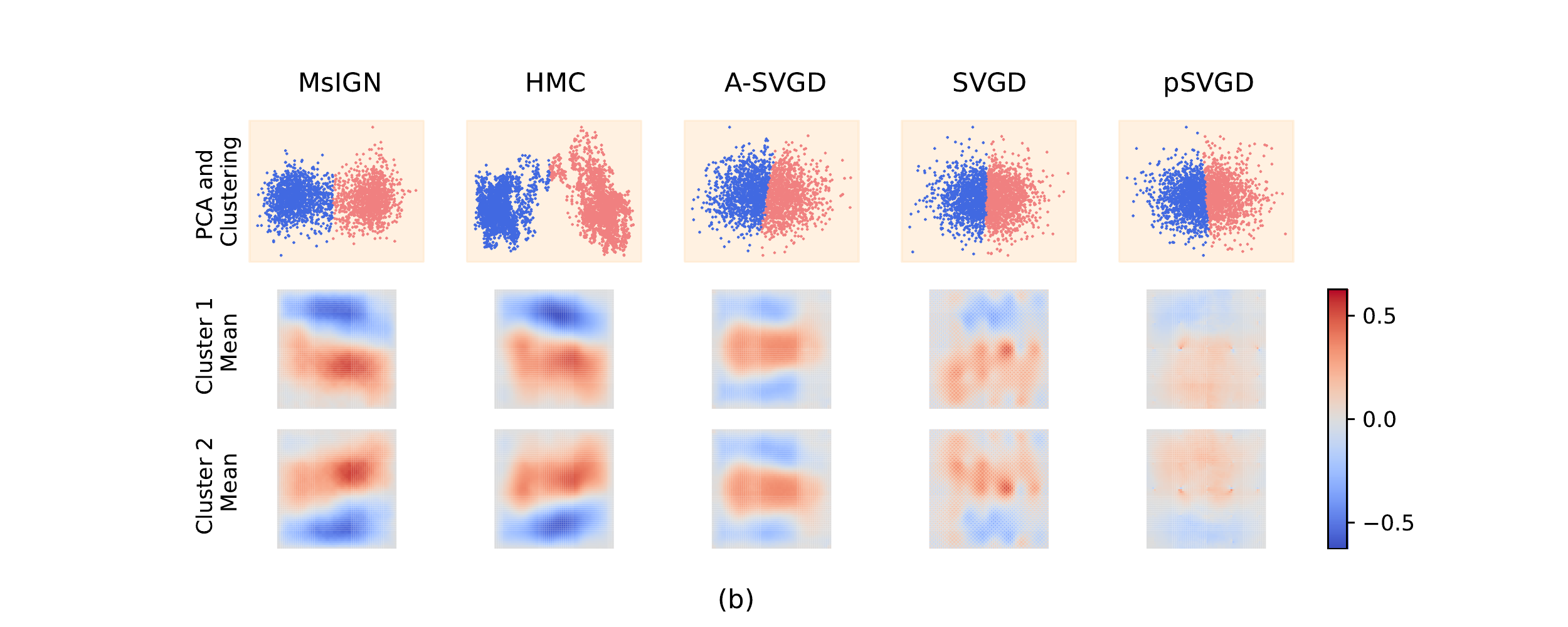}
\vspace{-4mm}
\caption{Results in the elliptic BIP. (a): Sample marginal distribution along the critical direction. MsIGN and HMC capture two modes in this marginal distribution, but the others fail. (b): Clustering result of samples. Samples of MsIGN are more balanced between two modes. The similarity of the cluster means of MsIGN and HMC implies that they both are likely to capture the correct modes.}
\label{fig: elliptic_BIP}
\vspace{-3mm}
\end{figure*}

In both experiments, we utilize average pooling with kernel size $2$ and stride $2$ as the operator $\A$, and stack several of the invertible block in Glow \citep{kingma2018glow} to build our invertible flow $F$, as mentioned in Section \ref{sec: network}.

\vspace{-2mm}
\subsection{Bayesian Inverse Problems}
\label{sec: exp_BIP}

We study two nonlinear and high-$d$ BIPs known to have at least two equally important modes in this section: one with true samples available as reference in Section \ref{sec: exp_syn_BIP}; one without true samples but close to real-world applications of subsurface flow in fluid dynamics in Section \ref{sec: exp_elp_BIP}. In both problems, sample $x$ of the target posterior $q$ is a vector on a 2-D uniform $64\times64$ lattice, which means the problem dimension $d$ is $4096$. Every $x$ is equivalent to a piece-wise constant function on the unit disk:  $x(s)$ for $s\in\Omega=\left[0, 1\right]^2$, and we don't distinguish between them thereafter. We equip $x$ with a Gaussian prior $\N\left(0, \Sigma\right)$ with $\Sigma$ as the discretization of $\beta^2\left(-\Delta\right)^{-1-\alpha}$, where $\alpha$, $\beta$ are parameters.

To make the high-$d$ inference more challenging, the target $q$ is built to be multi-modal by leveraging spatial symmetry. Combining properties of the prior defined above and the likelihood defined afterwards, the posterior is innately mirror-symmetric: $q(x)=q(x^\prime)$ if $x(s_1, s_2)=x^\prime(s_1, 1-s_2)$ for any $s=(s_1, s_2)\in\Omega$. Furthermore, we carefully select the prior and the likelihood so that $q$ has at least two modes. They are mirror-symmetric to each other and possess equal importance, see discussion in Appendix \ref{ap: exp_BIP}.

We train MsIGN following Algorithm \ref{alg: training} with $L=6$ scales. The problem dimension at scale $l$ is $d_l=2^l*2^l=4^l$. We compare MsIGN with representative approaches for high-$d$ BIPs: Hamiltonian Monte Carlo (short as HMC) \citep{neal2011mcmc}, SVGD \citep{liu2016stein}, amortized-SVGD (short as A-SVGD) \citep{feng2017learning}, and projected SVGD (short as pSVGD) \citep{chen2020projected}. Since simulating the forward map $\F$ dominates the training time cost, especially in Section \ref{sec: exp_elp_BIP} (more than $75\%$ of the wall clock time), we set a budget for the \textbf{n}umber of \textbf{f}orward \textbf{s}imulations (nFSs) for all methods for fair comparison in computational cost. For both problems, we aim at generating 2500 samples from the target. More details of experimental setting and additional numerical results can be found in Appendix \ref{ap: exp_BIP}.

\begin{figure*}[tbp]
    \centering
    \includegraphics[width=0.98\textwidth]{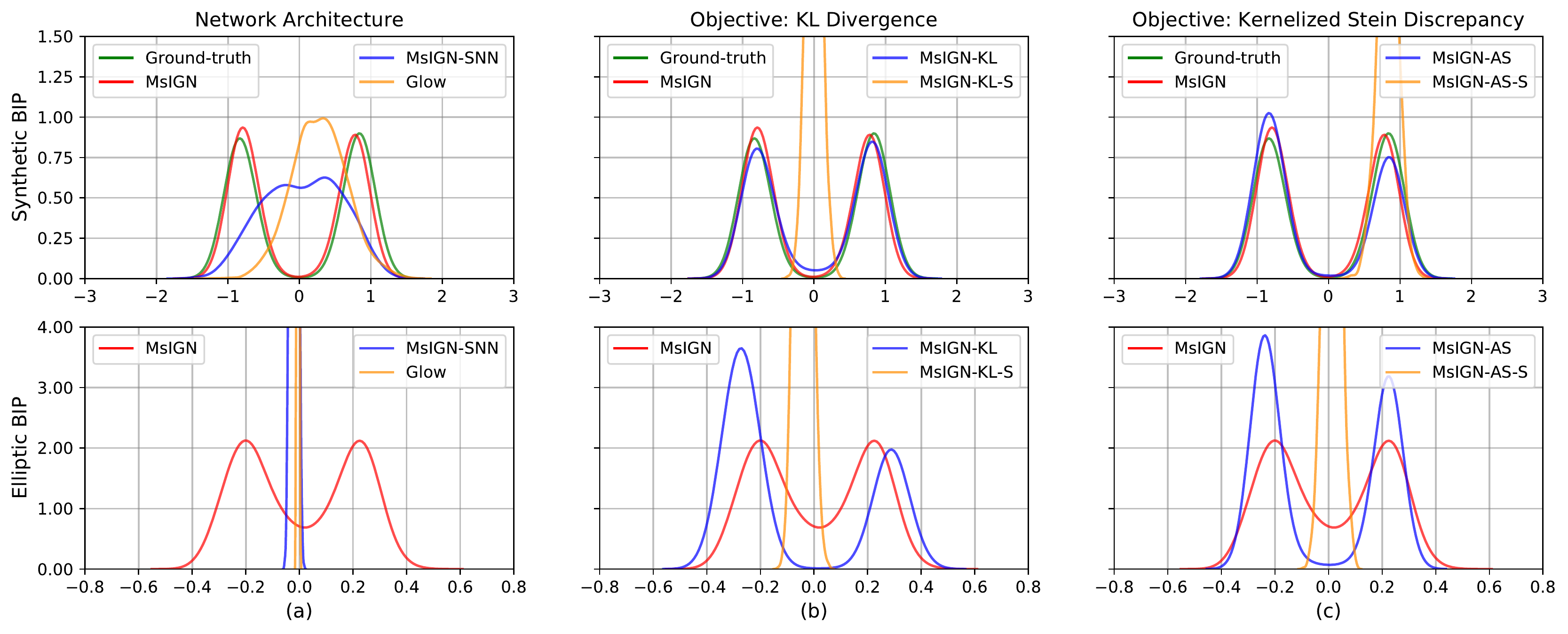}
    \vspace{-0.5cm}
    \caption{Ablation study of the network architecture and training strategy. ``MsIGN'' means our default setting: training MsIGN network with Jeffreys divergence and multi-stage strategy. Other models are named by a base model (MsIGN or Glow), followed by strings indicating its difference from the default setting. For example, ``MsIGN-KL'' refers to training MsIGN network with single KL divergence in a multi-stage way, while ``MsIGN-KL-S'' means training in a single-stage way. See Appendix \ref{ap: exp_BIP} for thorough discussion.}\label{fig: ablation_BIP}
\vspace{-3mm}
\end{figure*}

\vspace{-2mm}
\subsubsection{Synthetic Bayesian Inverse Problems}
\label{sec: exp_syn_BIP}

This problem allows access to ground-truth samples so the comparison is clear and solid. We set $\F(x)=\langle\varphi, x\rangle^2=(\int_\Omega \varphi(s)x(s)\rd s)^2$, where $\varphi(s)=\sin(\pi s_1)\sin(2\pi s_2)$. Together with the prior, our posterior can be factorized into 1-D sub-distributions, namely $q(x)=\prod_{k=1}^{d}q_{k}(\langle w_k, x\rangle)$ for some orthonormal basis $\{w_k\}_{k=1}^{d}$ of $\R^d$. This property gives us access to true samples via inversion cumulative function sampling along each direction $w_k$. Furthermore, these 1-D sub-distributions are all single modal except that there's one, which is the marginal distribution $q_{k^*}$ along direction $w_{k^*}$, with two symmetric modes. This confirms our construction of two equally important modes. The computation budget is fixed at $8\times 10^5$ nFSs.

\textbf{Multi-mode capture\hspace{3mm}} To visualize mode capture, we plot the marginal distribution of generated samples along the critical direction $w_{k^*}$, which is the source of double-modality. From Figure \ref{fig: toy_BIP}(a), MsIGN gives the best mode capture among our baselines in this $d=4096$ problem.

\textbf{Distribution approximation\hspace{3mm}} We use the root mean square errors (RMSE) of sample mean, standard deviation, and correlation, with the Jeffreys divergence to measure distribution approximation. We compare the sample mean, variance and correlation with theoretical ground-truths, and report the averaged RMSE of all sub-distributions at all scales in Figure \ref{fig: toy_BIP}(b). Additionally, since MsIGN and A-SVGD also gives density estimation, we report the Monte Carlo estimates of the Jeffreys divergence (\ref{eq: jeffreys}) with the target posterior in Table \ref{tab: jeffreys}. We can see that MsIGN has superior accuracy in approximating the target distribution.

\vspace{-5mm}
\begin{table}[H]
\caption{Jeffreys divergence $\Jef(p_\theta\vert q)$ in three independent runs.}
\label{tab: jeffreys}
\begin{center}
\begin{small}
\begin{sc}
\begin{tabular}{l|cc}
\toprule
Model & MsIGN & A-SVGD \\
\midrule
Error & 56.77$\pm$0.15 & 3372$\pm$21 \\
\bottomrule
\end{tabular}
\end{sc}
\end{small}
\end{center}
\vspace{-6mm}
\end{table}

\vspace{-2mm}
\subsubsection{Elliptic Bayesian Inverse Problems}
\label{sec: exp_elp_BIP}

This problem is a benchmark problem for high-$d$ inference from geophysics and fluid dynamics \citep{iglesias2014well, cui2016dimension}. The forward map $\F(x)=\mathcal{O}\circ\mathcal{S}(x)$, where $u=\mathcal{S}(x)$ is the solution to an elliptic partial differential equation with zero Dirichlet boundary condition:
$$-\nabla\cdot\left(e^{x(s)}\nabla u(s)\right)=f(s)\,,\quad s\in\Omega\,,$$
And $\mathcal{O}$ is linear measurements of the field function $u$:
$$\mathcal{O}(u)=\begin{bmatrix} \int_\Omega\varphi_1(s)u(s)\mathrm{d}s & \ldots & \int_\Omega\varphi_m(s)u(s)\mathrm{d}s \end{bmatrix}^T\,,$$where $f$ and $\varphi_k$ are given and fixed. The map $\mathcal{S}$ is solved by the finite element method with mesh size $1/64$. Unfortunately, there is no known access to true samples of $q$. But the trick of symmetry introduced in Section \ref{sec: exp_BIP} guarantees at least two equally important modes in the posterior. We put a $5\times 10^5$-nFS budget on our computation cost.

\textbf{Multi-mode capture\hspace{3mm}} Due to lack of true samples, we check the marginal distribution of the posterior along eigen-vectors of the prior, and pick a particular one to show if we capture double modes in Figure \ref{fig: elliptic_BIP}(a). We also confirm the capture of multiple modes by embedding samples by Principle Component Analysis (PCA) to a 2-D space. We report the clustering (by K-means) result and means of each cluster in Figure \ref{fig: elliptic_BIP}(b), where we can see that MsIGN has a more balanced capture of the symmetric posterior than HMC, while others fail to detect two modes. We refer readers to Appendix \ref{ap: exp_BIP} for more comprehensive study of mode capture ability of different methods.

\vspace{-2mm}
\subsubsection{Ablation Study}
\label{sec: exp_BIP_ablation}

We run extensive experiments to study the effectiveness of the network architecture and training strategy of MsIGN. Detailed setting and extra results are left in Appendix \ref{ap: exp_BIP}.

\textbf{Network architecture\hspace{3mm}} We replace the prior conditioning layer $\PC$ by two direct alternatives: a stochastic nearest-neighbor upsample layer independent of the prior (model named ``MsIGN-SNN''), or the split and squeeze layer in Glow design (it resumes Glow model, so we call it ``Glow''). Figure \ref{fig: ablation_BIP}(a) shows that the prior conditioning layer design is crucial to the performance of MsIGN on both problems, because neither alternatives has a successful mode capture.

\textbf{Training strategy\hspace{3mm}} We study the effectiveness of the Jeffreys divergence objective and multi-stage training. We try substituting the Jeffreys divergence with the KL divergence (marked with a suffix ``-KL'') or kernelized Stein discrepancy (which resumes A-SVGD algorithm, so we mark it with a suffix ``-AS''), and switching between multi-stage (the default, no extra suffix) or single-stage training (marked with a suffix ``-S''). We remark that single-stage training using Jeffreys divergence is infeasible because of the difficulty to estimate $\KL(q\Vert p_\theta)$. Figure \ref{fig: ablation_BIP}(b) and (c) show that, all models trained in the single-stage manner (``MsIGN-KL-S'', ``MsIGN-AS-S'') will face mode collapse. We observe that our multi-stage training strategy can benefit training with other objectives, see ``MsIGN-KL'' and ``MsIGN-AS''. We also notice that the Jeffreys divergence leads to a more balanced samples for these symmetric problems, especially for the complicated elliptic BIP.

\begin{table*}[t!]
    \caption{Bits-per-dimension value comparison with baseline models of flow-based generative networks. All models in this table do not use the ``variational dequantization'' technique in \citep{ho2019flow++}. *: Score obtained by our own reproducing experiment.}
    \label{tab: bits-per-dimension}
    \begin{center}
    \begin{small}
    \begin{sc}
    \begin{tabular}{l|ccccc}
    \toprule
    Model & MNIST & CIFAR-10 & CelebA 64 & ImageNet 32 & ImageNet 64 \\
    \midrule
    Real NVP\citep{dinh2016density} & 1.06 & 3.49 & 3.02 & 4.28 & 3.98 \\
    Glow\citep{kingma2018glow} & 1.05 & 3.35 & 2.20$^*$ & 4.09 & 3.81 \\
    FFJORD\citep{grathwohl2018ffjord} & 0.99 & 3.40 & -- & -- & -- \\
    Flow++\citep{ho2019flow++} & -- & 3.29 & -- & -- & -- \\
    i-ResNet\citep{behrmann2019invertible} & 1.05 & 3.45 & -- & -- & -- \\
    Residual Flow\citep{chen2019residual} & 0.97 & {\bf 3.28} & -- & {\bf 4.01} & 3.76 \\
    \midrule
    {\bf MsIGN} (Ours) & {\bf 0.93} & {\bf 3.28} & {\bf 2.15} & 4.03 & {\bf 3.73} \\
    \bottomrule
    \end{tabular}
    \end{sc}
    \end{small}
    \end{center}
    \vspace{-1mm}
\end{table*}

\begin{figure*}[t!]
    \centering
    \includegraphics[width=0.56\textwidth]{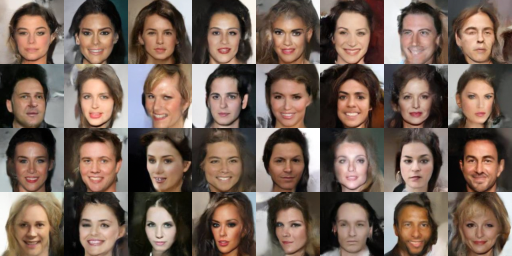}\hspace{0.02\textwidth}\includegraphics[width=0.42\textwidth]{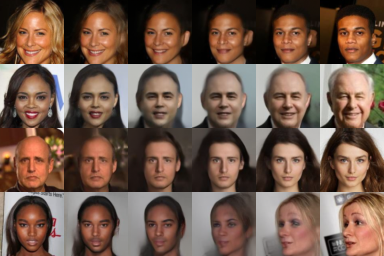}
    \vspace{-6mm}
    \caption{Left: Synthesized CelebA images of resolution $64\times 64$ with temperature 0.9. Right: Linear interpolation in latent space shows MsIGN's parameterization of natural image manifold is semantically meaningful. For images $x_1, x_2$ at the left and right ends, we retrieve their latent feature by $z_i=T^{-1}(x_i;\theta), i=1,2$, and then interpolate between them by $T((1-\lambda)z_1+\lambda z_2;\theta)$ for $\lambda=0.2, 0.4, 0.6, 0.8$.}
    \label{fig: img_synthesize_interpolate}
\end{figure*}

\vspace{-2mm}
\subsection{Image Synthesis}
\label{sec: exp_Image}

The transport map approach to Bayesian inference has two critical difficulties: the model capacity and the training effectiveness. Since the distribution of images is complicated and multi-modal, we present the image synthesis task result to show case the model capacity of the MsIGN. It also provides a good test bed for our MsIGN to benchmark with other flow-based generative networks.

We train MsIGN by maximum likelihood estimation. We assume a simple Gaussian prior $\rho$ for natural images, whose covariance is a scalar matrix learned from the data. See Appendix \ref{ap: exp_Image} for experimental details and additional results.

\begin{figure*}[t!]
    \vspace{1mm}
    \centering
    \includegraphics[width=\textwidth]{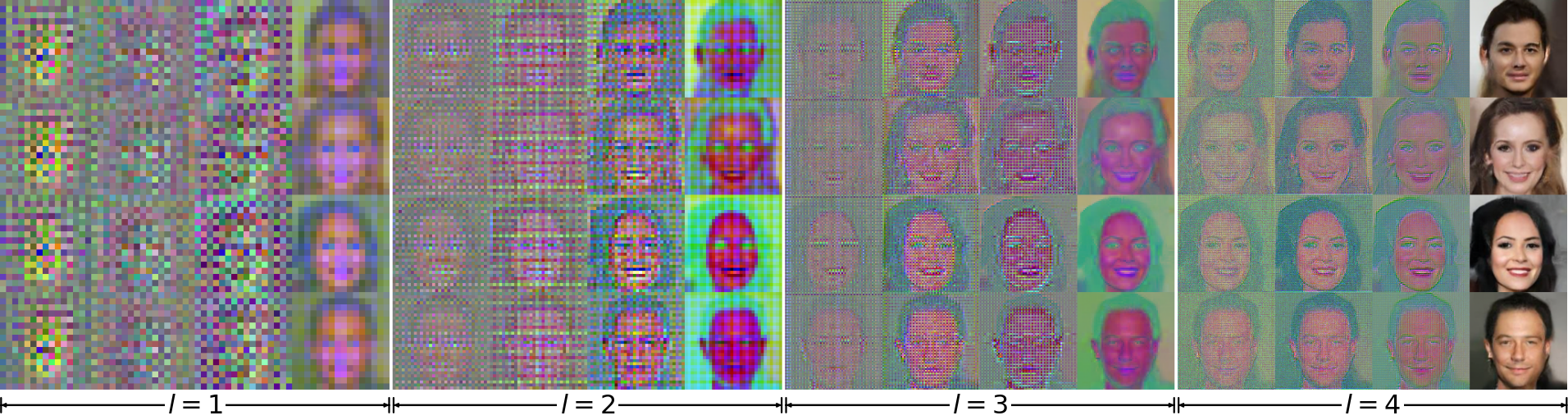}
    \vspace{-6mm}
    \caption{Visualization of internal activation shows the interpret-ability of MsIGN hidden neurons. This MsIGN model has $L=4$ scales. From left to right, we take $4$ snapshots (head, two trisection points, and tail) in each invertible flow $F_l$ for $l=1, 2, 3, 4$, to show how MsIGN progressively generates new samples from low to high resolution.}
    \label{fig: img_hierarchy}
    \vspace{-2mm}
\end{figure*}

We report the bits-per-dimension value comparison with baseline models in Table \ref{tab: bits-per-dimension}. Our MsIGN is superior in number and also is more efficient in parameter size: for example, MsIGN uses $24.4\%$ fewer parameters than Glow for CelebA 64, and uses $37.4\%$ fewer parameters than Residual Flow for ImageNet 64.

Figure \ref{fig: img_synthesize_interpolate} shows synthesized images of MsIGN from CelebA data set, and linear interpolation of real images in the latent feature space. In Figure \ref{fig: img_hierarchy}, we visualize internal activations at checkpoints in the invertible flow at different scales which demonstrate the interpret-ability of MsIGN.
\vspace{-2mm}

\section{Conclusion}
\label{sec: conclude}

For high-dimensional Bayesian inference problems with multiscale structure, we propose Multiscale Invertible Generative Networks (MsIGN) and associated training algorithms to approximate the posterior. We demonstrate the potential of this approach in high-dimensional (up to 4096) Bayesian inference problems, leaving several important directions as future work. The network architecture also achieves superior performance over benchmarks in various image synthesis tasks. We plan to apply this methodology to other Bayesian inference problems, e.g., Bayesian deep learning with multiscale structure in model width or depth (e.g., \citep{chang2017multi,haber2018learning}) and data assimilation problem with multiscale structure in the temporal variation (e.g., \citep{giles2008multilevel}).

\bibliography{reference}
\bibliographystyle{icml2021}

\clearpage

\appendix

\section{Proof of Theorem \ref{thm: gaussian_condition}}
\label{ap: proof_gaussian_condition}

In this section, we prove Theorem \ref{thm: gaussian_condition} which gives closed-form formulation for the prior conditional layer $\PC$ in the Gaussian prior case.

We first introduce a powerful tool named partition of unity in Lemma in order to prove Theorem \ref{thm: gaussian_condition}. We adopt the notations in Section \ref{sec: network} here.

\begin{lemma}
Assume $\A\in\R^{d_c\times d}$ ($d_c<d$) has full row-rank, i.e. $\rank(\A)=d_c$, there exists a matrix $\tilde{\A}\in\R^{(d-d_c)\times d}$ such that $\A\tilde{\A}^T=0\in\R^{d_c\times(d-d_c)}$. And for any symmetric positive definite matrix $\Sigma$, we have the following decomposition of the identity (unit) matrix $I_{d}\in\R^{d\times d}$:
\begin{align*}
    I_{d} =& \Sigma^{\frac{1}{2}}\A^T(\A\Sigma\A^T)^{-1}\A\Sigma^{\frac{1}{2}} \\
    &+ \Sigma^{-\frac{1}{2}}\tilde{\A}^T(\tilde{\A}\Sigma^{-1}\tilde{\A}^T)^{-1}\tilde{\A}\Sigma^{-\frac{1}{2}}
\end{align*}
\label{lem: identity_decomposition}
\end{lemma}

{\it Proof}: The matrix $\tilde{\A}$ is in fact the orthogonal complement of $\A$. Let $V\in\R^{d}$ be the row space of $\A$, then $\dim(V)=d_c<d$, so the orthogonal complement $V^\perp$ of the subspace $V\subset\R^d$ is non-trivial: $\dim(V^\perp)=d-d_c>0$. Collect a basis of $V^\perp$ and pack them in rows, we have a matrix $\tilde{\A}\in\R^{(d-d_c)\times d}$. By construction we know $\A\tilde{\A}^T=0$, because $V$ and $V^\perp$ are orthogonal to each other.

Now consider the following matrix $\Omega\in\R^{d\times d}$:
\begin{align*}
    \Omega := \begin{bmatrix}
    (\A\Sigma\A^T)^{-\frac{1}{2}}\A\Sigma^{\frac{1}{2}} \\
    (\tilde{\A}\Sigma^{-1}\tilde{\A}^T)^{-\frac{1}{2}}\tilde{\A}\Sigma^{-\frac{1}{2}}
    \end{bmatrix}^T\,.
\end{align*}
We have
\begin{align*}
    &\Omega^T\Omega\\
    =&\begin{bmatrix}
    (\A\Sigma\A^T)^{-\frac{1}{2}}\A\Sigma^{\frac{1}{2}} \\
    (\tilde{\A}\Sigma^{-1}\tilde{\A}^T)^{-\frac{1}{2}}\tilde{\A}\Sigma^{-\frac{1}{2}}
    \end{bmatrix}\begin{bmatrix}
    (\A\Sigma\A^T)^{-\frac{1}{2}}\A\Sigma^{\frac{1}{2}} \\
    (\tilde{\A}\Sigma^{-1}\tilde{\A}^T)^{-\frac{1}{2}}\tilde{\A}\Sigma^{-\frac{1}{2}}
    \end{bmatrix}^T\,, \\
    =&\begin{bmatrix}
    \left(\Omega^T\Omega\right)_{11} & \left(\Omega^T\Omega\right)_{12} \\
    \left(\Omega^T\Omega\right)_{21} & \left(\Omega^T\Omega\right)_{22} \\
    \end{bmatrix}\,,
\end{align*}
where, since $\A\tilde{\A}^T=0$ and $\Sigma$ is symmetric: $\Sigma=\Sigma^T$,
\begin{align*}
    \left(\Omega^T\Omega\right)_{11} &= (\A\Sigma\A^T)^{-\frac{1}{2}}\A\Sigma\A^T(\A\Sigma\A^T)^{-\frac{1}{2}}=I_{d_c}\,,\\
    \left(\Omega^T\Omega\right)_{12} &= (\A\Sigma\A^T)^{-\frac{1}{2}}\A\Sigma^{\frac{1}{2}}\Sigma^{-\frac{1}{2}}\tilde{\A}^T(\tilde{\A}\Sigma^{-1}\tilde{\A}^T)^{-\frac{1}{2}}\\
    &= (\A\Sigma\A^T)^{-\frac{1}{2}}\A\tilde{\A}^T(\tilde{\A}\Sigma^{-1}\tilde{\A}^T)^{-\frac{1}{2}}=0\,,\\
    \left(\Omega^T\Omega\right)_{21} &= (\tilde{\A}\Sigma^{-1}\tilde{\A}^T)^{-\frac{1}{2}}\tilde{\A}\Sigma^{-\frac{1}{2}}\Sigma^{\frac{1}{2}}\A^T(\A\Sigma\A^T)^{-\frac{1}{2}}\\
    &= (\tilde{\A}\Sigma^{-1}\tilde{\A}^T)^{-\frac{1}{2}}\tilde{\A}\A^T(\A\Sigma\A^T)^{-\frac{1}{2}}=0\,,\\
    \left(\Omega^T\Omega\right)_{22} &= (\tilde{\A}\Sigma^{-1}\tilde{\A}^T)^{-\frac{1}{2}}\tilde{\A}\Sigma^{-1}\tilde{\A}^T(\tilde{\A}\Sigma^{-1}\tilde{\A}^T)^{-\frac{1}{2}}\\&=I_{d-d_c}\,.
\end{align*}
So $\Omega$ is in fact a $d\times d$ orthonormal matrix, because
\begin{align*}
    \Omega^T\Omega&=\begin{bmatrix}
    I_{d_c} & \\
     & I_{d-d_c}
    \end{bmatrix}=I_{d}\,.
\end{align*}

The orthonormality of $\Omega$ also implies  $\Omega\Omega^T=I_{d}$, which can expand as
\begin{align*}
    &I_{d}=\Omega\Omega^T\\
    =&\begin{bmatrix}
    (\A\Sigma\A^T)^{-\frac{1}{2}}\A\Sigma^{\frac{1}{2}} \\
    (\tilde{\A}\Sigma^{-1}\tilde{\A}^T)^{-\frac{1}{2}}\tilde{\A}\Sigma^{-\frac{1}{2}}
    \end{bmatrix}^T\begin{bmatrix}
    (\A\Sigma\A^T)^{-\frac{1}{2}}\A\Sigma^{\frac{1}{2}} \\
    (\tilde{\A}\Sigma^{-1}\tilde{\A}^T)^{-\frac{1}{2}}\tilde{\A}\Sigma^{-\frac{1}{2}}
    \end{bmatrix} \\
    =&\Sigma^{\frac{1}{2}}\A^T(\A\Sigma\A^T)^{-1}\A\Sigma^{\frac{1}{2}}\\
    &+ \Sigma^{-\frac{1}{2}}\tilde{\A}^T(\tilde{\A}\Sigma^{-1}\tilde{\A}^T)^{-1}\tilde{\A}\Sigma^{-\frac{1}{2}}\,.
\end{align*}
And this proves Lemma \ref{lem: identity_decomposition}. $\square$

Now we give the proof to Theorem \ref{thm: gaussian_condition}.

\begin{theorem_repeat}{thm: gaussian_condition}
Suppose that $\rho$ is a Gaussian with density $\N(x; 0, \Sigma)$ where the covariance $\Sigma$ is positive definite, then with $U^c := \Sigma \A^T(\A\Sigma\A^T)^{-1}\in\R^{d\times d_c}$ and $\Sigma^c := \Sigma-\Sigma \A^T(\A\Sigma\A^T)^{-1}\A\Sigma\in\R^{d\times d}$, we have
$$\rho(x|\A x=x_{c}) = \N(x; U^c x_{c}, \Sigma^c)\,.$$
Furthermore, there exists a matrix $W\in\R^{d\times(d-d_c)}$ such that $\Sigma^c=WW^T$, and the prior conditioning layer $\PC$ can be given as, with $z\in\R^{d-d_c}$ being standard Gaussian
$$x=\PC(x_c, z)=U^cx_c+Wz\,,$$
and $\PC$ is invertible between $x$ and $(x_c, z)$.
\end{theorem_repeat}

{\it Proof}: The conditional probability rule suggests that
$$\rho(x|\A x=x_{c})=\left.\rho(x)\middle/\left(\int_{\left\{x^\prime: \A x^\prime=x_c\right\}}\rho(x^\prime)\rd x^\prime\right)\right.$$
When $x_c$ is given and fixed, the denominator in the above is a constant with respect to $x$. Therefore, since we recall that the prior $\rho$ is a Gaussian $\N(0, \Sigma)$, we have
$$\log\rho(x|\A x=x_c)=\log\rho(x) - C^\prime=-\frac{1}{2}x^T\Sigma^{-1}x + C\,,$$
where $C$ is a constant that only depends on $x_c$ and $\Sigma$. Since $\log\rho(x|\A x=x_c)$ is a quadratic function of $x$, $\rho(x|\A x=x_c)$ should also be a Gaussian distribution. To determine this distribution we only need to calculate its mean $\E\left[x|\A x=x_c\right]$ and covariance $\Cov\left[x|\A x=x_c\right]$.

With $U^c := \Sigma\A^T(\A\Sigma\A^T)^{-1}$, we decompose $x=(x-U^c\A x)+U^c\A x$. We will prove later that $x-U^c\A x$ is independent from $\A x$, so that,
\begin{align*}
    &\E\left[x|\A x=x_c\right]=\E\left[(x-U^c\A x)+U^c\A x|\A x=x_c\right]\\
    =&\E\left[x-U^c\A x|\A x=x_c\right]+\E\left[U^c\A x|\A x=x_c\right]\\
    =&0+U^cx_c=U^cx_c\,.
\end{align*}

To show $x-U^c\A x=(I_{d}-U^c\A)x$ is independent from $\A x$, where $I_d\in\R^{d\times d}$ is the identity (unit) matrix, we notice that they are both linear transformation of the Gaussian variable $x$, so their joint distributions should also be a Gaussian, and their covariance can be computed as
\begin{align*}
    \Cov\left[(I_{d}-U^c\A)x, \A x\right]=(I_{d}-U^c\A)\Sigma\A^T\,.
\end{align*}
Notice that $U^c\A\Sigma\A^T=\Sigma\A^T(\A\Sigma\A^T)^{-1}\A\Sigma\A^T=\Sigma\A^T$, so $(I_{d}-U^c\A)\Sigma\A^T=\Sigma\A^T-\Sigma\A^T=0$. Thus, $x-U^c\A x=(I_{d}-U^c\A)x$ is independent from $\A x$.

Finally, since $\E\left[x|\A x=x_c\right]=U^cx_c$, we calculate
$$\Cov\left[x|\A x=x_c\right]=\Cov\left[x-U^c\A x|\A x=x_c\right]\,.$$
Because $x-U^c\A x=(I_{d}-U^c\A)x$ is independent from $\A x$, we can drop the condition and write:
\begin{align*}
    &\Cov\left[x|\A x=x_c\right]=\Cov\left[x-U^c\A x\right]\\
    =&(I_{d}-U^c\A)\Sigma(I_{d}-U^c\A)^T\\
    =&\Sigma-U^c\A\Sigma-\Sigma\A^T\left(U^c\right)^T+U^c\A\Sigma\A^T\left(U^c\right)^T\,.
\end{align*}
Plug in the definition of $U^c$, we find
$$\Cov\left[x|\A x=x_c\right]=\Sigma-\Sigma\A^T(\A\Sigma\A^T)^{-1}\A\Sigma=\Sigma^c\,.$$

Therefore we can conclude that
$$\rho(x|\A x=x_c) = \N(x; U^c x_c, \Sigma^c)\,.$$

For the close form of $\PC$, we first notice that
\begin{align*}
    \Sigma^c &= \Sigma - \Sigma\A^T(\A\Sigma\A^T)^{-1}\A\Sigma\\
    &= \Sigma^{\frac{1}{2}}\left(I_d - \Sigma^{\frac{1}{2}}\A^T(\A\Sigma\A^T)^{-1}\A\Sigma^{\frac{1}{2}}\right)\Sigma^{\frac{1}{2}}\,.
\end{align*}
Using the identity decomposition in Lemma \ref{lem: identity_decomposition}, we have
\begin{align*}
    \Sigma^c 
    &=\Sigma^{\frac{1}{2}}\Sigma^{-\frac{1}{2}}\tilde{\A}^T(\tilde{\A}\Sigma^{-1}\tilde{\A}^T)^{-1}\tilde{\A}\Sigma^{-\frac{1}{2}}\Sigma^{\frac{1}{2}}\\
    &=\tilde{\A}^T(\tilde{\A}\Sigma^{-1}\tilde{\A}^T)^{-1}\tilde{\A}\,.
\end{align*}
Now set $W=\tilde{\A}^T(\tilde{\A}\Sigma^{-1}\tilde{\A}^T)^{-\frac{1}{2}}$, then $W\in\R^{d\times(d-d_c)}$ and $\Sigma^c=WW^T$. With the existence of $W$, it remains to show that $U^cx_c+Wz$ follows the same distribution as $\rho(x|\A x=x_c)$ for a given $x_c$, and Gaussian noise $z$.

We first check if the condition $\A x=x_c$ is satisfied,
\begin{align}
    &\A\left(U^cx_c+Wz\right)=\A U^cx_c+\A Wz\nonumber\\
    =&\A U^cx_c+\A\tilde{\A}^T(\tilde{\A}\Sigma^{-1}\tilde{\A}^T)^{-\frac{1}{2}}z=\A U^cx_c\nonumber\\
    =&\A\Sigma\A^T(\A\Sigma\A^T)^{-1}x_c=x_c\,.\label{eq: PCinverse1}
\end{align}
Thus the condition is satisfied. On the other hand, with $x_c$ given and fixed, and $z$ being Gaussian noise, $U^cx_c+Wz$ follows a Gaussian distribution with mean $U^cx_c$ and covariance $WW^T=\Sigma^c$. Therefore, the prior conditioning layer $\PC$ can be given as
$$x=\PC(x_c, z)=U^cx_c+Wz\,.$$

Finally, to show the invertibility of $\PC$ between $x$ and $(x_c, z)$, it remains to show how to map $x$ back to $x_c$ and $z$. We claim that the inversion is given by
$$(x_c, z)=\PC^{-1}(x)=(\A x, (\tilde{\A}\Sigma^{-1}\tilde{\A}^T)^{-\frac{1}{2}}\tilde{\A}\Sigma^{-1}x)\,.$$
The first part holds true because
$$\A x=\A\left(U^cx_c+Wz\right)=x_c$$
as shown in (\ref{eq: PCinverse1}). The second part holds because, when plug in $x=U^cx_c+Wz$, we notice that
\begin{align*}
    &(\tilde{\A}\Sigma^{-1}\tilde{\A}^T)^{-\frac{1}{2}}\tilde{\A}\Sigma^{-1}U^c\\
    =&(\tilde{\A}\Sigma^{-1}\tilde{\A}^T)^{-\frac{1}{2}}\tilde{\A}\Sigma^{-1}\Sigma\A^T(\A\Sigma\A^T)^{-1}=0\,,
\end{align*}
and similarly
\begin{align*}
    &(\tilde{\A}\Sigma^{-1}\tilde{\A}^T)^{-\frac{1}{2}}\tilde{\A}\Sigma^{-1}W\\
    =&(\tilde{\A}\Sigma^{-1}\tilde{\A}^T)^{-\frac{1}{2}}\tilde{\A}\Sigma^{-1}\tilde{\A}^T(\tilde{\A}\Sigma^{-1}\tilde{\A}^T)^{-\frac{1}{2}}=I\,.
\end{align*}
Therefore, $(\tilde{\A}\Sigma^{-1}\tilde{\A}^T)^{-\frac{1}{2}}\tilde{\A}\Sigma^{-1}x=0x_c+Iz=z$. So the invertibility of $\PC$ is guaranteed. $\square$

We remark that $\PC$ is not unique, as for any orthonormal matrix $P\in\R^{(d-d_c)\times(d-d_c)}$, the map $x=U^cx_c+WPz$ is also a valid candidate for $\PC$.

\section{Comparison of the Jeffreys divergence and Kullback-Leibler divergence}
\label{ap: Jef_KL_comparison}

The KL divergence sometimes can be inefficient to detect multi-modes: it could be easily trapped by a local minimum that misses some modes or is far from the ground-truth. We support our claim by a concrete example below.

Given $\sigma>0$, let $q$ be a 1-D Gaussian mixture model, with parameters $\mu_1$ and $\mu_2$ unknown but fixed:
$$q(x)=\frac{1}{2}\left(\N(x; \mu_1, \sigma^2)+\N(x; \mu_2, \sigma^2)\right)\,.$$
Our parametric model $p$ is also a 1-D Gaussian mixture model with parameter $\theta=(\theta_1, \theta_2)$: 
$$p_\theta(x)=\frac{1}{2}\left(\N(x; \theta_1, \sigma^2)+\N(x; \theta_2, \sigma^2)\right)\,.$$
Setting $\mu_1=-\mu_2=1.5$, and $\sigma=0.25$, we plot the landscape of single-sided KL divergences $\KL(p_\theta\Vert q)$ and $\KL(q \Vert p_\theta)$, and the Jeffreys divergence $\Jef(p_\theta\Vert q)$ as functions of $\theta=(\theta_1, \theta_2)$ in Figure \ref{fig: KL_vs_Jeffreys}.

It is now clear that $\KL(p \Vert q)$ alone might guide the training towards the local minima around $(1.5, 1.5)$ or $(-1.5, -1.5)$, where only one mode of $q$ is captured, see Figure \ref{fig: KL_vs_Jeffreys}. We explain this phenomenon as $\KL(p \Vert q)=\E_{p}\left[\log(p/q)\right]=\int p(x)\left(\log p(x)-\log q(x)\right)\rd x$ becomes small as long as $p$ is close to zero wherever $q$ close to zero. \citep{nielsen2009sided} describes this property as ``zero-forcing", and observes that $\KL(p \Vert q)$ will be small when high-density region of $p$ is covered by that of $q$. However, it doesn't strongly enforce $p$ to capture all high-density region of $q$. In our example, when $(\theta_1, \theta_2)=(1.5, 1.5)$ or $(-1.5, -1.5)$, the only high-density region of $p$ (around $1.5$ \textbf{or} $-1.5$) is a strict subset of high-density region of $q$ (around both $1.5$ and $-1.5$), and thus it attains a local minimum of $\KL(p \Vert q)$.

\begin{figure}[tbp]
    \centering
    \includegraphics[width=0.45\columnwidth]{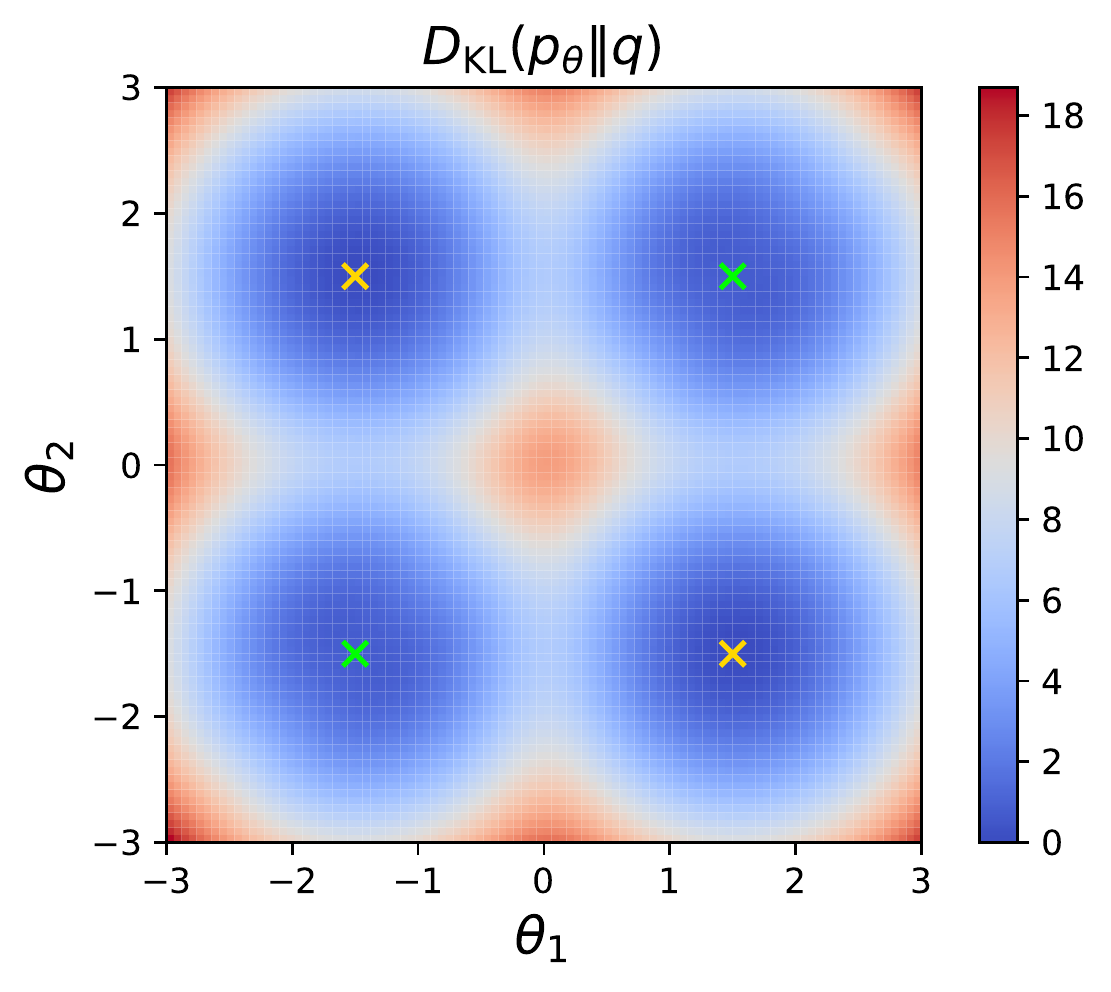}
    \includegraphics[width=0.45\columnwidth]{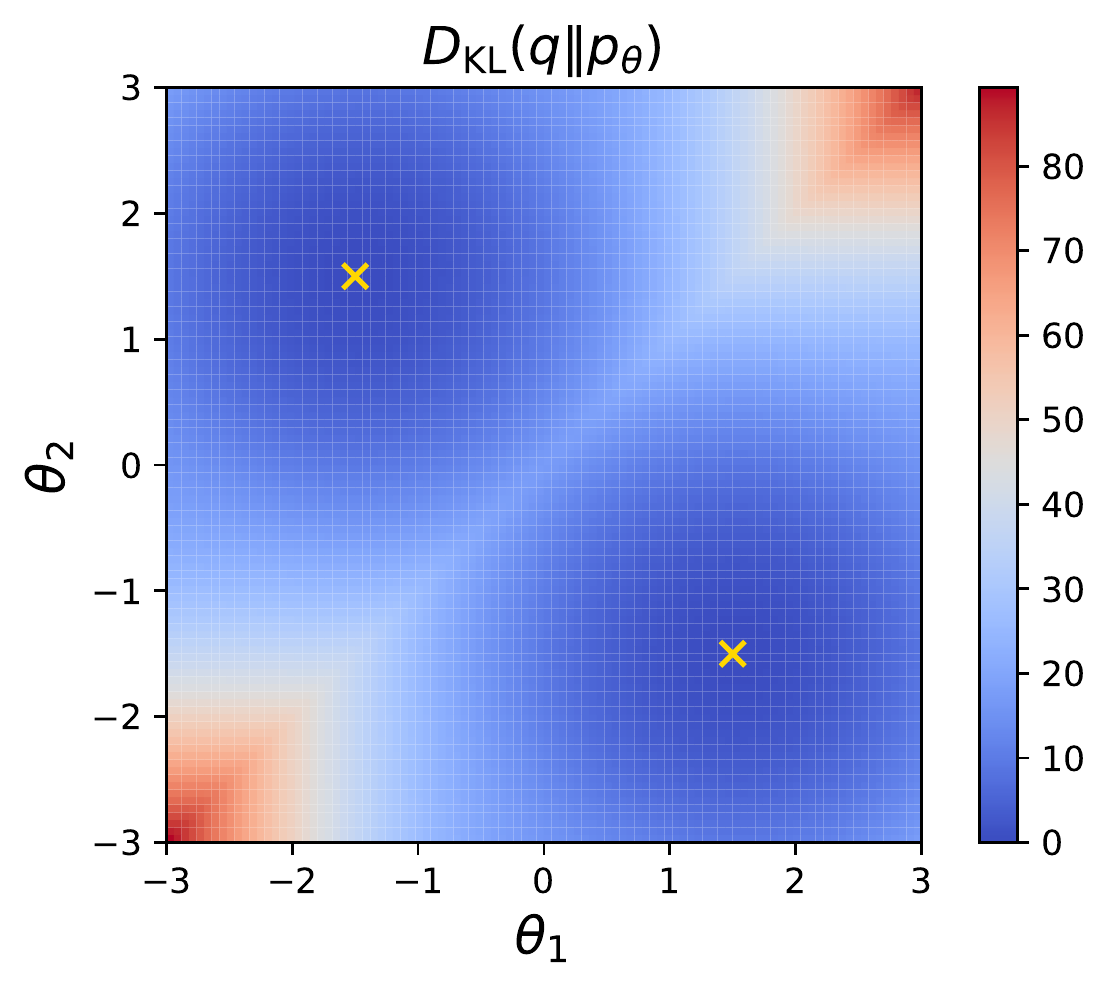}\\
    \includegraphics[width=0.45\columnwidth]{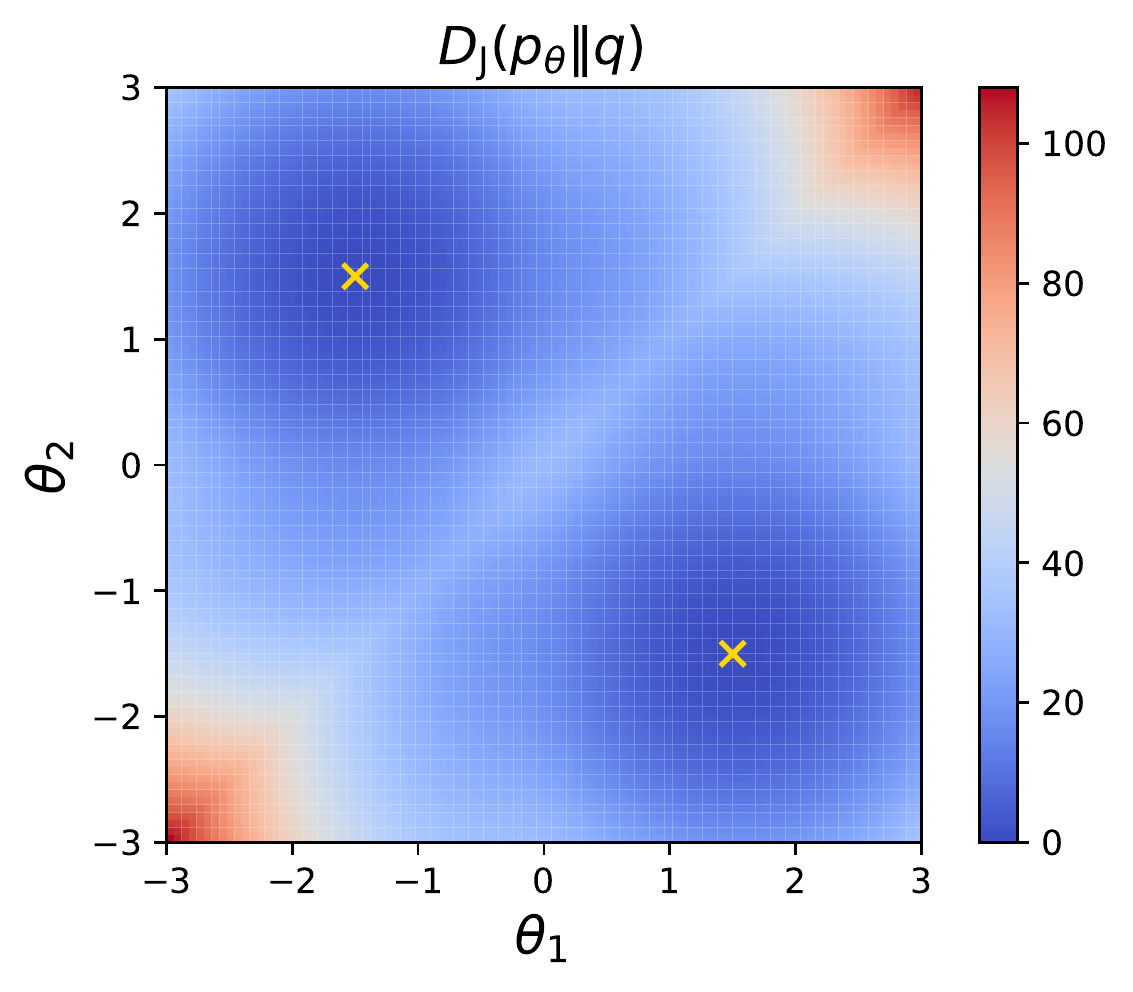}\includegraphics[width=0.45\columnwidth]{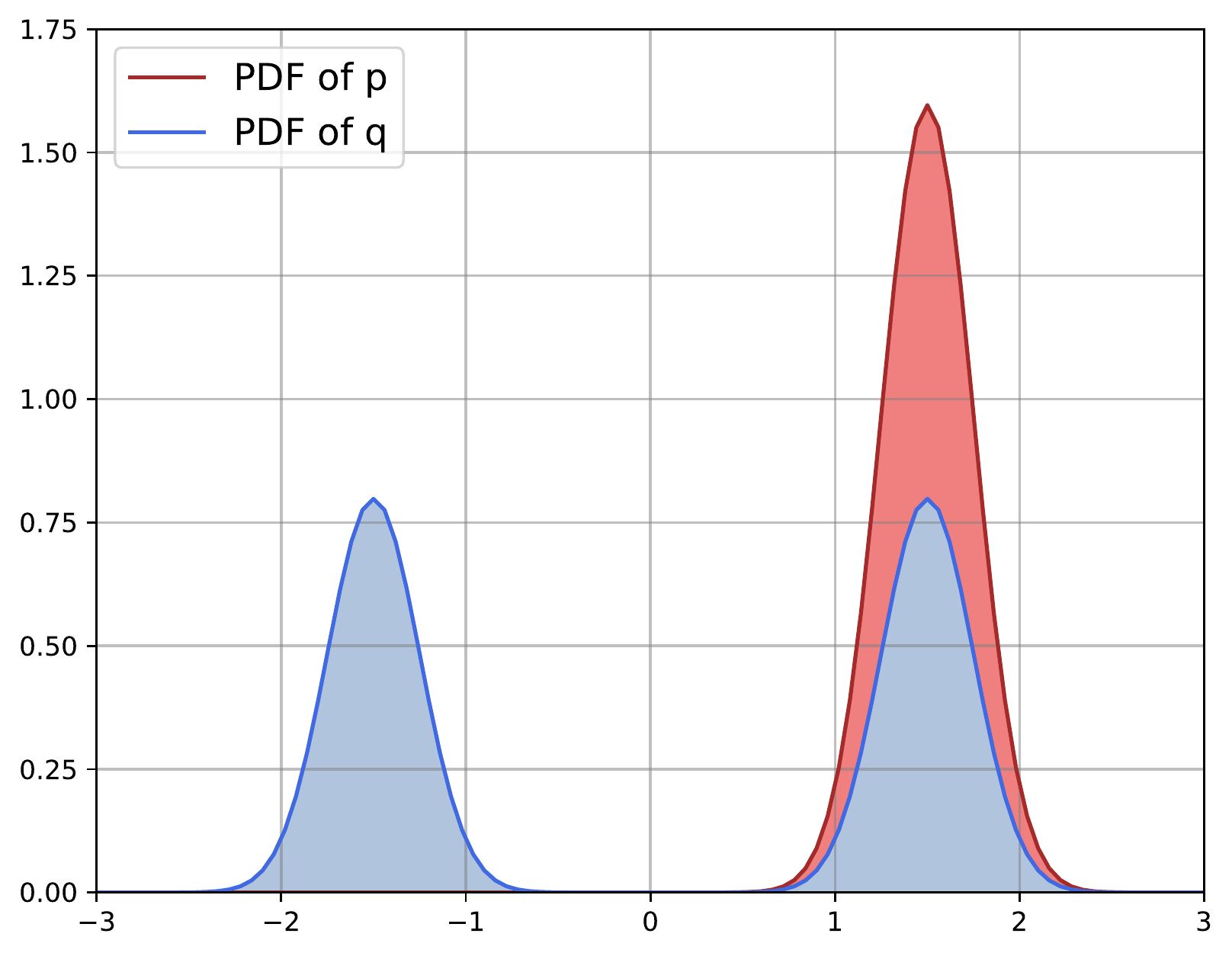}
    \caption{Landscape of $\KL(p_\theta \Vert q)$ (upper left), $\KL(q \Vert p_\theta)$ (upper right), and $\Jef(p_\theta \Vert q)$ (lower left), density function of $p_\theta$ and $q$ when they reach one of the local minima (lower right). We mark the global minima (ground-truth) by golden cross, and other local minima by green cross.}
    \vspace{-5mm}
    \label{fig: KL_vs_Jeffreys}
\end{figure}

We also argue that the other KL divergence $\KL(q \Vert p)$ alone faces the risk as well. Similarly, $\KL(q \Vert p)=\E_{q}\left[\log(q/p)\right]=\int q(x)\left(\log q(x)-\log p(x)\right)\rd x$ becomes small as long as $q$ is close to zero wherever $p$ is close to zero. Thus if $p$ captures all modes in $q$ but also contains some extra modes, described as ``zero-avoiding" in \citep{nielsen2009sided}, we could also observe a small value of $\KL(q \Vert p)$. Therefore, we choose to use the Jeffreys divergence as a robust learning objective to capture multi-modes.

\section{Proof of Theorem \ref{thm: jeffreys}}
\label{ap: proof_jeffreys}

\begin{theorem_repeat}{thm: jeffreys}
The Jeffreys divergence and its derivative to $\theta$ admit the following formulation which can be estimated by the Monte Carlo method  {\it without} samples from $q$,
\vspace{-0.1cm}
\begin{align}
    \Jef (p_\theta \Vert q)=&~\E_{p_\theta}\left[ \log\frac{p_\theta}{q} \right]+\E_{\tilde{q}} \left[ \frac{q}{\tilde{q}} \log\frac{q}{p_\theta} \right]\,.\tag{\ref{eq: compute_jeffreys}} \\
    \begin{split}
        \frac{\partial}{\partial\theta}\Jef ( p_\theta \Vert q )=&~\E_{p_\theta}\left[\left(1+\log\frac{p_\theta}{q}\right)\frac{\partial \log p_\theta}{\partial\theta}\right]\\
        &~-\E_{\tilde{q}}\left[\frac{q}{\tilde{q}}\frac{\partial \log p_\theta}{\partial\theta}\right]\,.
    \end{split}\tag{\ref{eq: compute_grad_jeffreys}}
\end{align}
Furthermore, the Monte Carlo estimation doesn't need the normalizing constant $Z$ in (\ref{eq: bayes}) as it can cancel itself.
\end{theorem_repeat}

{\it Proof}: Equation (\ref{eq: compute_jeffreys}) can be seen from
\begin{align*}
    &\E_{\tilde{q}} \left[ \frac{q}{\tilde{q}} \log\frac{q}{p_\theta} \right]=\int \tilde{q}(x)\frac{q(x)}{\tilde{q}(x)} \log\frac{q(x)}{p_\theta(x)}\rd x\\
    =&\int q(x)\log\frac{q(x)}{p_\theta(x)}\rd x=\E_{q} \left[\log\frac{q}{p_\theta} \right]\,,
\end{align*}
so the right hand side of (\ref{eq: compute_jeffreys}) resumes the definition of Jeffreys divergence in (\ref{eq: jeffreys}).

For (\ref{eq: compute_grad_jeffreys}), we have, by definition
$$\frac{\partial}{\partial\theta}\Jef ( p_\theta \Vert q )=\frac{\partial}{\partial\theta}\E_{ p_\theta} \left[\log\frac{p_\theta}{q}\right]+\frac{\partial}{\partial\theta}\E_{ \tilde{q}} \left[\frac{q}{\tilde{q}}\log\frac{q}{p_\theta}\right]\,.$$
We compute
\begin{align*}
     &\frac{\partial}{\partial\theta}\E_{ p_\theta} \left[\log\frac{p_\theta}{q}\right]=\frac{\partial}{\partial\theta}\int p_\theta(x)\log\frac{p_\theta(x)}{q(x)}\rd x\\
     =&\int\left(\frac{\partial p_\theta(x)}{\partial\theta}\log\frac{p_\theta(x)}{q(x)}+p_\theta(x)\frac{\partial \log p_\theta(x)}{\partial\theta}\right)\rd x\,,
\end{align*}
and
\begin{align*}
     \frac{\partial}{\partial\theta}\E_{ \tilde{q}} \left[\frac{q}{\tilde{q}}\log\frac{q}{p_\theta}\right]=-\E_{\tilde{q}}\left[\frac{q}{\tilde{q}}\frac{\partial \log p_\theta}{\partial\theta}\right]\,.
\end{align*}
Now since $\frac{\partial}{\partial\theta}\log p_\theta(x)=\frac{1}{p_\theta(x)}\frac{\partial}{\partial\theta}p_\theta(x)$, we have
$$\frac{\partial}{\partial\theta}p_\theta(x)=p_\theta(x)\frac{\partial}{\partial\theta}\log p_\theta(x)\,.$$
So the term $\frac{\partial}{\partial\theta}\E_{ p_\theta} \left[\log\frac{p_\theta}{q}\right]$ further simplifies to
\begin{align*}
     \frac{\partial}{\partial\theta}\E_{ p_\theta} \left[\log\frac{p_\theta}{q}\right]=&\int\left(p_\theta(x)\frac{\partial\log p_\theta(x)}{\partial\theta}\log\frac{p_\theta(x)}{q(x)}\right.\\
     &+\left.p_\theta(x)\frac{\partial \log p_\theta(x)}{\partial\theta}\right)\rd x\\
     =&\E_{ p_\theta}\left[\left(1+\log\frac{p_\theta}{q}\right)\frac{\partial \log p_\theta}{\partial\theta}\right]\,.
\end{align*}
So we can conclude (\ref{eq: compute_grad_jeffreys}).

Now instead of the normalized density $q$, suppose we only have its unnomralized version $Zq$, with $Z$ unknown. When we replace $q$ with $Zq$ in (\ref{eq: compute_grad_jeffreys}), we get
\begingroup
\allowdisplaybreaks
\begin{align*}
    &\E_{p_\theta}\left[\left(1+\log\frac{p_\theta}{Zq}\right)\frac{\partial \log p_\theta}{\partial\theta}\right] \nonumber-\E_{\tilde{q}}\left[\frac{q}{\tilde{q}}\frac{\partial \log p_\theta}{\partial\theta}\right]\\
    =&\E_{p_\theta}\left[\left(1+\log\frac{p_\theta}{q}\right)\frac{\partial \log p_\theta}{\partial\theta}\right]-\E_{\tilde{q}}\left[\frac{q}{\tilde{q}}\frac{\partial \log p_\theta}{\partial\theta}\right]\\
    &-\log Z~\E_{p_\theta}\left[\frac{\partial \log p_\theta}{\partial\theta}\right]\\
    =&\frac{\partial}{\partial\theta}\Jef ( p_\theta \Vert q )-\log Z\int p_\theta(x)\frac{\partial \log p_\theta(x)}{\partial\theta}\rd x\\
    =&\frac{\partial}{\partial\theta}\Jef ( p_\theta \Vert q )-\log Z\int \frac{\partial p_\theta(x)}{\partial\theta}\rd x\\
    =&\frac{\partial}{\partial\theta}\Jef ( p_\theta \Vert q )-\log Z~\frac{\partial }{\partial\theta}\left(\int p_\theta(x)\rd x\right)\\
    =&\frac{\partial}{\partial\theta}\Jef ( p_\theta \Vert q )\,,
\end{align*}%
\endgroup
as $\int p_\theta(x)\rd x=1$. We remark that we don't have the importance weight term like $Zq/\tilde{q}$ in this case, because we can use the self-normalized importance weight. In practice, if we have $\tilde{x}_i$ sampled i.i.d. from $\tilde{q}$ for $i=1, \ldots, M$, the importance weight for $\tilde{x}_i$ is given by $w_i=\hat{w}_i/\sum_{j=1}^M\hat{w}_j$, where $\hat{w}_j=Zq(\tilde{x}_j)/\tilde{q}(\tilde{x}_j)$, for $j=1, \ldots, M$. We can see that the weight $w_i$ is independent from $Z$ as it cancels itself. The similar argument goes for (\ref{eq: compute_jeffreys}).

So we conclude that the Monte Carlo estimation of (\ref{eq: compute_jeffreys}) and (\ref{eq: compute_grad_jeffreys}) doesn't need to know the normalizing constant $Z$ in $q$ as defined in (\ref{eq: bayes}). $\square$

\section{The Recursive Multiscale Structure}
\label{ap: multiscale_property}

Here we detail the definitions and properties related to the multiscale structure. Recall the recursive design introduced in Section \ref{sec: network}, and set $L$ be the number of scales. At scale $l$ ($1\leq l\leq L$), the problem dimension is $d_l$, and $d_l$ increases with $l$: $d_1<d_2<\ldots<d_L=d$.

For $2\leq l\leq L$, the downsample operator $\A_l$ at scale $l$, introduced in Section \ref{sec: network}, is a linear operator from $\R^{d_{l}}$ to $\R^{d_{l-1}}$. It links the variable $x_l$ at scales $l$ to the variable $x_{l-1}$ at scales $l-1$ by $x_{l-1}=\A_lx_l$. Similarly, the upsample operator $\B_l$ at scale $l$, introduced in Section \ref{sec: theory}, is a linear operator from $\R^{d_{l-1}}$ to $\R^{d_{l}}$, for $2\leq l\leq L$.

The prior $\rho_l$ at scale $l$ is defined recursively: at the finest scale $l=L$, the prior $\rho_L=\rho$, and as for scale $l$ ($1\leq l< L$), $\rho_l$ is the density of $\A_{l+1}x_{l+1}$ if $x_{l+1}$ follows the last scale prior $\rho_{l+1}$. In other words, $\rho_l$ is the push-forward density of $\rho_{l+1}$ by $A_{l+1}$ for $1\leq l<L$.

To define the posterior $q_l$ at scale $l$, we first let $\hat{\B}_l=\B_L\B_{L-1}\ldots\B_{l+1}$ be the linear upsample operator from $\R^{d_l}$ to $\R^{d_L}=\R^d$, for $1\leq l<L$. It maps $x_l\in\R^{d_l}$ to a valid input in $\R^d$ for $\F$. For consistency, we define $\hat{\B}_L=I_{d_L}$, the identity map. Then we can introduce the likelihood $\L_l$ at scale $l$ as, for $1\leq l\leq L$,
$$\L_l(y|x_l):=\L(y|\hat{\B}_lx_l)=\N(y-\F(\hat{\B}_lx_l);0,\Gamma)\,.$$
Now we define the posterior $q_l$ at scale $l$, for $1\leq l\leq L$, as
$$q_l(x_l)=\frac{1}{Z_l}\rho_l(x_l)\L_l(y|x_l)\,,$$
where $Z_l$ is the normalizing constant.

\begin{figure}[!b]
    \centering
    \includegraphics[width=\columnwidth]{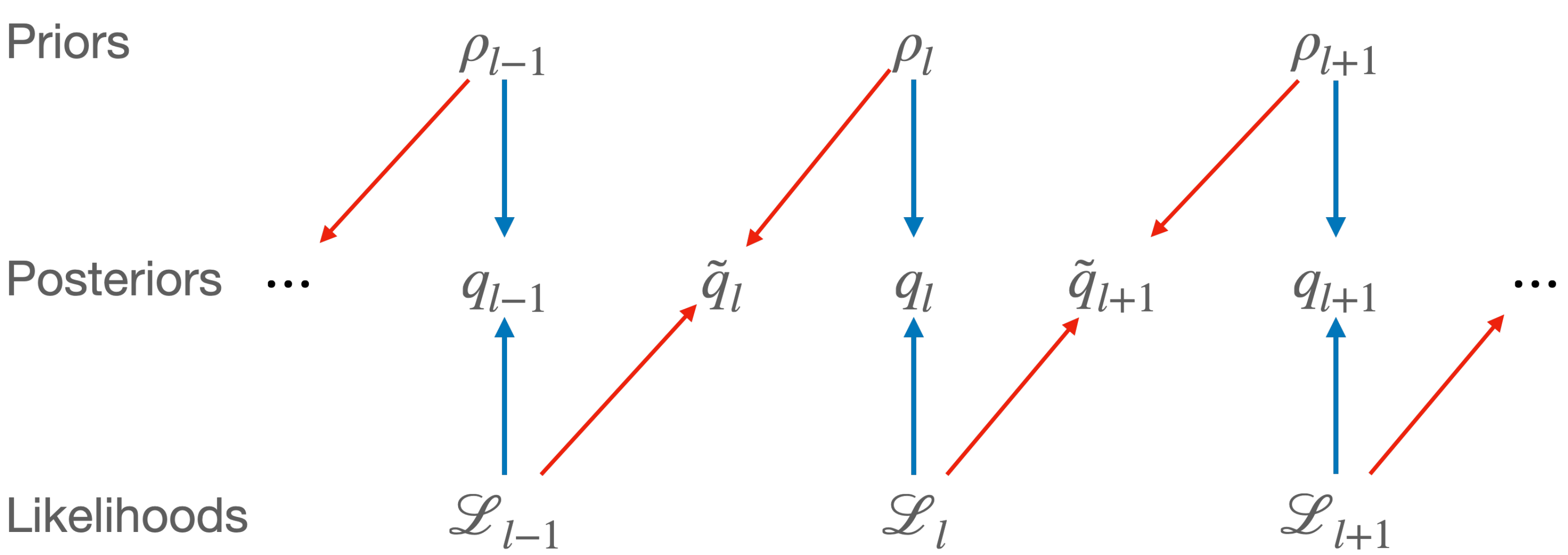}
    \caption{Conceptual diagram of the definitions. Arrows mean that ``contribute to the definition of''. We further remark that, $(i)$ $\tilde{q}_l$ is the upsampling of $q_{l-1}$ by $\rho_l(x_l|x_{l-1})$, because $\rho_l$ is the upsampling of $q_{l-1}$ by $\rho_l(x_l|x_{l-1})$, and $(ii)$ $q_l$ can be well approximated by $\tilde{q}_l$, because $\L_l(y|x_l)$ can be well approximated by $\L_{l-1}(y|\A x_l)$.}
    \label{fig: definitions}
\end{figure}

The auxiliary distribution $\tilde{q}_l$ at scale $l$, for $2\leq l\leq L$, introduced in Section \ref{sec: theory}, is defined as
$$\tilde{q}_l(x_l)=\frac{1}{\tilde{Z}_l}\rho_l(x_l)\L_{l-1}(y|\A_l x_l)\,,$$
where $\tilde{Z}_l$ is the normalizing constant. To see why $\tilde{q}_l$ approximates $q_l$ well, we notice that $\A_lx_l$ is a coarse-scale version of $x_l$, and by the multiscale property, $\F(\hat{\B}_lx_l)\approx\F(\hat{\B}_{l-1}\A_lx_l)$, so
$$\L_{l}(y|x_l)\approx\L_{l-1}(y|\A_l x_l)\,,$$
which implies that $q_l\approx\tilde{q_l}$.

We also notice that, the hierarchical definition of $\rho_l$ implies the following decoupling, for $x_{l-1}=\A_lx_l$,
$$\rho_l(x_l)=\rho_{l-1}(x_{l-1})\rho_l(x_l|x_{l-1})\,.$$
This decoupling is due to the conditional probability rule: 
$$\rho_{l}(x_l|x_{l-1})=\rho_{l}(x_l|\A x_l=x_{l-1})=\rho_l(x_l)/\rho_{l-1}(x_{l-1})\,.$$
Therefore, we arrive at an alternative formulation of $\tilde{q}_l$:
\begin{align*}
    \tilde{q}_l(x_l)&:=\frac{1}{\tilde{Z}_l}\rho_l(x_l)\L_{l-1}(y|\A_l x_l)\\
    &=\frac{1}{\tilde{Z}_l}\rho_{l-1}(\A x_l)\rho_l(x_l|\A_l x_l)\L_{l-1}(y|\A_l x_l)\\
    &=\frac{Z_{l-1}}{\tilde{Z}_l}\rho_l(x_l|\A x_{l})q_{l-1}(\A x_{l})\,,
\end{align*}
which suggests that a sample $x_l$ of $\tilde{q}_l$ can be generated in the following way:
$(i)$ sample $x_{l-1}$ from $q_{l-1}$, and $(ii)$ then sample $x_l$ from $\rho_l(x_l|x_{l-1})$. The relation of $\rho_l$, $\L_l$, $q_l$ and $\tilde{q}_l$ is shown in Figure \ref{fig: definitions}.

\section{More Discussion about Related Work}
\label{ap: related_work}

In this section we provide more discussion and comparison of our approach to related works.

In \citep{parno2016multiscale}, a similar notion of multiscale structure is developed as follows. A likelihood function has the {\it \citep{parno2016multiscale}-multiscale structure}, if there exists a coarse-scale {\it random variable} $\gamma$ of dimension $d_c$ ($d_c < d$) and a likelihood $\L_c$ such that 
\begin{align}\label{eq: multiscale_marzouk}
    \L(y|x, \gamma) = \L_c(y|\gamma)\,.
\end{align}
Then the {\it joint posterior distribution} of the fine- and coarse-scale parameters $(x, \gamma)$ can be decoupled as
\begin{align}
    q(x, \gamma) &\propto \rho(x, \gamma)\L(y|x, \gamma) \stackrel{(i)}{=} \rho(x, \gamma) \L_c(y|\gamma)\nonumber\\
    &\stackrel{(ii)}{=} \rho(x|\gamma) \rho(\gamma) \L_c(y|\gamma) \stackrel{(iii)}{=} \rho(x|\gamma) q_c(\gamma)\,,\label{eq: decouplescale_marzouk}
\end{align}
with normalizing constants omitted in the equivalence relations. We use the \citep{parno2016multiscale}-multiscale structure (\ref{eq: multiscale_marzouk}) in $(i)$, and the conditional probability rule $\rho(x, \gamma)=\rho(x|\gamma)\rho(\gamma)$ in $(ii)$. In $(iii)$ we define $q_c(\gamma):=\rho(\gamma)\L_c(y|\gamma)$ as the \citep{parno2016multiscale}-posterior in coarse scale.

There are two important differences in these two definitions. First, our coarse-scale parameter $x_c$ is a deterministic function of the fine-scale parameter $x$, while in \citep{parno2016multiscale}, $\gamma$ is a random variable that may contain extra randomness outside $x$ (as demonstrated in numerical examples in \citep{parno2016multiscale}). This difference in definition results in significant difference in modeling: our invertible model has $d$-dimensional random noise $z$ as input to approximate the target posterior $q(x)$, while models in \citep{parno2016multiscale} has $(d+d_{c})$-dimensional random noise as input to approximate the joint-posterior $q(x, \gamma)$. Another consequence is that users need to define the joint prior $\rho(x, \gamma)$ in \citep{parno2016multiscale}, while in our definition the prior of $x_c$ is naturally induced by the prior of $x$.

Second, our multiscale structure (\ref{eq: key_obs}) is an approximate relation and we use invertible flow $F$ in MsIGN to model this approximation, while in \cite{parno2016multiscale} the multiscale structure (\ref{eq: decouplescale_marzouk}) is an exact relation and authors treat the prior-upsampled solution $\rho(x|\gamma) q_c(\gamma)$ (right hand side of (\ref{eq: decouplescale_marzouk})) as the final solution. Our approximate multiscale relation and further treatment by transform $F$ enables us to apply the method recursively in a multiscale fashion, while in \citep{parno2016multiscale} the proposed method is essentially a two-scale method and there is not further correction based on the prior-upsampled solution $\rho(x| \gamma) q_c( \gamma)$ at the fine-scale.

Finally, as we discussed in Section \ref{sec: related}, the invertible model in \citep{parno2016multiscale} is polynomials, which suffer from the exponential growth of polynomial coefficients as dimension grows. In this work, the invertible model is deep generative networks, whose number of parameters are independent of the problem dimension.

We also observe that \citep{spantini2015optimal,chen2019projected,chen2020projected} seeks a best low-rank approximation of the posterior, and treat the approximation as the final solution with no extra modification. As we will see in Appendix \ref{ap: exp_BIP}, the true posterior could still be far away from the prior-upsampled solution, especially in the first few coarse scales.

In addition, while in \citep{ardizzone2018analyzing} flow-based generative models are also used to in distribution capture in inverse problems, their definition of posterior is not equivalent to ours, as they assume no error in measurement. Furthermore, as their training strategy looks to capture the target distribution while simultaneously learning the forward map $\F$, they mainly focused on low-$d$ Bayesian inference problems, in contrast with our high-$d$ setting here.

\section{Experimental Setting and Additional Results for BIPs in Section \ref{sec: exp_BIP}}
\label{ap: exp_BIP}

\subsection{Experimental Setting of BIPs}
\label{ap: exp_setting_BIP}

As introduced in Section \ref{sec: exp_BIP}, we don't distinguish between the vector representation of $x$ as grid values on the 2-D $64\times 64$ uniform lattice: $x\in\R^d$ with $d=64*64=4096$, and the piece-wise constant function representation of $x$ on the unit disk: $x(s)$ for $s\in\Omega=\left[0,1\right]^2$.

We place a Gaussian distribution $\N(0, \Sigma)$ with the covariance $\Sigma$ as the discretization of $\beta^2(-\Delta)^{-1-\alpha}$ for both of our Bayesian inverse problem examples. Here the discretization of the Laplacian operator $\Delta$ can be understood as a graph Laplacian when we consider $x$ gives grid values on a 2-D uniform lattice. We choose zero Dirichlet boundary condition for $\Delta$. As for the distribution to model noise (error) as in (\ref{eq: likelihood}), we set $\Gamma=\gamma^2I$, where $I$ is the identity matrix. We list the setting of $(\alpha,\beta,\gamma)$ for both BIPs in Table \ref{tab: BIP_param}.

\begin{table}[tbp]
\caption{Hyper-parameter $(\alpha,\beta,\gamma)$ setting in BIPs}
\label{tab: BIP_param}
\begin{center}
\begin{small}
\begin{sc}
\begin{tabular}{l|ccc}
\toprule
Problem Name & $\alpha$ & $\beta$ & $\gamma$ \\
\midrule
Synthetic (Section \ref{sec: exp_syn_BIP}) & 0.1 & 2.0 & 0.2 \\
Elliptic (Section \ref{sec: exp_elp_BIP}) & 0.5 & 2.0 & 0.02 \\
\bottomrule
\end{tabular}
\end{sc}
\end{small}
\end{center}
\end{table}

The synthetic BIP sets its ground-truth for $x$ as $x(s)=\sin(\pi s_1)\sin(2\pi s_2)$, and defines its forward map as a nonlinear measurement of $x$:
$$\F(x)=\langle\varphi, x\rangle^2=\left(\int_\Omega \varphi(s)x(s)\rd s\right)^2\,,$$
where $\varphi(s)=\sin(\pi s_1)\sin(2\pi s_2)$.

The elliptic BIP is a benchmark problem for high-$d$ inference from geophysics and fluid dynamics \citep{iglesias2014well, cui2016dimension}. It also sets its ground-truth for $x$ as $x(s)=\sin(\pi s_1)\sin(2\pi s_2)$. However, the forward map is defined as $\F(x)=\mathcal{O}\circ\mathcal{S}(x)$, where $u=\mathcal{S}(x)$ is the solution to an elliptic partial differential equation with zero Dirichlet boundary condition:
$$-\nabla\cdot\left(e^{x(s)}\nabla u(s)\right)=f(s)\,,\quad s\in\Omega\,,$$
And $\mathcal{O}$ is linear measurements of the field function $u$:
$$\mathcal{O}(u)=\begin{bmatrix} \int_\Omega\varphi_1(s)u(s)\mathrm{d}s & \ldots & \int_\Omega\varphi_m(s)u(s)\mathrm{d}s \end{bmatrix}^T\,.$$
The force term $f$ of the elliptic PDE is set as
\begin{align*}
    f(s)=&\frac{50}{\pi}\left(2e^{-10\|s-f_1\|^2}+2e^{-10\|s-f_2\|^2}\right.\\
    &\left.-e^{-10\|s-f_3\|^2}-e^{-10\|s-f_4\|^2}\right)\,,
\end{align*}
where $f_1=(0.25, 0.3)$, $f_2=(0.25, 0.7)$, $f_3=(0.7, 0.3)$, $f_4=(0.7, 0.3)$, and $\|\cdot\|$ is the Euclidean norm in $\R^2$. $f$ is mirror-symmetric along the $s_2$ direction: $f(s_1, s_2)=f(s_1, 1-s_2)$. As for the measurement functions $\varphi_k$ ($1\leq k\leq m$), we set $m=15$ and each $\varphi_k$ gives local detection of $u$. They are also mirror-symmetry along the $s_2$ direction. See Figure \ref{fig: elliptic_measurement_force} for the visualization of $f$ and $\varphi_k$.

\begin{figure}[tbp]
    \centering
    \includegraphics[width=0.41\columnwidth]{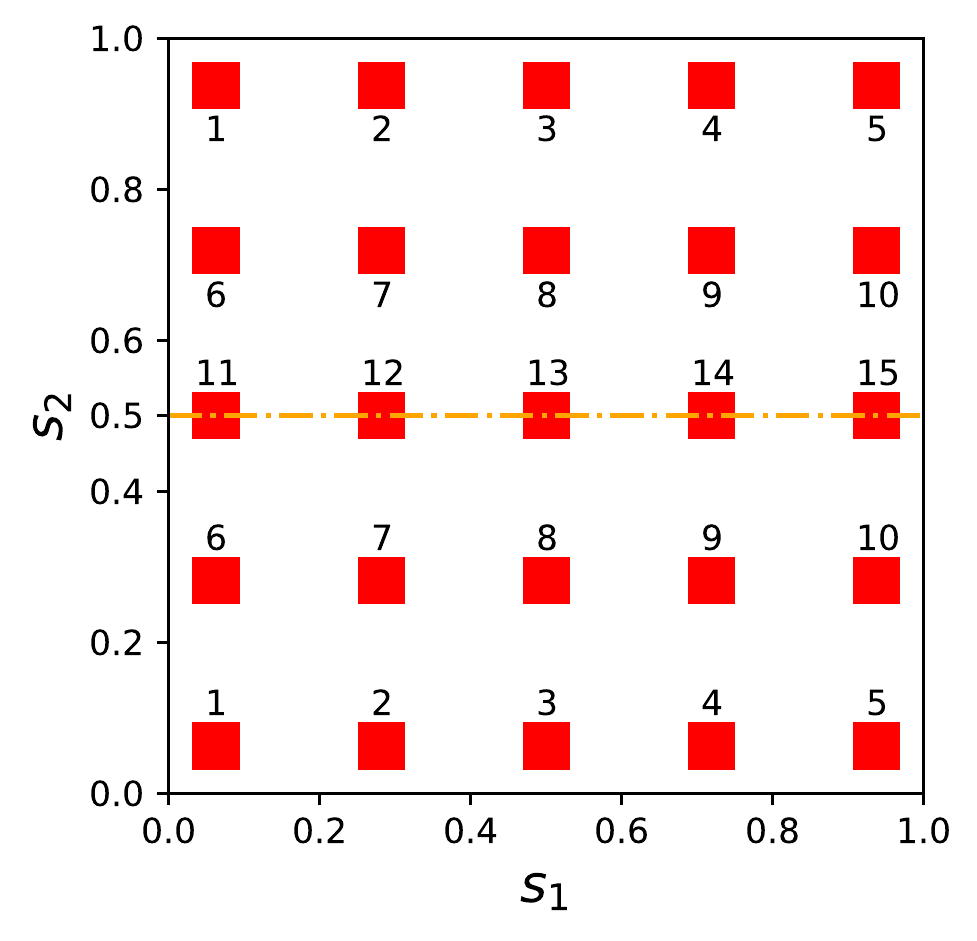}
    \includegraphics[width=0.48\columnwidth]{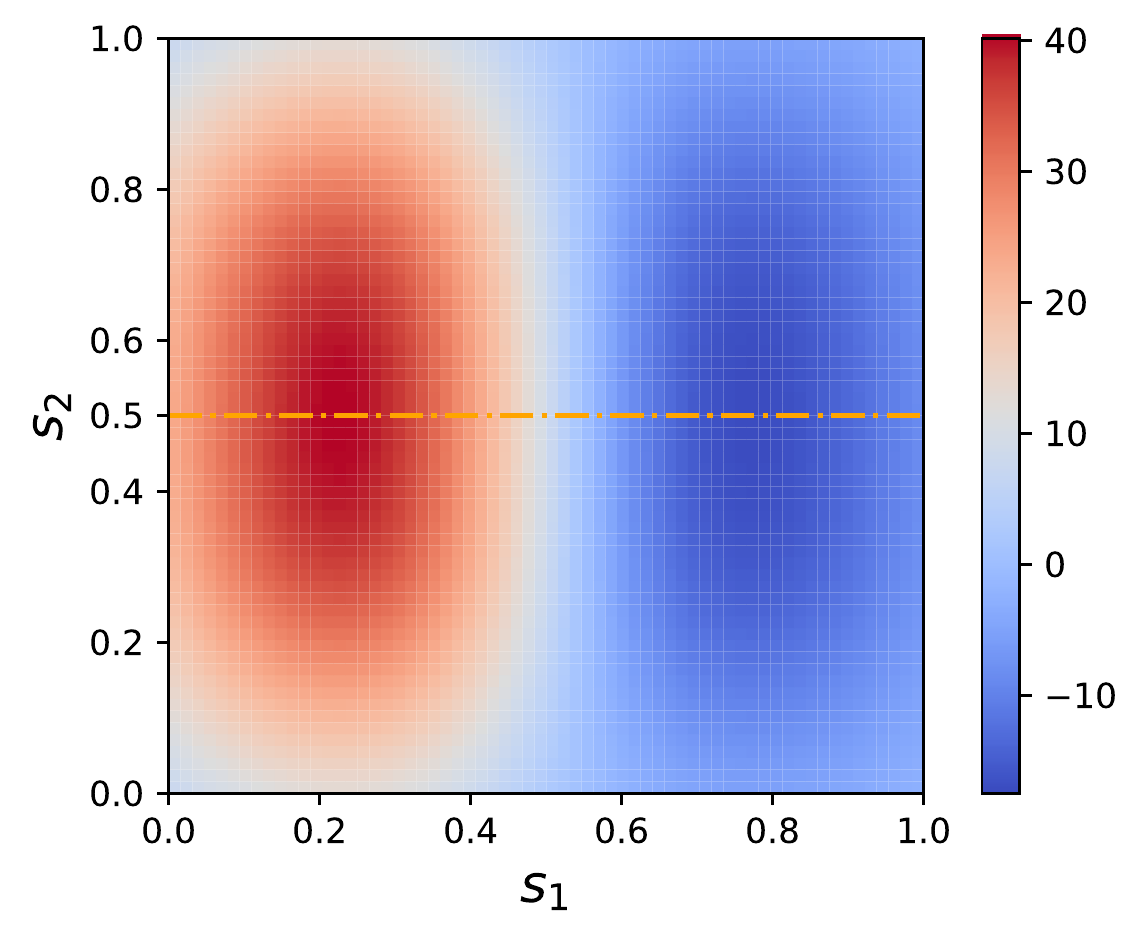}
    \caption{Left: 15 measurement functions in $\mathcal{O}$. Here we plotted non-zero patches of $\varphi_k$, $k=1, \ldots, 15$, with $k$ labeled next to them. $\varphi_k$ has a constant non-zero value on its patch(es) and is zero else where. The constant value here is chosen so that we have $\|\varphi_k\|_{L^2}=1$. Right: The force term $f$ of the elliptic PDE in $\mathcal{S}$. We remark that both measurement functions and the force term are mirror-symmetric along the $s_2$ direction (the orange dash line).}
    \label{fig: elliptic_measurement_force}
\end{figure}

By the symmetric design, our posterior $q$ has the property $q(x)=q(x^\prime)$, where, when considered in the representation of function, $x$ and $x^\prime$ is linked by $x(s_1, s_2)=x^\prime(s_1, 1-s_2)$ for $s=(s_1, s_2)\in\Omega$. We carefully choose our hyper-parameters $(\alpha, \beta, \gamma)$, as in Table \ref{tab: BIP_param}, such that it ends up to be q not only mirror-symmetric, but also double-modal posterior distribution. To certify the multi-modality, we run multiple gradient ascent searching of maximum-a-posterior points, starting from different initial points. They all converge to two mutually mirror-symmetric points $x^*$ and $x^{*\prime}$: for $s=(s_1, s_2)\in\Omega$, $x^*(s_1, s_2)=x^{*\prime}(s_1, 1-s_2)$. Visualization of the 1D landscape profile of the posterior $q$ on the line passing through $x^*$ and $x^{\prime*}$ also shows a clear double-modal feature.

To simulate the forward process $\F$, we solve the PDE in map $\mathcal{S}$ by the Finite Element Method with mesh size $1/64$. We remark here this setting is independent of the scale $l$ ($1\leq l\leq L$) in our recursive strategy.

When counting the number of forward simulations (nFSs) as our indicator for computational cost, we notice that all SVGD-type methods: A-SVGD, SVGD and pSVGD, require not only the log posterior $\log q(x)$ but also its gradient: $\partial_x\log q(x)$. Thanks to the adjoint method, the gradient can be computed with only one extra forward simulation.

In Table \ref{tab: BIP_train} we report our hyperparameter of network setting in BIPs. To initialize our multi-stage training as in line 2 of Algorithm \ref{alg: training}, we still try to minimize the Jeffreys divergence $\Jef(p_\theta\Vert q)=\E_{p_\theta}[\log(p_\theta/q)]+\E_{q}[\log(q/p_\theta)]$ , but this time it is directly estimated by the Monte Carlo method with samples from distribution $p_\theta$ and $q$. $p_\theta$ samples come from the model itself and $q$ samples come from an HMC chain. We remark that at $l=1$, the posterior lies in 4-D space, which is relatively a low-$d$ problem, so an HMC run can approximate the target distribution $q_1$ well. Our present solution of seeking help from HMC can be replaced by some other strategies, like other MCMC methods, and deep generative networks.

For A-SVGD, we choose Glow \citep{kingma2018glow} as its network design, with the same network hyperparameter in Table \ref{tab: BIP_train}. Due to the fact that MsIGN is more parameter-saving than Glow with the same hyperparameter, A-SVGD model has more trainable parameters than our MsIGN model, reducing the possibility that that its network is not expressive enough to capture the modes.

As for our training of HMC, we grid search its hyperparameters, and use curves of acceptance rate and autocorrelation as evidence of mixing. We consider our HMC chain mixing successfully if the acceptance rate stabilizes and falls between $30\%-75\%$, as suggested by \citep{neal2011mcmc}, and the autocorrelation decays fast with respect to lag.

\begin{table}[tbp]
\caption{Hyperparameter setting for MsIGN in Section \ref{sec: exp_BIP}. The meaning of terms can be found in \citep{kingma2018glow}.}
\label{tab: BIP_train}
\begin{center}
\begin{small}
\begin{sc}
\begin{tabular}{l|cc}
\toprule
Problem Name & Synthetic & Elliptic \\
\midrule
Minibatch Size & 100 & 100 \\
Scales (L) & 6 & 6 \\
$\sharp$ of Glow Blocks (K) & 16 & 32 \\
$\sharp$ of Hidden Channels & 32 & 64 \\
\bottomrule
\end{tabular}
\end{sc}
\end{small}
\end{center}
\end{table}

As for the ablation study shown in Section \ref{sec: exp_BIP_ablation}, all models involved Glow or MsIGN adopt network hyperparameters as shown in Table \ref{tab: BIP_train}.We remark that it is not straightforward to design multi-stage strategy for Glow models, because their channel size increases with $l$. So for models with different number of scales $L$, there is no direct way to initialize one model with another. Therefore for methods using Glow, we don't consider multi-stage training.

Also, as will be seen in Appendix \ref{ap: exp_result_elp_BIP}, the elliptic problem at $l=1$ is ill-posed, its posterior is highly rough, and MsIGN variants (like MsIGN trained by the KL divergence) can hardly capture its two modes, see Table \ref{tab: ablation_result} and Figure \ref{fig: ablation_more}. We report that in general it is unlikely for multi-stage training to pick up the missing mode. Therefore, to make more convincing comparison, for models with multi-stage training, we use pretrained MsIGN model at $l=1$ (who captures $q_1$ well) as their initialization for $l=2$.

\subsection{Additional Results of BIPs}
\label{ap: exp_result_BIP}

In this section we provide more results on the Bayesian inverse problems examples in Section \ref{sec: exp_BIP}.

\subsubsection{Synthetic Bayesian Inverse Problem}
\label{ap: exp_result_syn_BIP}

In Figure \ref{fig: toy_elliptic_marginal} we provide comparison of the marginal distribution in the critical direction $w^*$ at intermediate scales $l=1, \ldots, 5$. For the final scale $l=6$ please refer to Figure \ref{fig: toy_BIP}(a). We can see that as the dimension increases, A-SVGD and SVGD become less robust in mode capture and collapse to one mode. Besides, HMC becomes imbalanced between modes, and pSVGD is a bit biased for $q_6$ in Figure \ref{fig: toy_BIP}(a). We remark here that in $q_1$, A-SVGD failed to capture both modes as it did to $q_2$. This phenomenon might be caused by the aliasing effect. Very rough resolution at this scale pushes the prior to penalize the smoothness much, and also adds the sensitivity to likelihood because entries of $x$ can easily influence its global behavior. Therefore, there is a larger log density gap between modes in the posterior $q_1$ than other scales, which adds up to the difficulty of multi-mode capture. A similar effect is observed in the elliptic example as in the next section.

The learning curve in Figure \ref{fig: training_necessity} shows the effectiveness of our multi-stage training of MsIGN. As we can see, the training process at $l=6$ did improve the model, with the Jeffreys divergence dropped from $252$ to $56.8$. Rather than simply refining the resolution, our multi-stage training strategy does improve our approximation to the distribution when entering the next scale. We will show more evidence about this in the next section.

\begin{figure*}[tbp]
    \centering
    \includegraphics[width=0.36\textwidth]{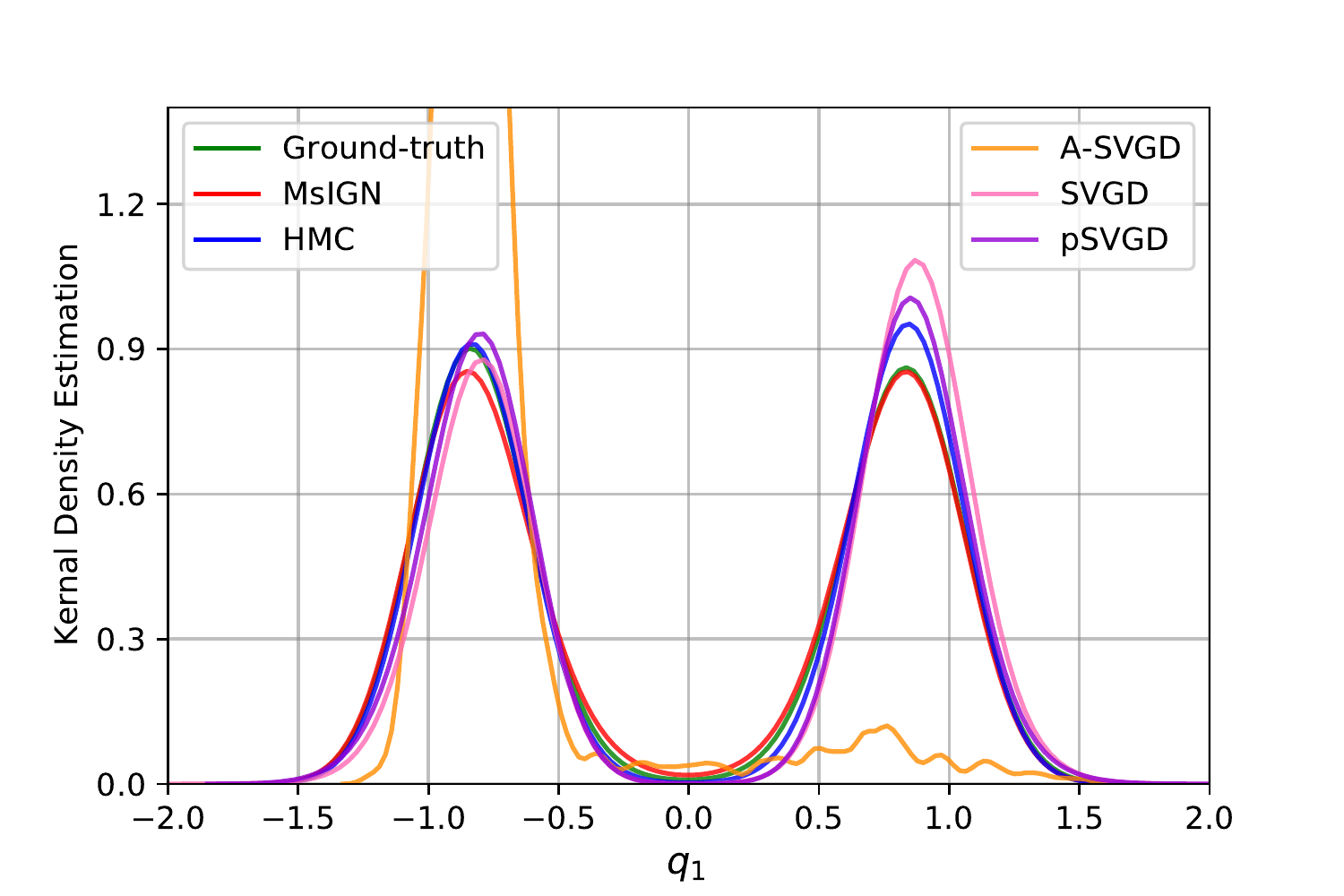}\includegraphics[width=0.36\textwidth]{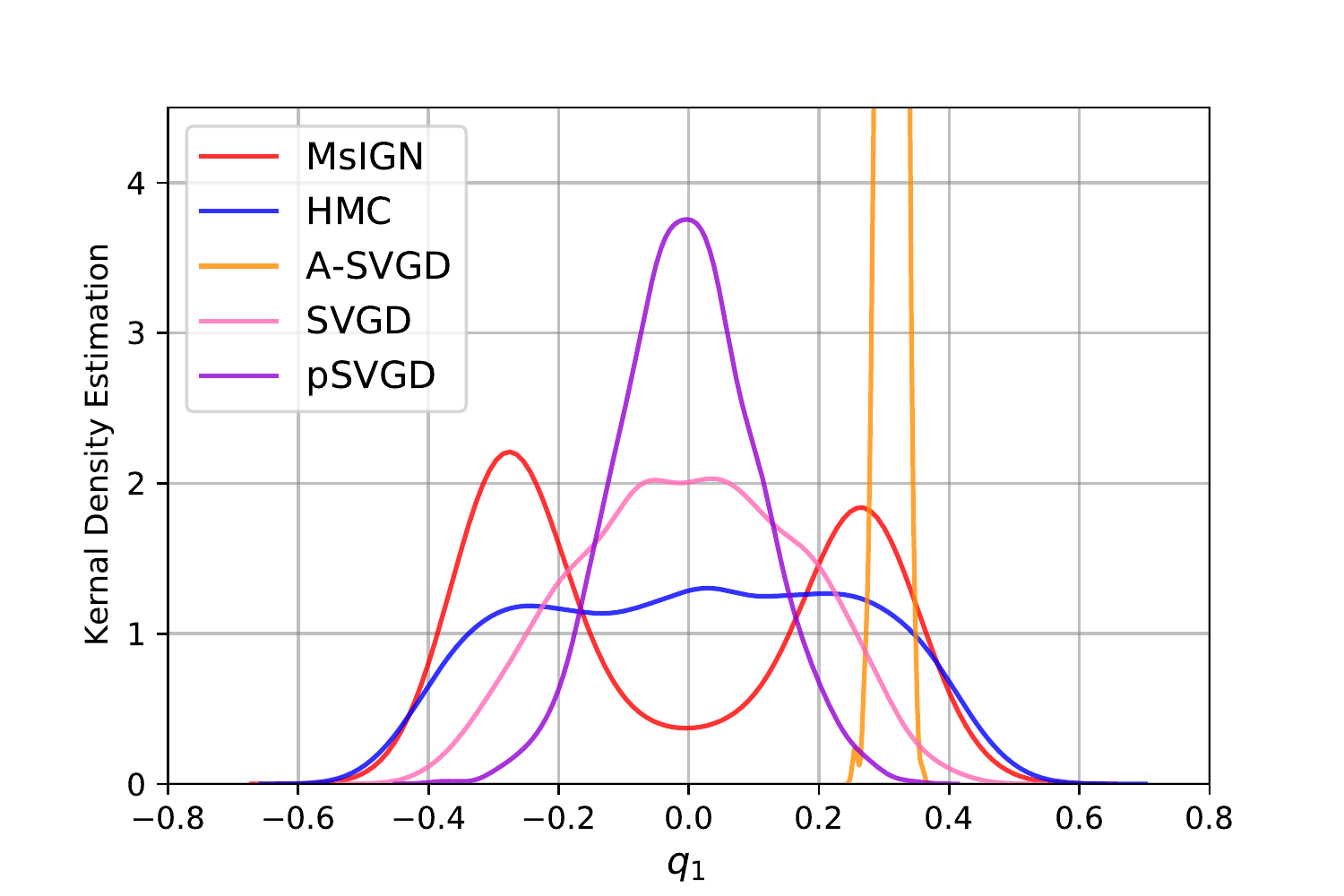}\\
    \includegraphics[width=0.36\textwidth]{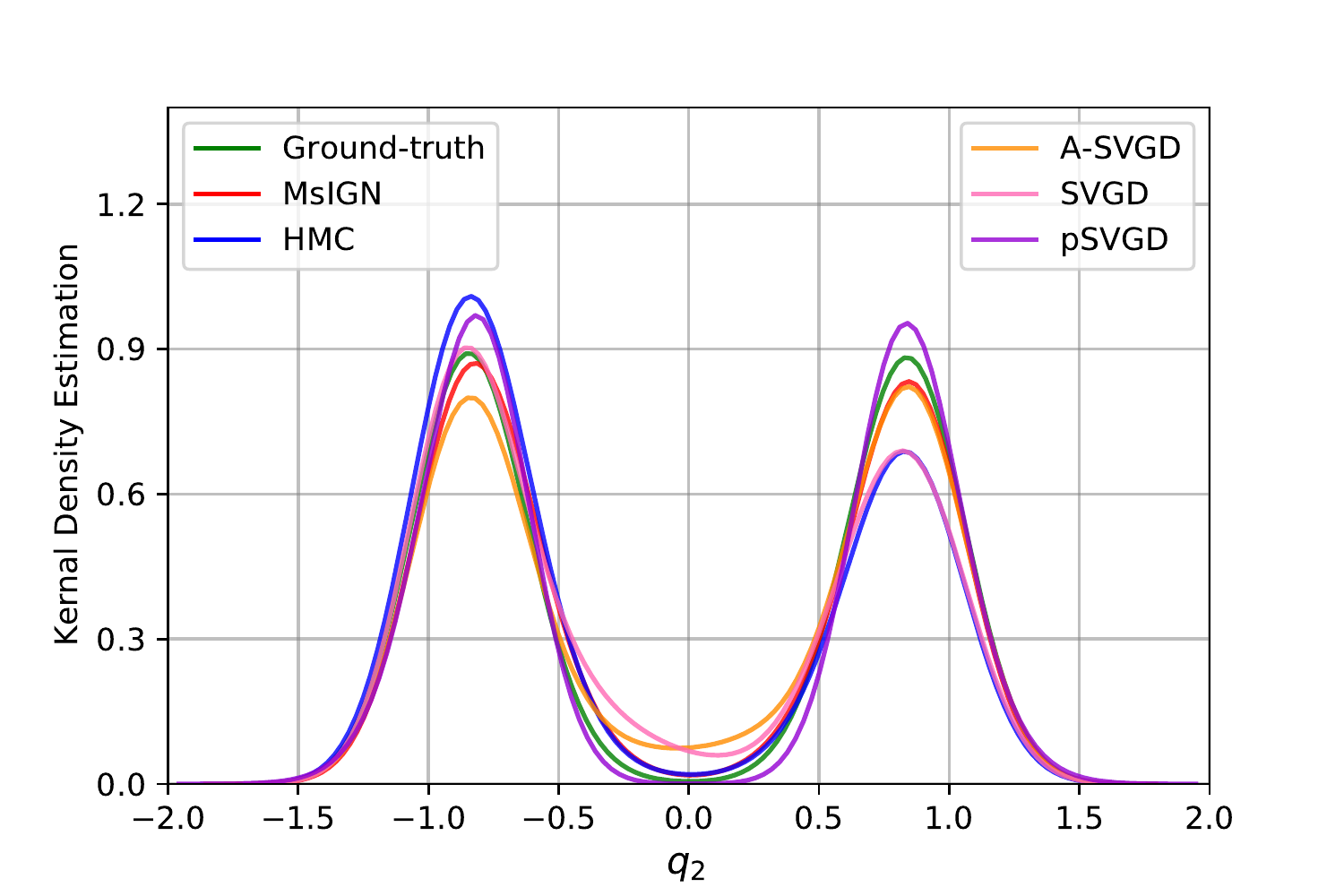}\includegraphics[width=0.36\textwidth]{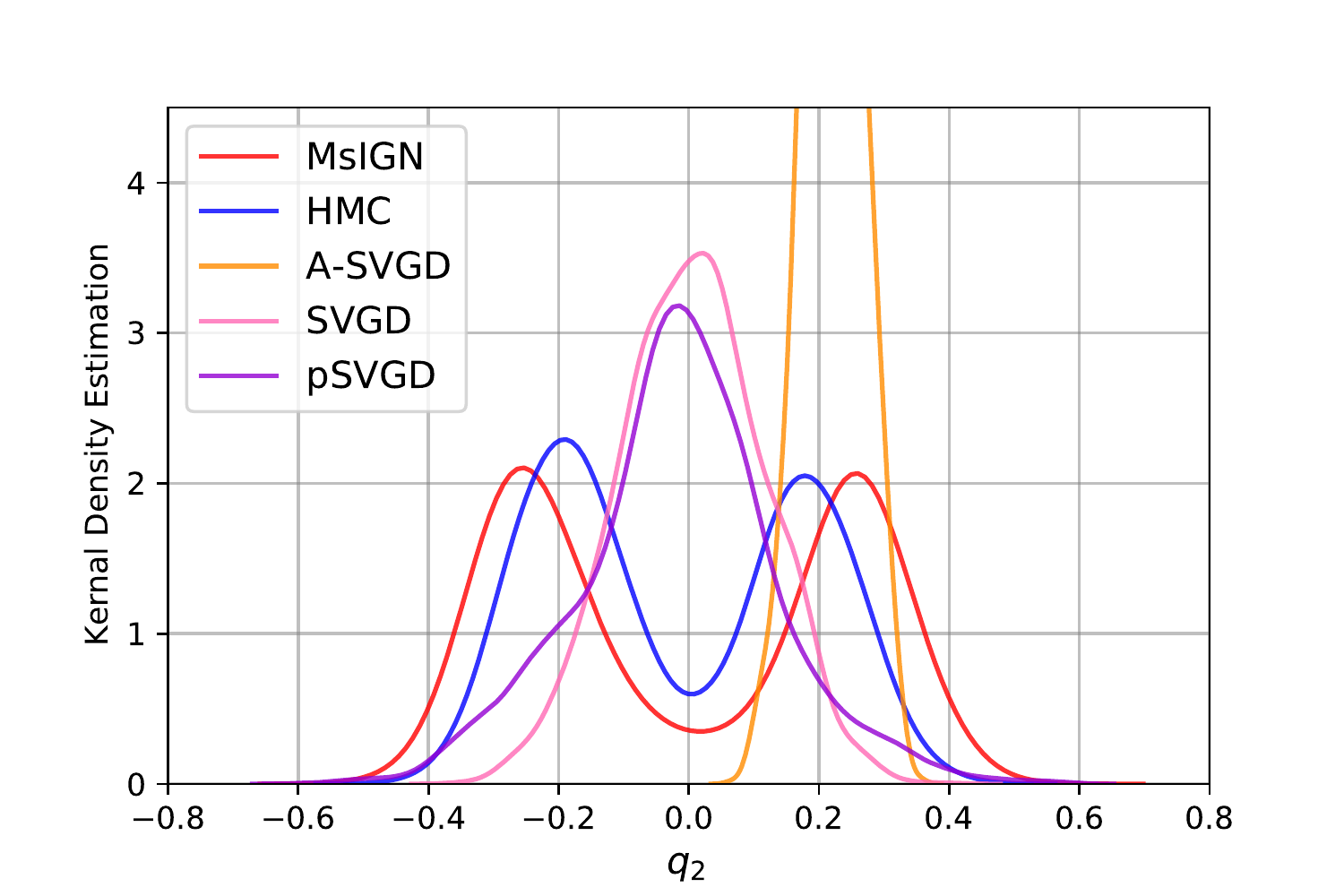}\\
    \includegraphics[width=0.36\textwidth]{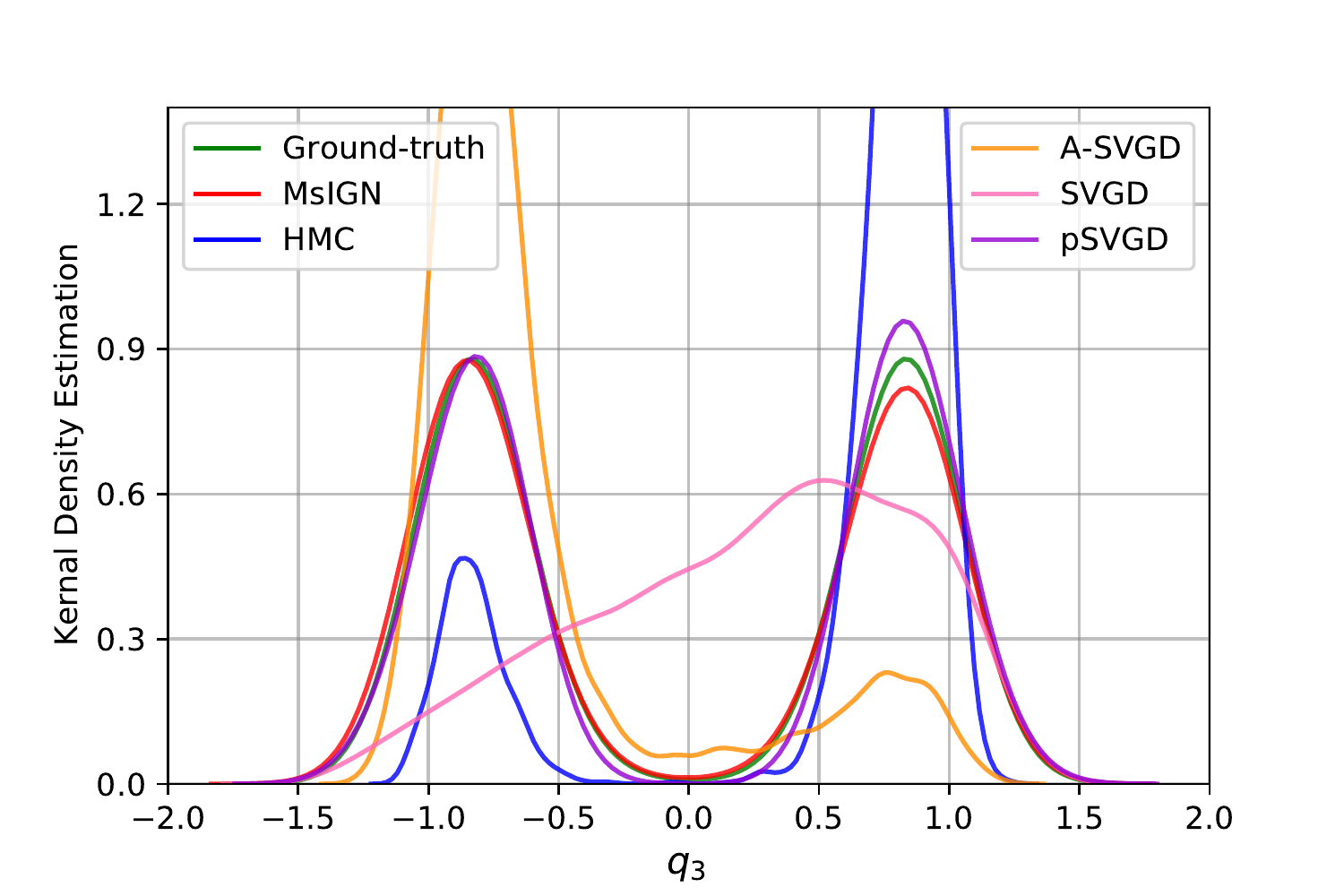}\includegraphics[width=0.36\textwidth]{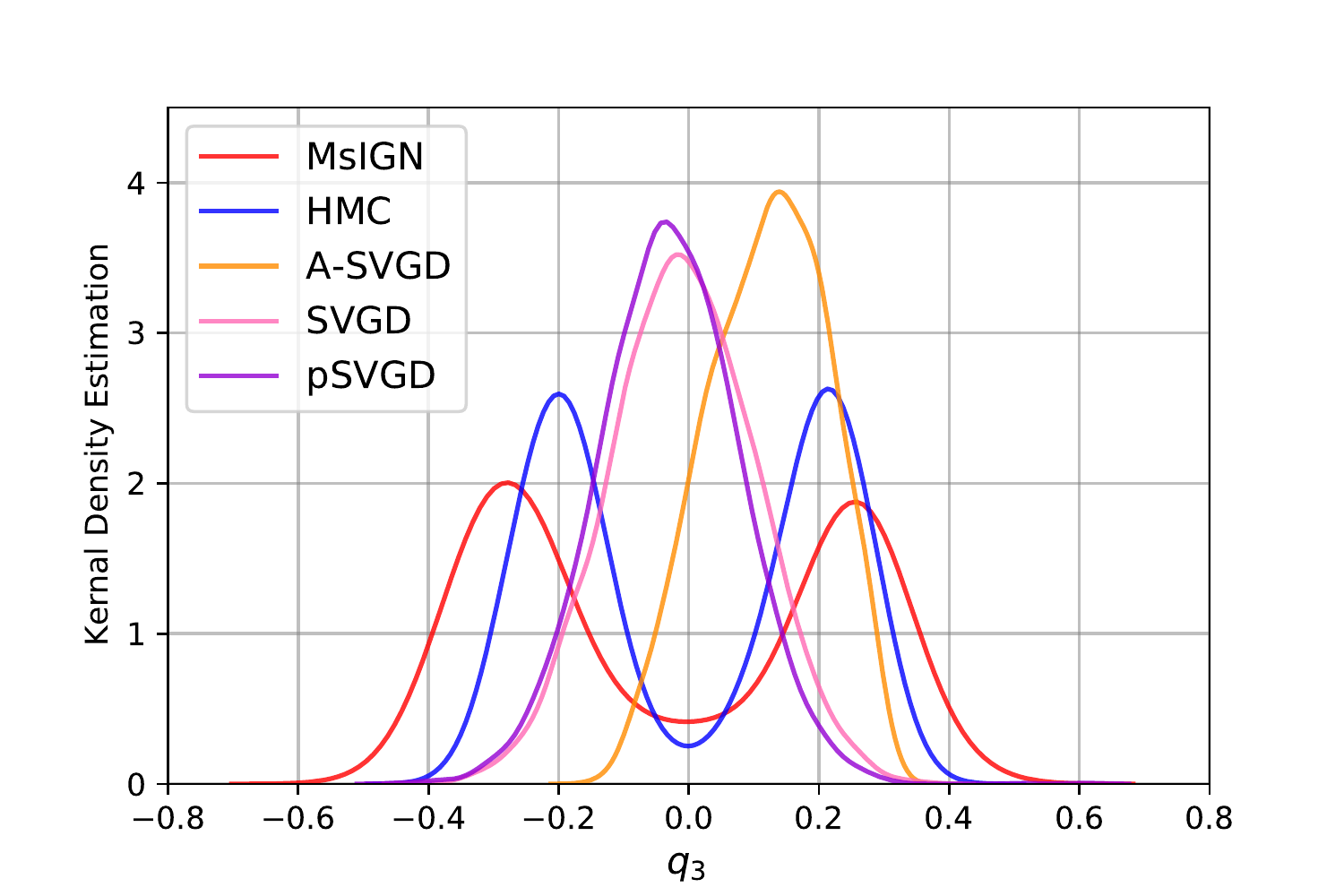}\\
    \includegraphics[width=0.36\textwidth]{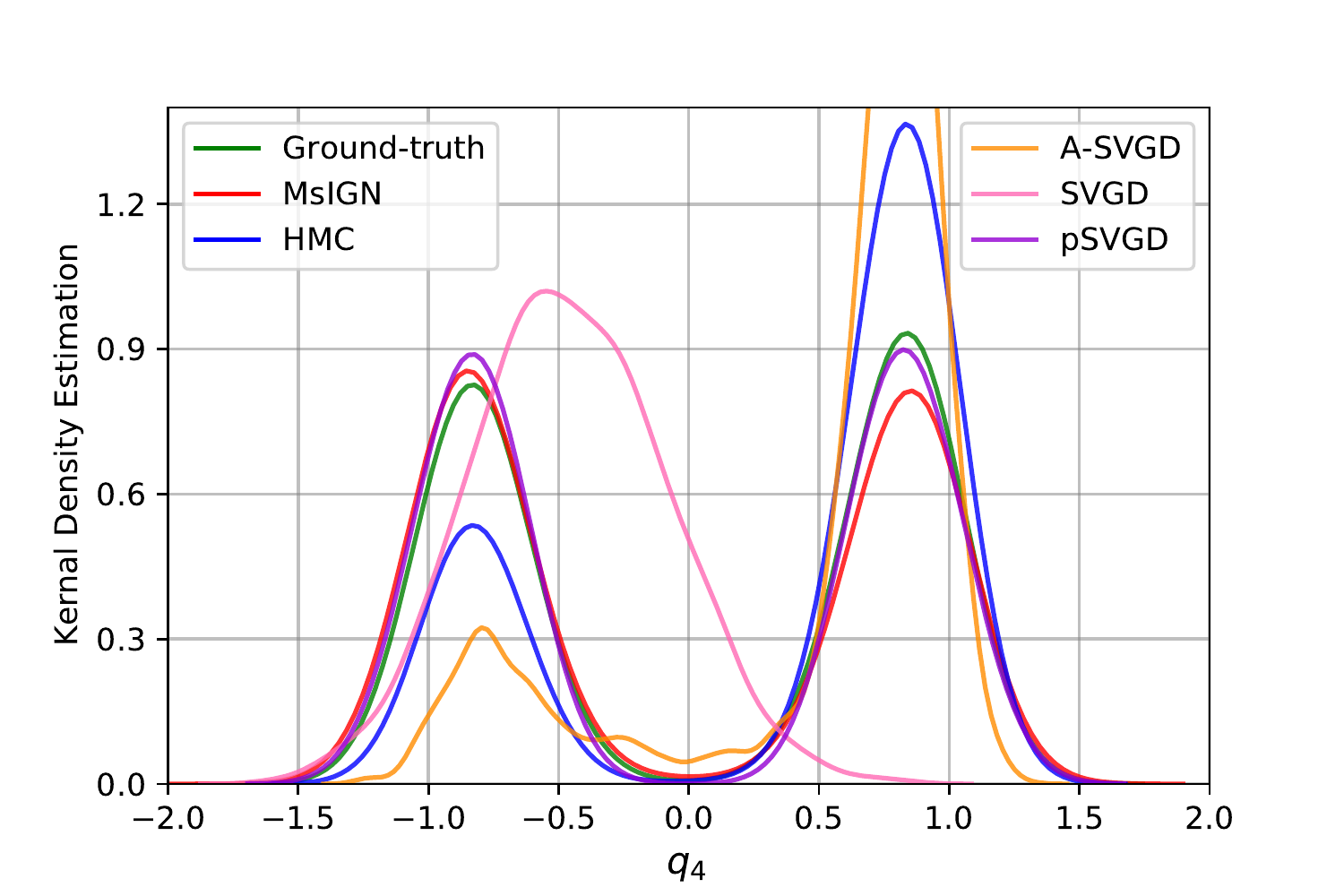}\includegraphics[width=0.36\textwidth]{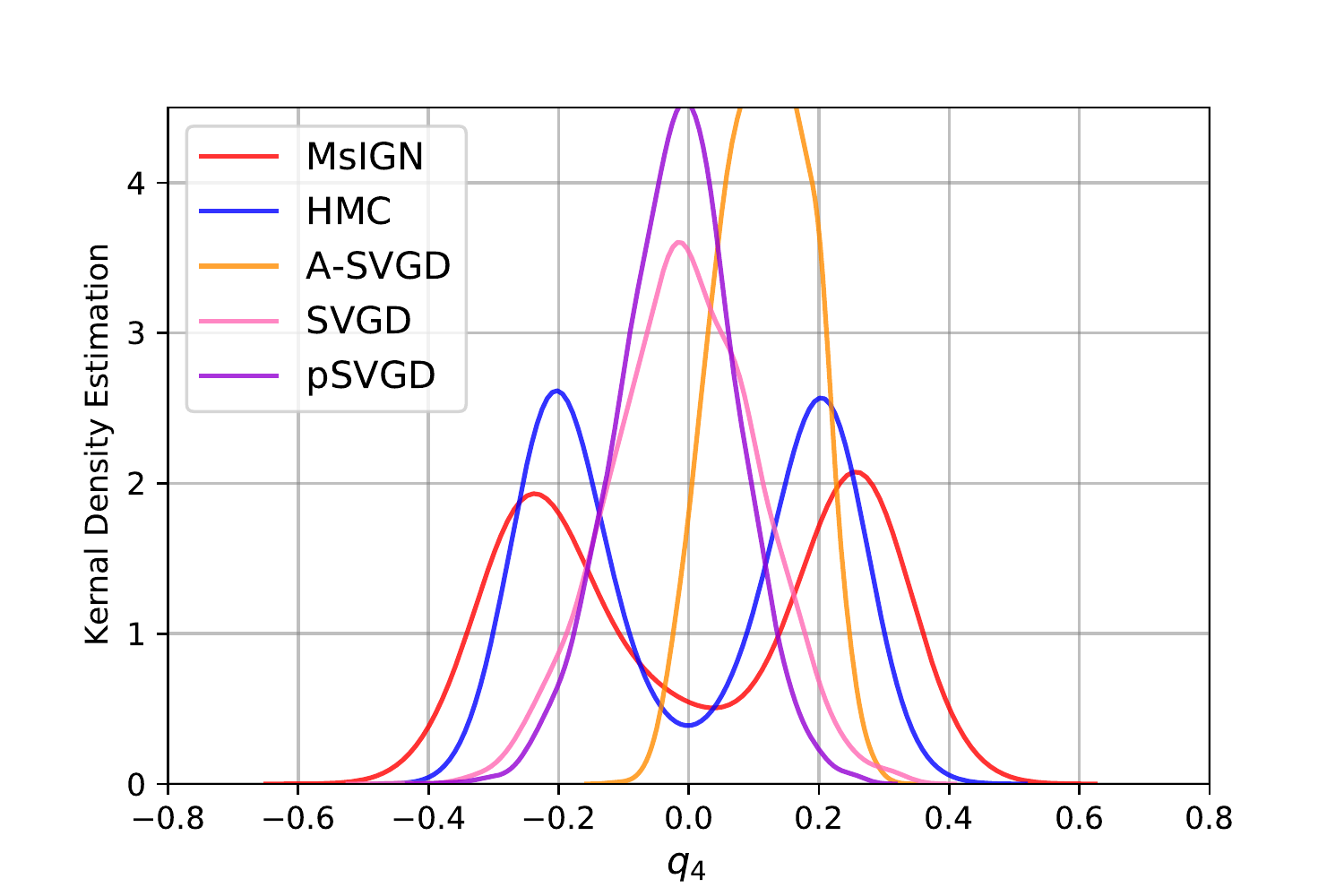}\\
    \includegraphics[width=0.36\textwidth]{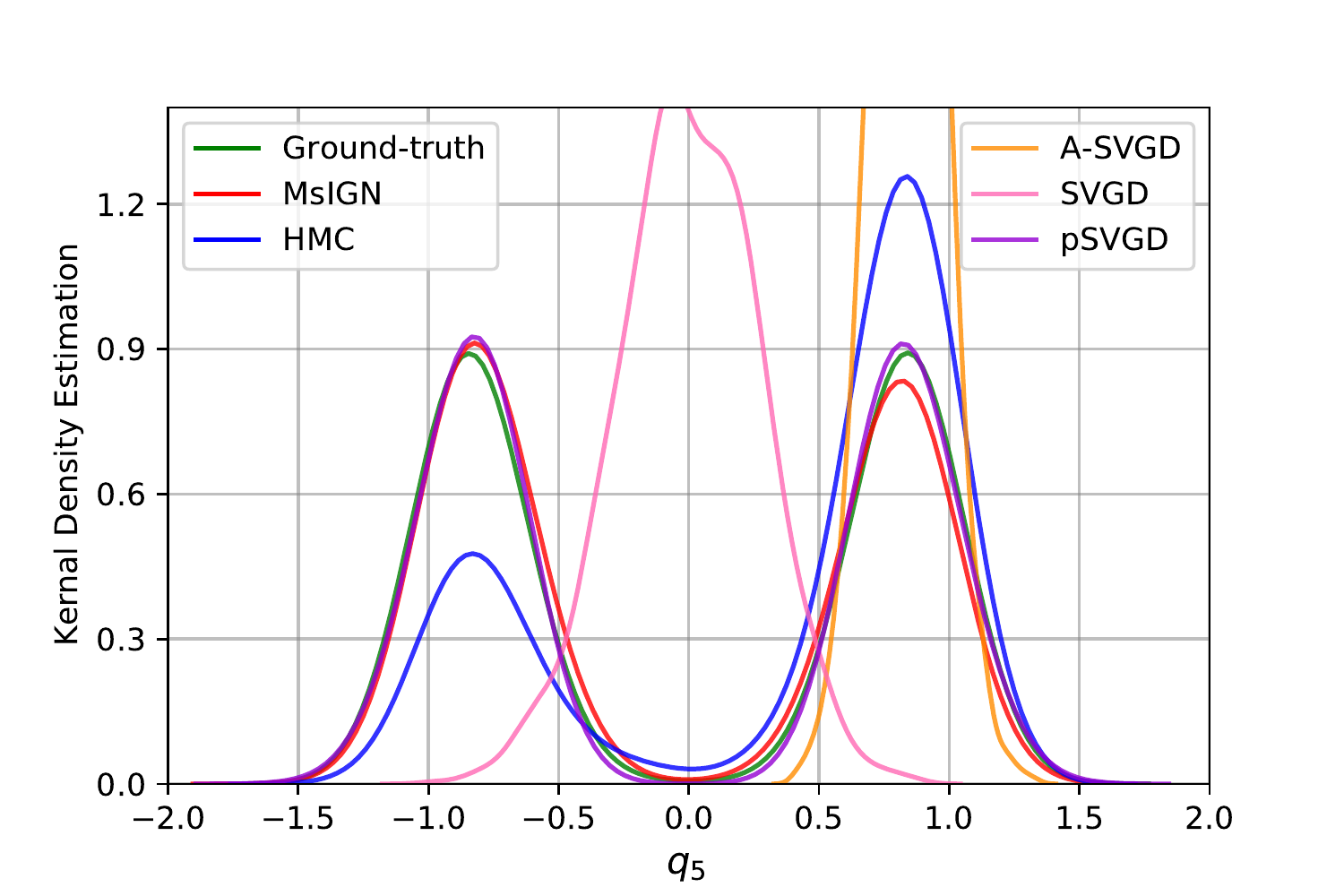}\includegraphics[width=0.36\textwidth]{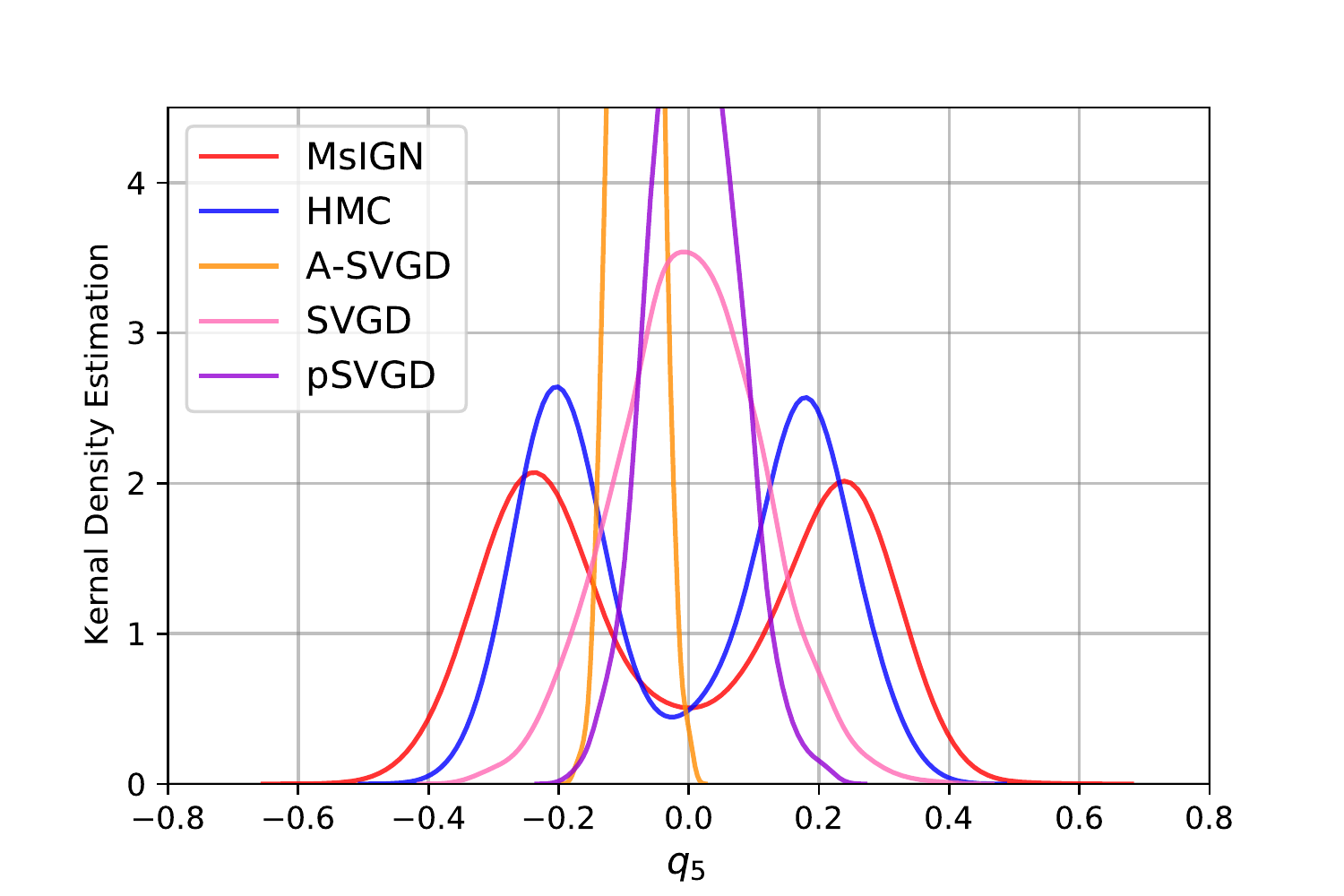}\\
    \caption{Marginal comparison at the intermediate scales $l=1, \ldots, 5$. Left: Synthetic BIP; Right: Elliptic BIP. In the synthetic example, as the dimension increases, SVGD and A-SVGD failed in mode capture. Besides, HMC becomes imbalanced between modes, and pSVGD is a bit biased for $q_6$ in Figure \ref{fig: toy_BIP}(a). In the elliptic example, all methods except MsIGN and HMC failed in detecting all modes, and could even get stuck in the middle. HMC has acceptable performance, but still suffers from imbalanced modes at some scales.}
    \label{fig: toy_elliptic_marginal}
\end{figure*}

\subsubsection{Elliptic Bayesian Inverse Problem}
\label{ap: exp_result_elp_BIP}

In Figure \ref{fig: toy_elliptic_marginal} we provide comparison of marginal comparison in the critical direction $w^*$ at intermediate scales $l=1, \ldots, 5$. For $l=6$ please refer to Figure \ref{fig: elliptic_BIP}(a). Again, for this complicated posterior we observe that all methods except MsIGN and HMC failed in detecting all modes, and could even get stuck in the middle. In this testbed, HMC seems to capture both modes well. However we will point out that its samples can't be treated like a reference solution. The failure of HMC at $q_1$ is due to the aliasing effect: the prior penalizes fluctuation in spatial directions heavily, and the likelihood is also very strong. As a consequence, the posterior $q_1$ is highly twisted, and the log density gap between two modes becomes significant.

In Figure \ref{fig: training_necessity}, we also show the necessity of training after prior conditioning. In other words, $q_l$ is not the same as the prior-conditioned surrogate $\tilde{q}_{l-1}$, though they are similar. We plot one of the modes we detected by our models for $l=4,5,6$. Comparing figures of Figure \ref{fig: training_necessity}, we can see the location, shape and scale of bumps and caves are different, which means the learned $q_l$ is different from the prior-conditioned surrogate $\tilde{q}_{l-1}$, who serves as its initialization. Our multi-stage training does learn more information at each scale, rather than simply scale up the resolution.

\begin{figure*}[tbp]
    \centering
    \includegraphics[width=0.4\textwidth]{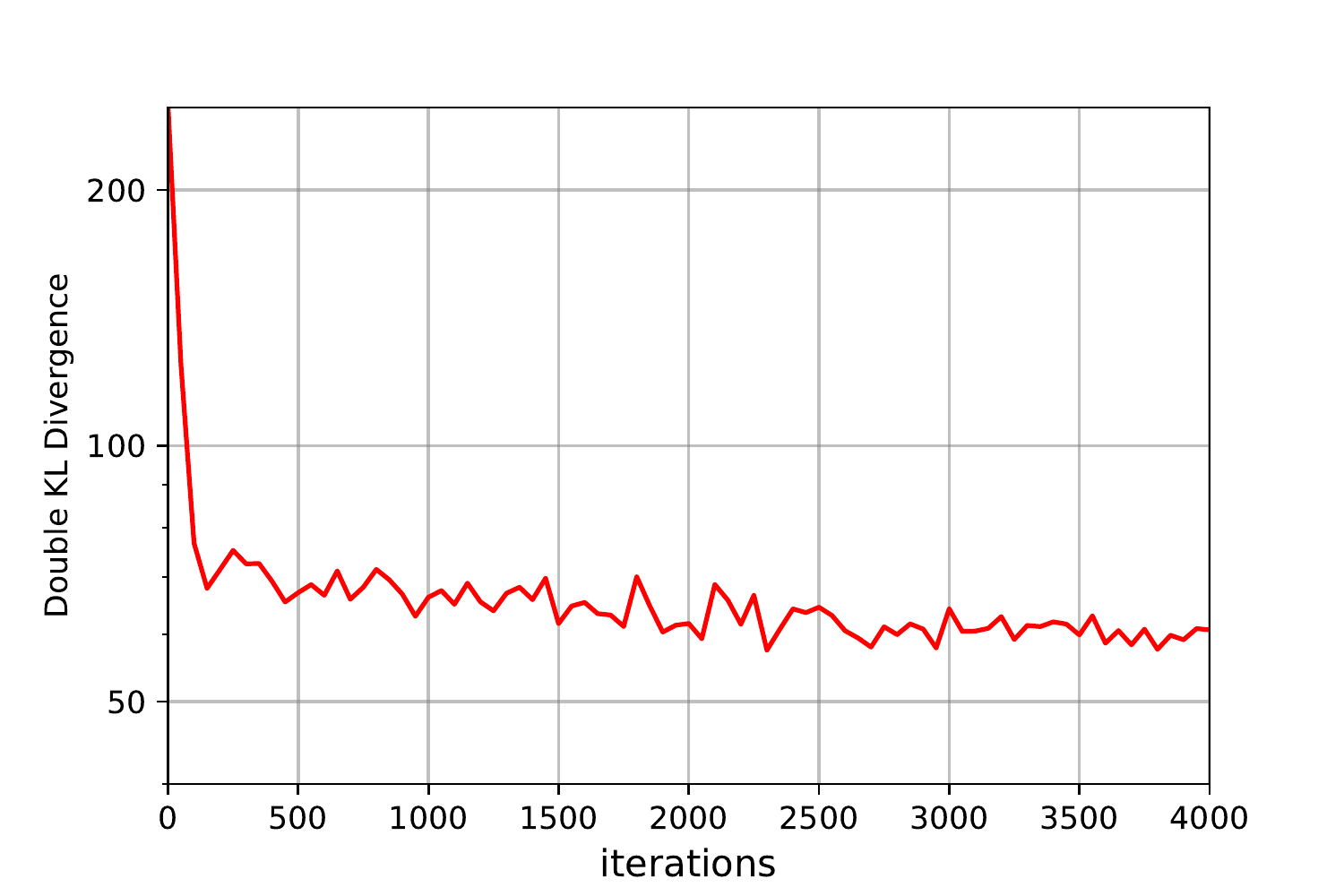}\includegraphics[width=0.56\textwidth]{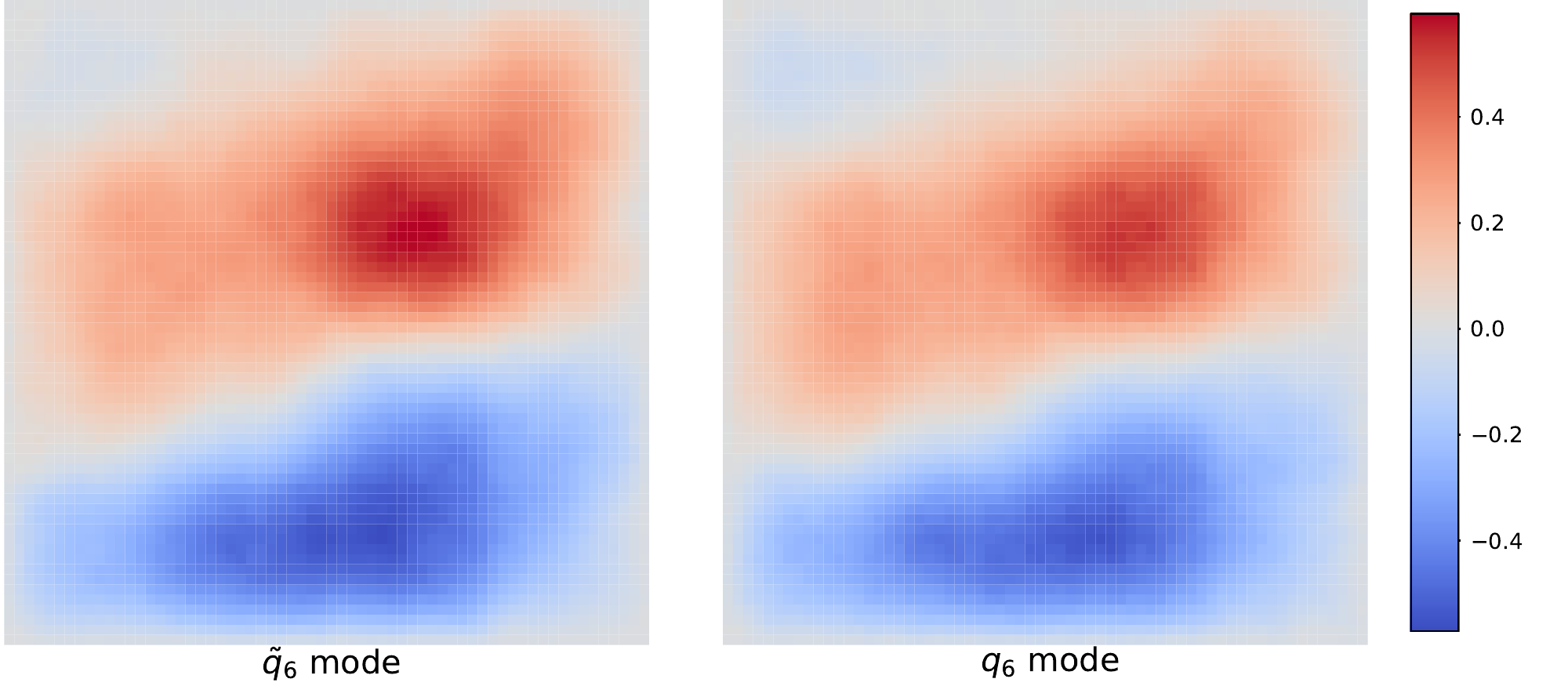}\\
    \caption{Necessity of training after prior conditioning. Left: learning curve of multi-stage MsIGN at $l=6$ in the synthetic BIP example; Middle and Right: comparison of the modes captured by the prior conditioned untrained model and the trained model in the elliptic BIP example. The learning curve shows that the model distribution is constantly getting closer to the target distribution in the last stage of training, supporting the necessity of training after prior conditioning. The mode comparison shows that bumps and caves in the left images are different from the right ones, especially in scale, as shown by color shade. Therefore, prior conditioning provides a good initial guess, but training is still necessary.}
    \label{fig: training_necessity}
\end{figure*}

\subsubsection{Ablation Study of Bayesian Inverse Problem}
\label{ap: exp_result_abl_BIP}

In Figure \ref{fig: ablation_BIP} we compared different variants of MsIGN and its training strategy at scale $l=6$. In Figure \ref{fig: ablation_more} we plot the same comparison at intermediate scales $l=1, \ldots, 5$. Since the curves overlap each other heavily in Figure \ref{fig: ablation_more}, we conclude their results of mode capturing (together with Figure \ref{fig: ablation_BIP}) in Table \ref{tab: ablation_result}.

\begin{table*}[tbp]
\caption{Table for mode capturing results by eye ball norm. Upper: synthetic Bayesian inverse problem; Lower: elliptic Bayesian inverse problem. ``T'' demotes the successful capturing of two modes, ``F'' denotes mode collapse, while ``I'' denotes biased, not well-separated modes capturing. For results marked with ``I'', we refer readers to Figure \ref{fig: ablation_more} for detail information. $^*$: we initialize the $l=2$ model by our MsIGN $l=1$ pretrained model, see Appendix \ref{ap: exp_setting_BIP}.}
\label{tab: ablation_result}
\begin{center}
\begin{small}
\begin{sc}
\begin{tabular}{l|cccccc}
\toprule
Scale & $l=1$ & $l=2$ & $l=3$ & $l=4$ & $l=5$ & $l=6$ \\
\midrule
Glow & T & F & F & F & F & F \\
MsIGN-SNN & T & T & T & T & I & I \\
MsIGN-KL-S & T & F & F & F & I & F \\
MsIGN-KL$^*$ & T & T & T & T & T & T \\
MsIGN-AS-S & T & F & F & F & F & F \\
MsIGN-AS$^*$ & T & T & T & I & I & I \\
\textbf{MsIGN} & \textbf{T} & \textbf{T} & \textbf{T} & \textbf{T} & \textbf{T} & \textbf{T} \\
\bottomrule
\end{tabular}
\quad
\begin{tabular}{l|cccccc}
\toprule
Scale & $l=1$ & $l=2$ & $l=3$ & $l=4$ & $l=5$ & $l=6$ \\
\midrule
Glow & F & F & F & F & F & F \\
MsIGN-SNN & F & F & F & F & F & F \\
MsIGN-KL-S & F & F & F & F & F & F \\
MsIGN-KL$^*$ & F & I & I & I & I & I \\
MsIGN-AS-S & F & F & F & F & F & F \\
MsIGN-AS$^*$ & F & I & I & T & I & I \\
\textbf{MsIGN} & \textbf{T} & \textbf{T} & \textbf{T} & \textbf{T} & \textbf{T} & \textbf{T} \\
\bottomrule
\end{tabular}
\end{sc}
\end{small}
\end{center}
\end{table*}

We can see from Table \ref{tab: ablation_result} that our framework and strategy outperforms all its variants in these two Bayesian inverse problems, which proved the necessity of our prior conditioning layer, network design, multi-stage training strategy, and Jeffreys divergence. In particular, the experiment of MsIGN-SNN supports our prior conditioning layer design, the experiment of MsIGN-KL supports our use of the Jeffreys divergence and MsIGN-KL-S supports our use of multi-stage training strategy.

Besides that, we can also see that multi-stage training also benefits other models like MsIGN with KL divergence objective or A-SVGD with MsIGN. By carefully comparing the marginals plotted in Figure \ref{fig: ablation_more}, we can also conclude that Jeffreys divergence can help capture more balanced modes than KL divergence.

\begin{figure*}[tbp]
    \centering
    \includegraphics[width=0.36\textwidth]{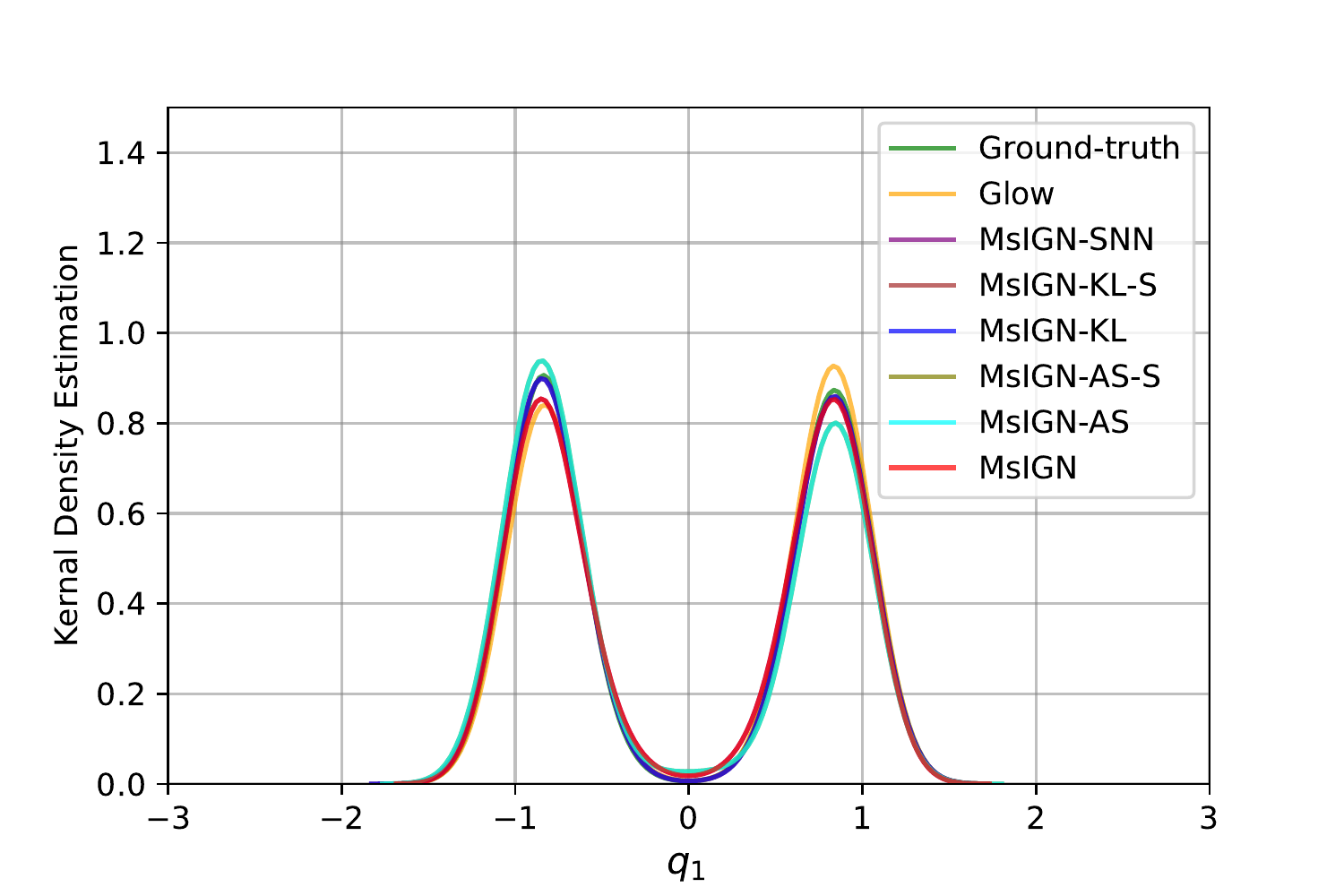}\includegraphics[width=0.36\textwidth]{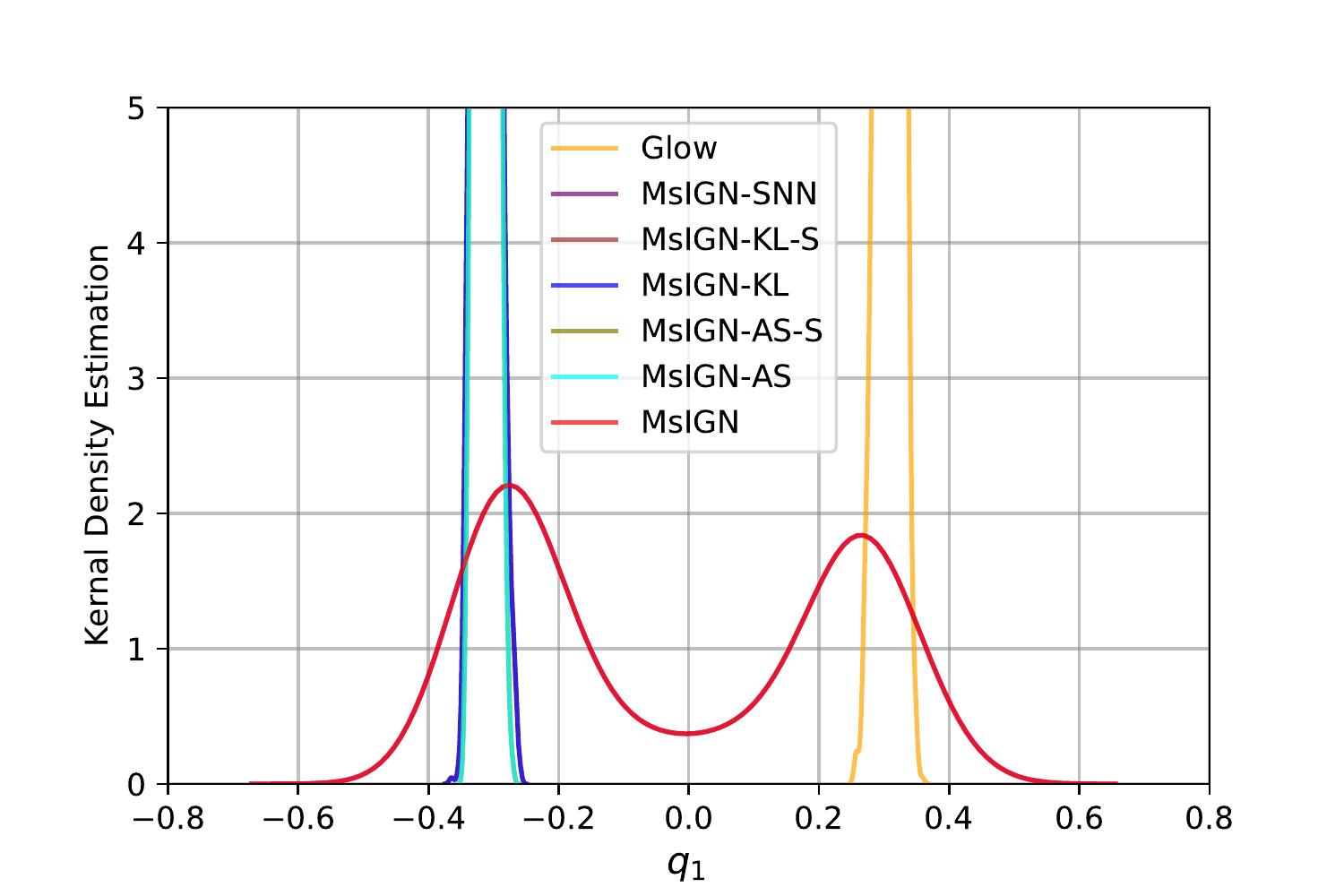}\\
    \includegraphics[width=0.36\textwidth]{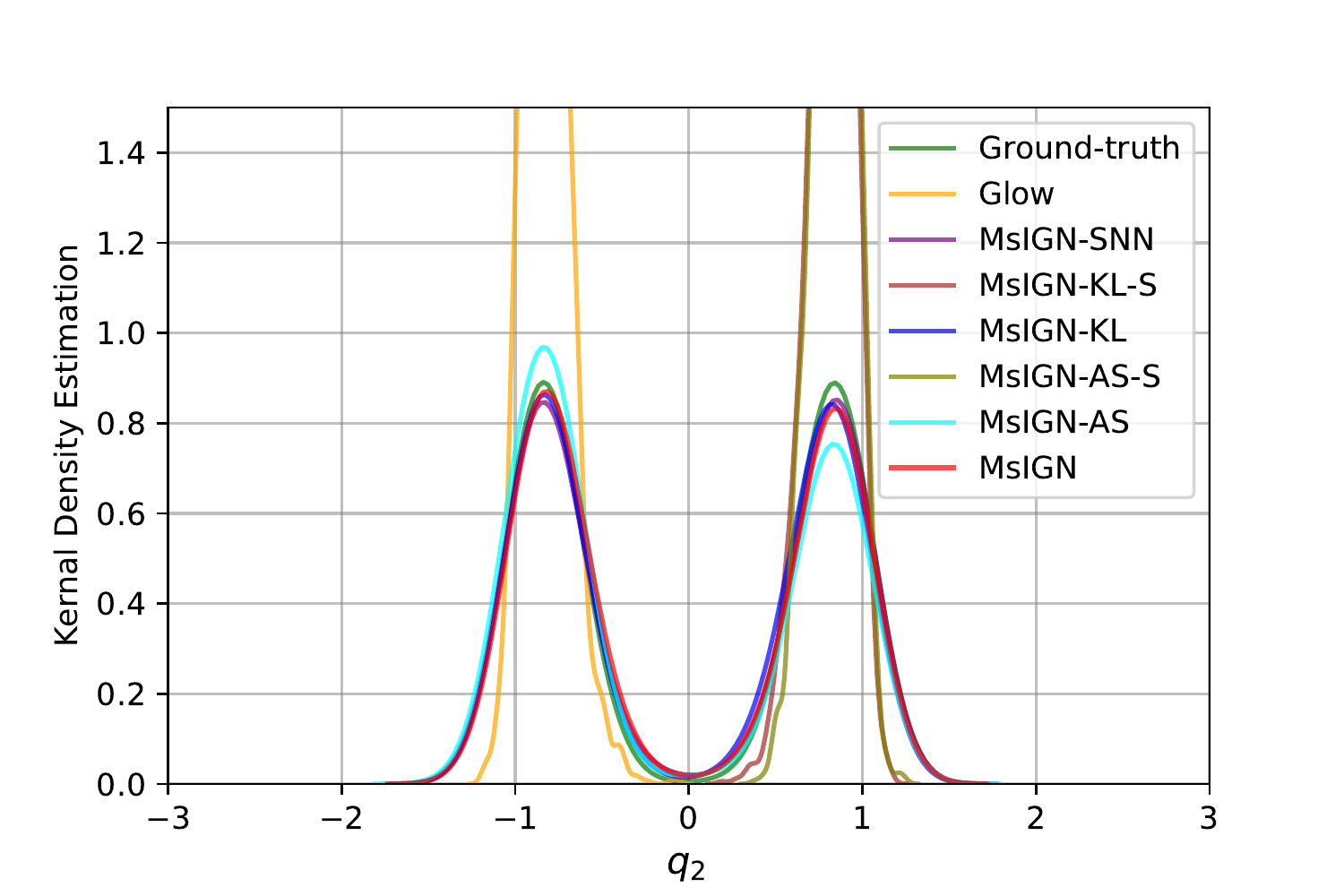}\includegraphics[width=0.36\textwidth]{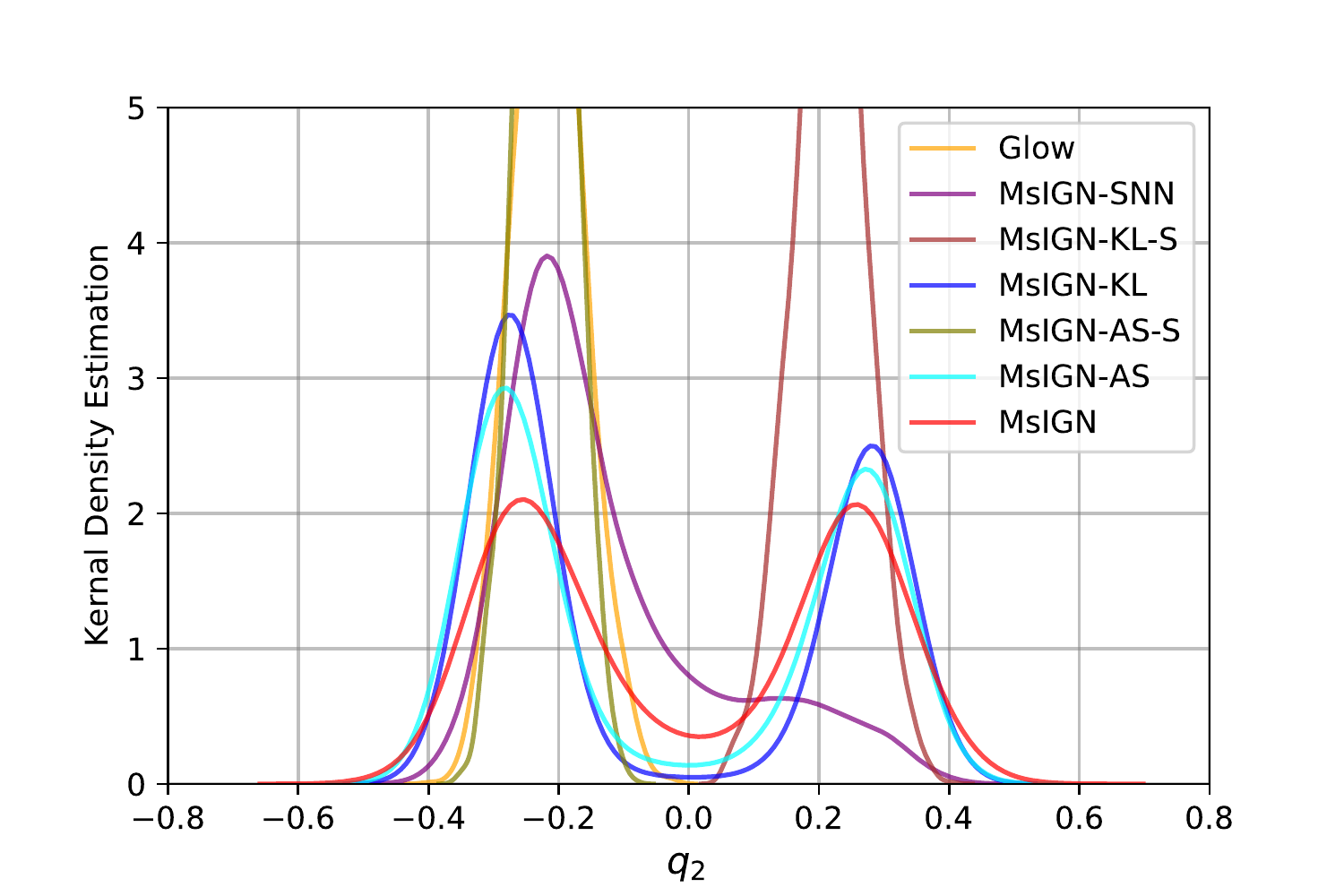}\\
    \includegraphics[width=0.36\textwidth]{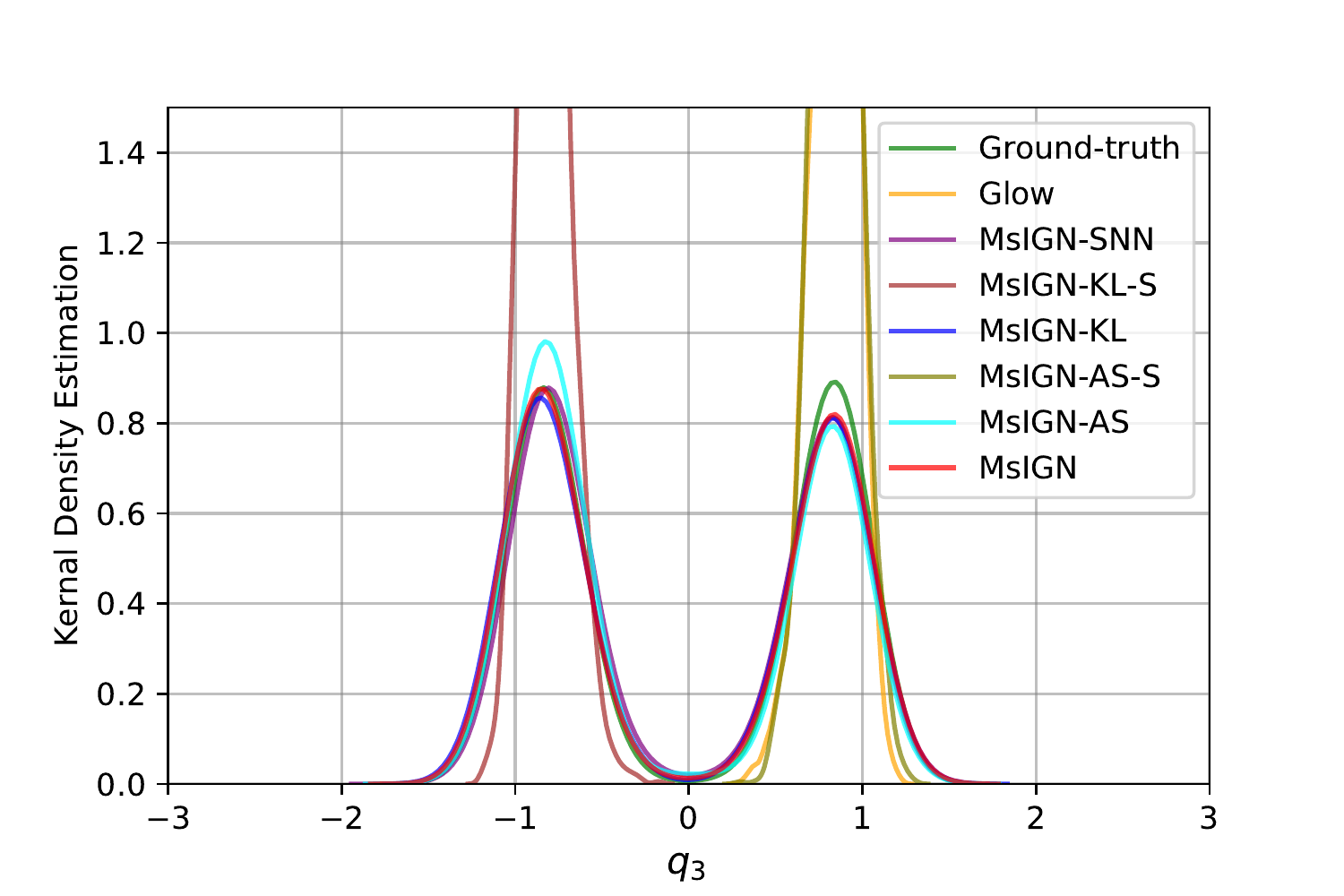}\includegraphics[width=0.36\textwidth]{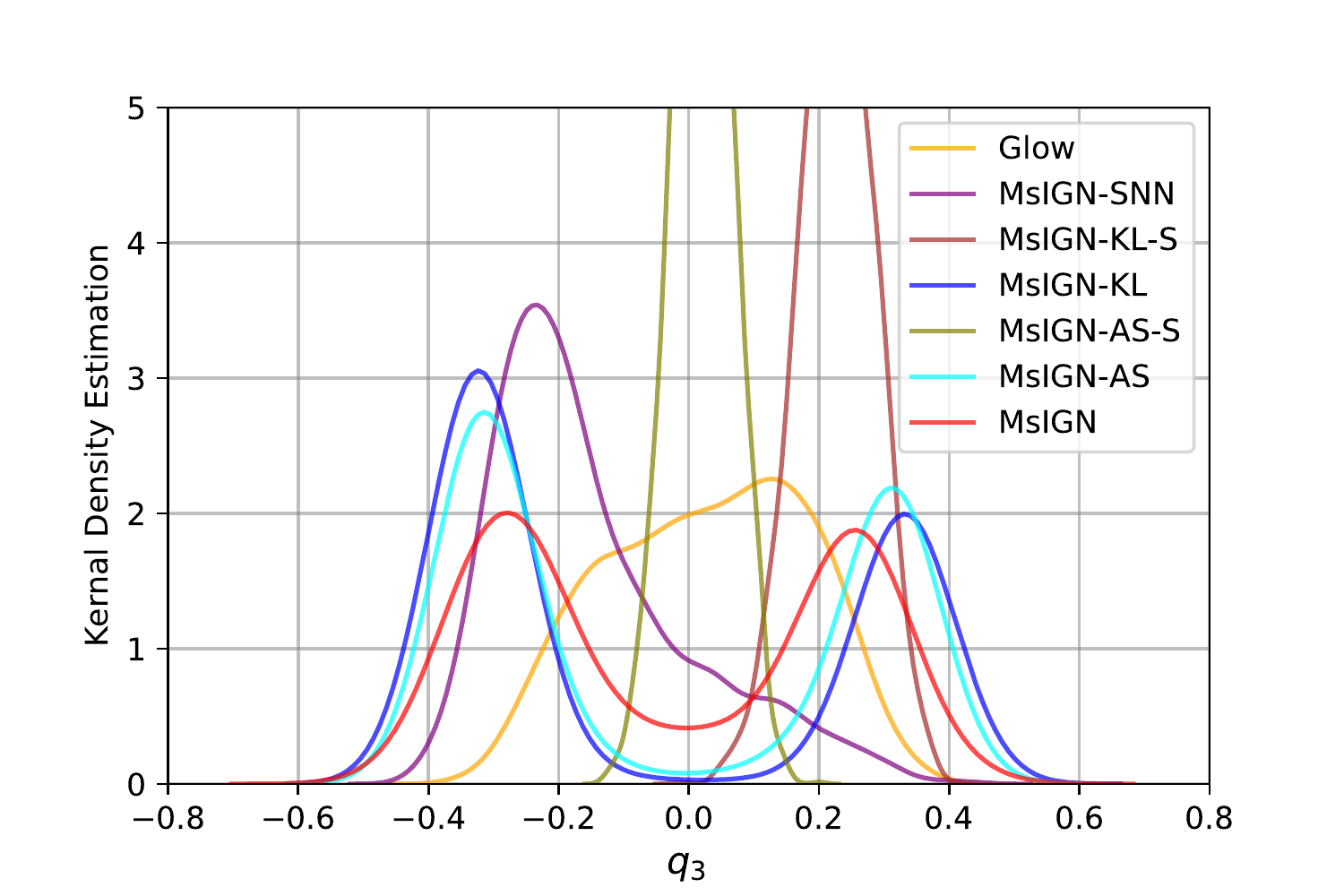}\\
    \includegraphics[width=0.36\textwidth]{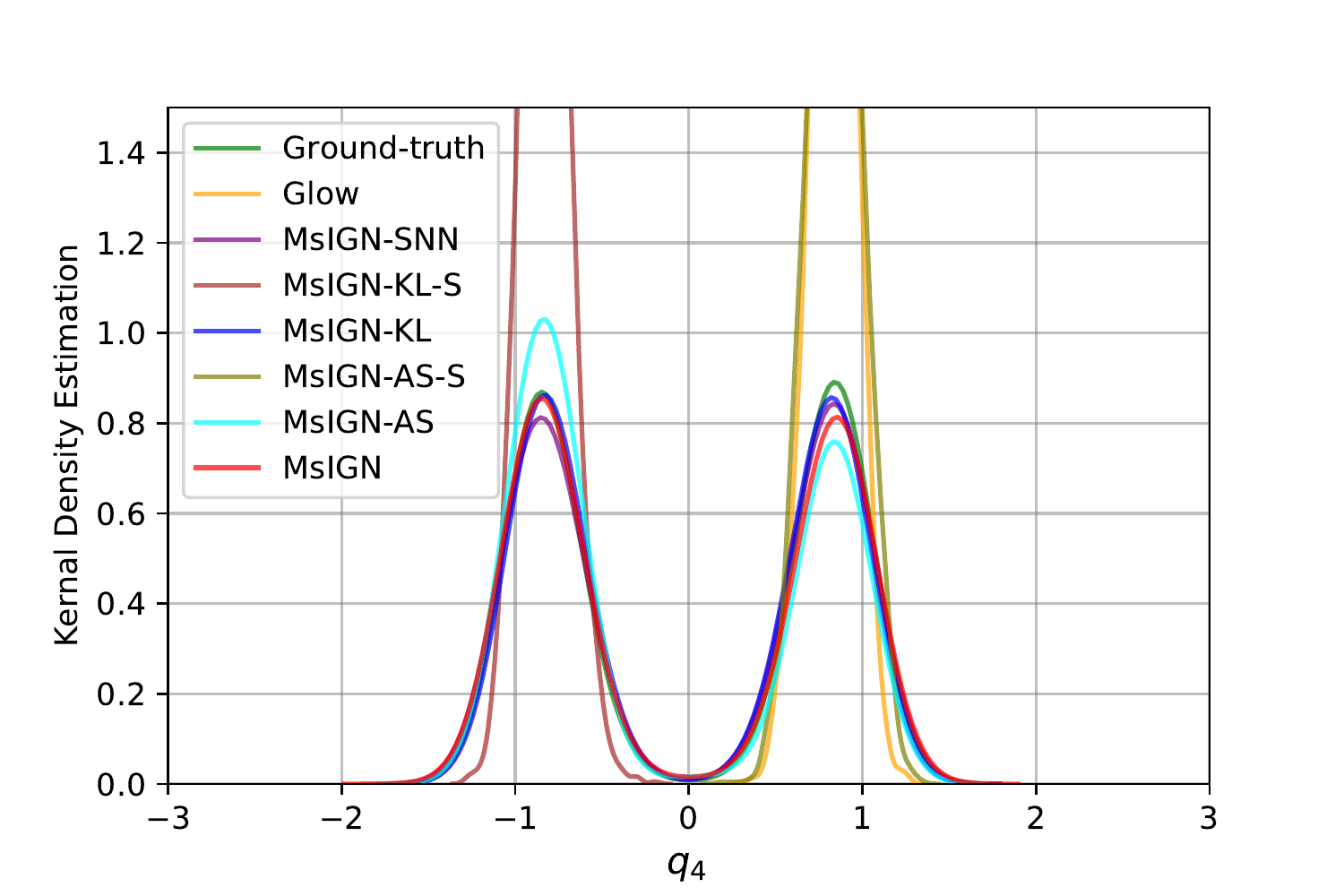}\includegraphics[width=0.36\textwidth]{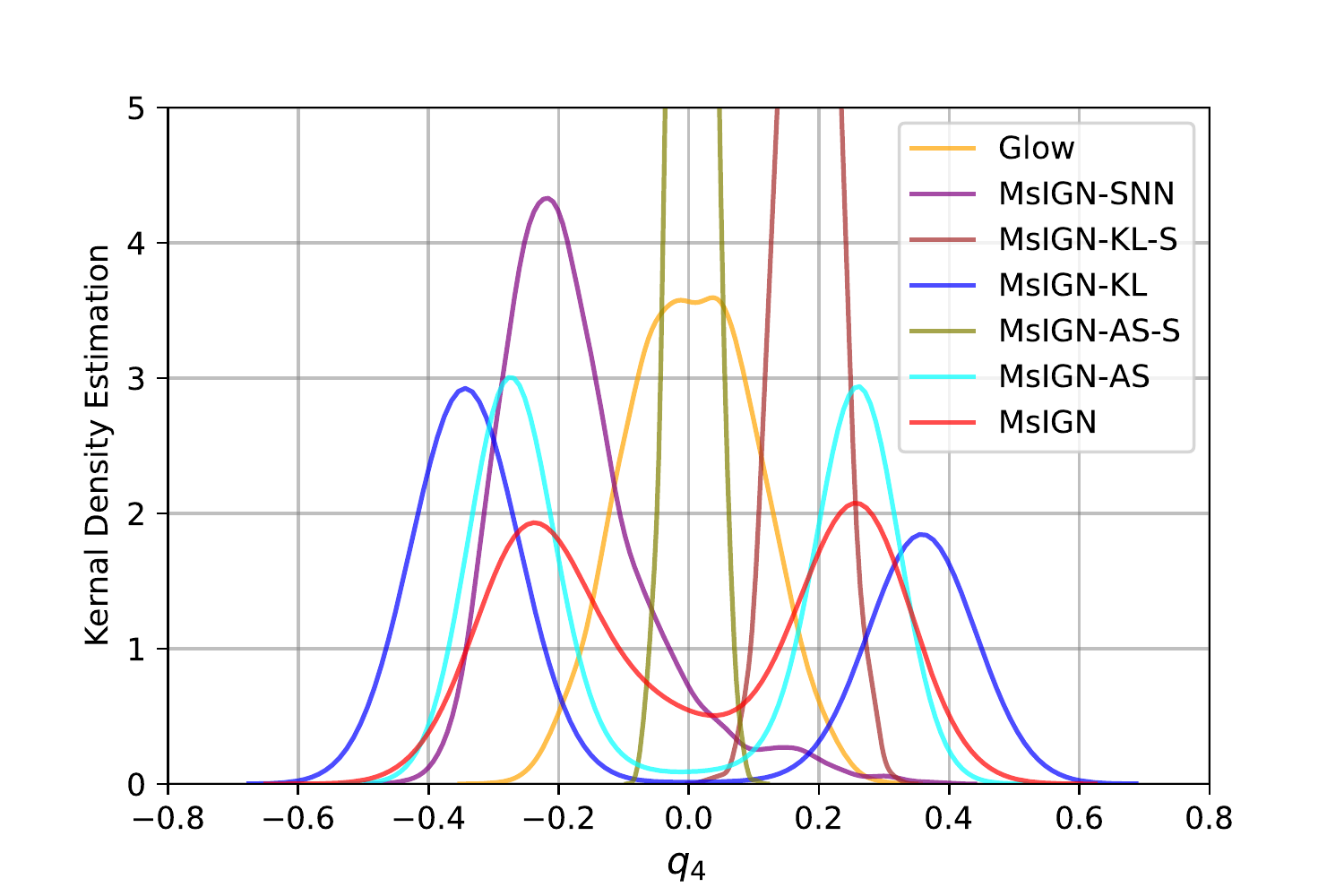}\\
    \includegraphics[width=0.36\textwidth]{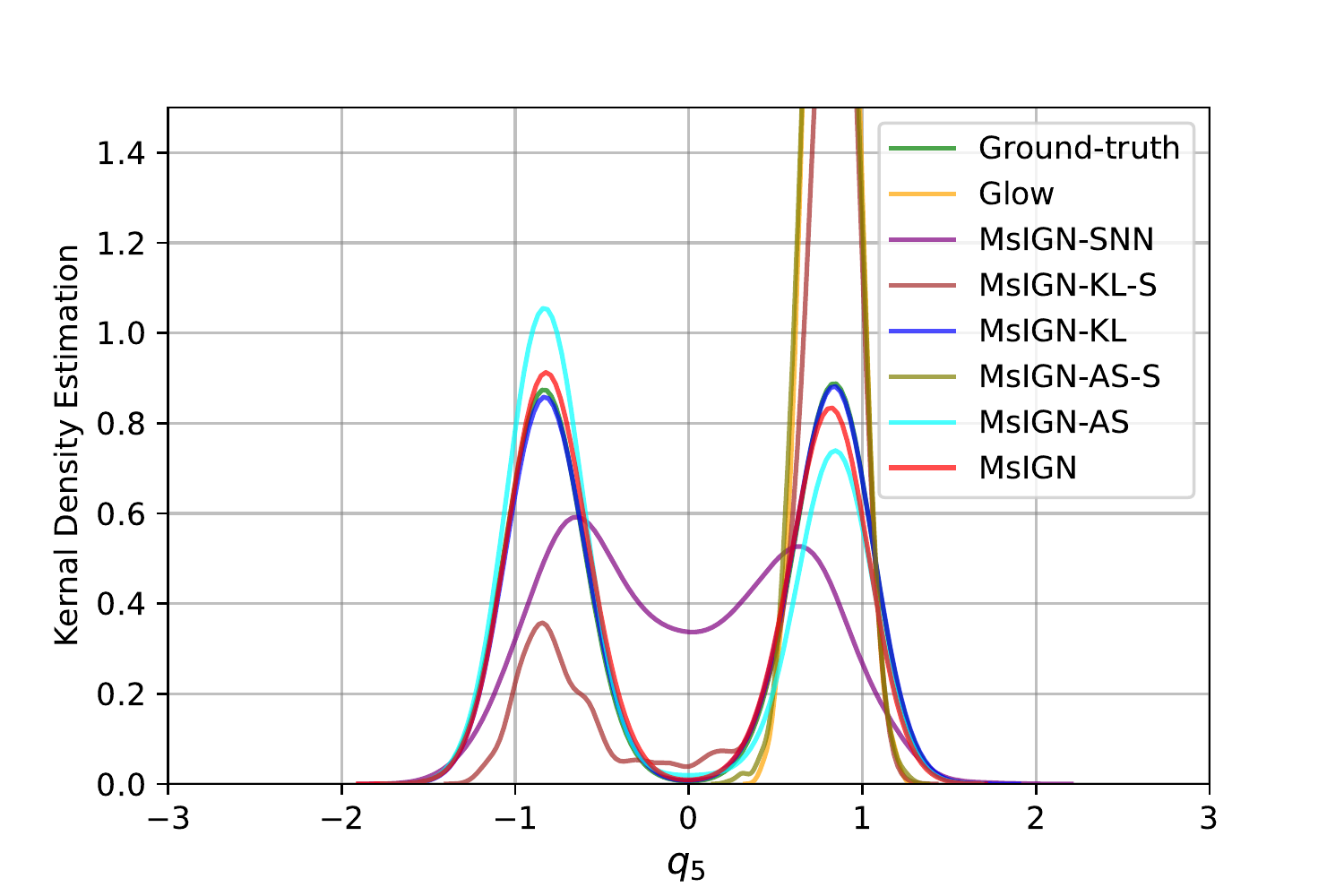}\includegraphics[width=0.36\textwidth]{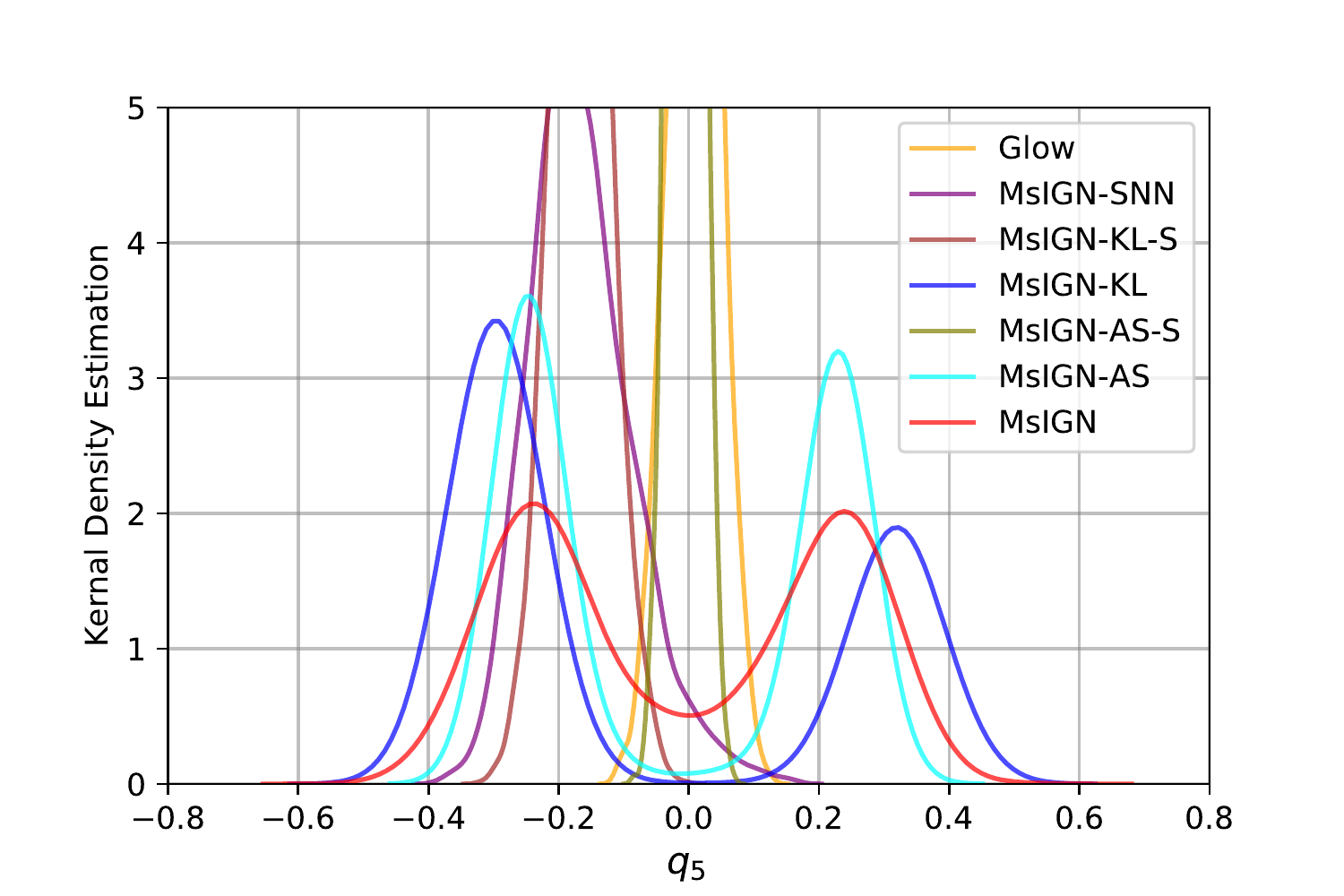}\\
    \caption{Ablation study at intermediate scales $l=1, \ldots, 5$. Left: Synthetic Bayesian inverse problem; Right: Elliptic Bayesian inverse problem. For MsIGN-AS and MsIGN-KL, we initialize their $l=2$ models by our MsIGN $l=1$ pretrained model, see Appendix \ref{ap: exp_setting_BIP}.}
    \label{fig: ablation_more}
\end{figure*}

\section{Experimental Setting and Additional Results for Image Synthesis in Section \ref{sec: exp_Image}}
\label{ap: exp_Image} 

\subsection{Experimental Setting of Image Synthesis}
\label{ap: exp_setting_Image}

Although there is no posterior for natural images, we can still use MsIGN to capture the distribution of natural images. We still feed Gaussian noises to MsIGN, and hope to get high-quality images from it as in (\ref{eq: MsIGN}). The training of MsIGN is now governed by the Maximal Likelihood Estimation due to the lack of the posterior density. In other words, we train our MsIGN by maximizing $\E_{x\sim q}[\log p_\theta(x)]$, which is equivalent to minimizing $\KL(q\Vert p_\theta)$, where $q$ is the empirical distribution of natural images given by the data set. As for the multiscale strategy, we naturally take $q_l$ to be the distribution of (downsampled) images at resolution $d_l$.

We use the invertible block introduced in \citep{kingma2018glow} as our model for the invertible flow. For our numbers in Table \ref{tab: bits-per-dimension}, we report our hyperparameter settings in Table \ref{tab: image_param}. Samples from those data sets are treated as $8$-bit images. For all experiments we use Adam \citep{kingma2014adam} optimizer with $\alpha=0.001$ and default choice of $\beta_1$, $\beta_2$. For models here that requires mutli-stage training in Algorithm \ref{alg: training}, non-final stages ($l<L$) will only be trained for $125$ epochs.

\begin{table*}[tbp]
\caption{Hyperparameter setting for results in Table \ref{tab: bits-per-dimension}. Here the meaning of terms can be found in \citep{kingma2018glow}.}
\label{tab: image_param}
\begin{center}
\begin{small}
\begin{sc}
\begin{tabular}{l|ccccc}
\toprule
Data Set & MNIST & CIFAR-10 & CelebA & ImageNet 32 & ImageNet 64 \\
\midrule
Minibatch Size & 400 & 400 & 200 & 400 & 200 \\
Scales (L) & 2 & 3 & 3 & 3 & 3 \\
$\sharp$ of Glow Blocks (K) & 32 & 32 & 32 & 32 & 32 \\
$\sharp$ of Hidden Channels & 512 & 512 & 512 & 512 & 512 \\
$\sharp$ of Epochs & 2000 & 2000 & 1000 & 400 & 200 \\
\bottomrule
\end{tabular}
\end{sc}
\end{small}
\end{center}
\end{table*}

To establish the prior conditioning layer $\PC$ in this image application, we let the downsample operator $\A_l$ from scale $l$ to scale $l-1$ be the average pooling operator with kernel size 2 and stride 2. We further assume the covariance $\Sigma_l$ at each scale be a scalar matrix, i.e. a diagonal matrix with equal diagonal elements.

\begin{figure}[H]
    \vspace{-1mm}
    \centering
    \includegraphics[width=0.24\columnwidth]{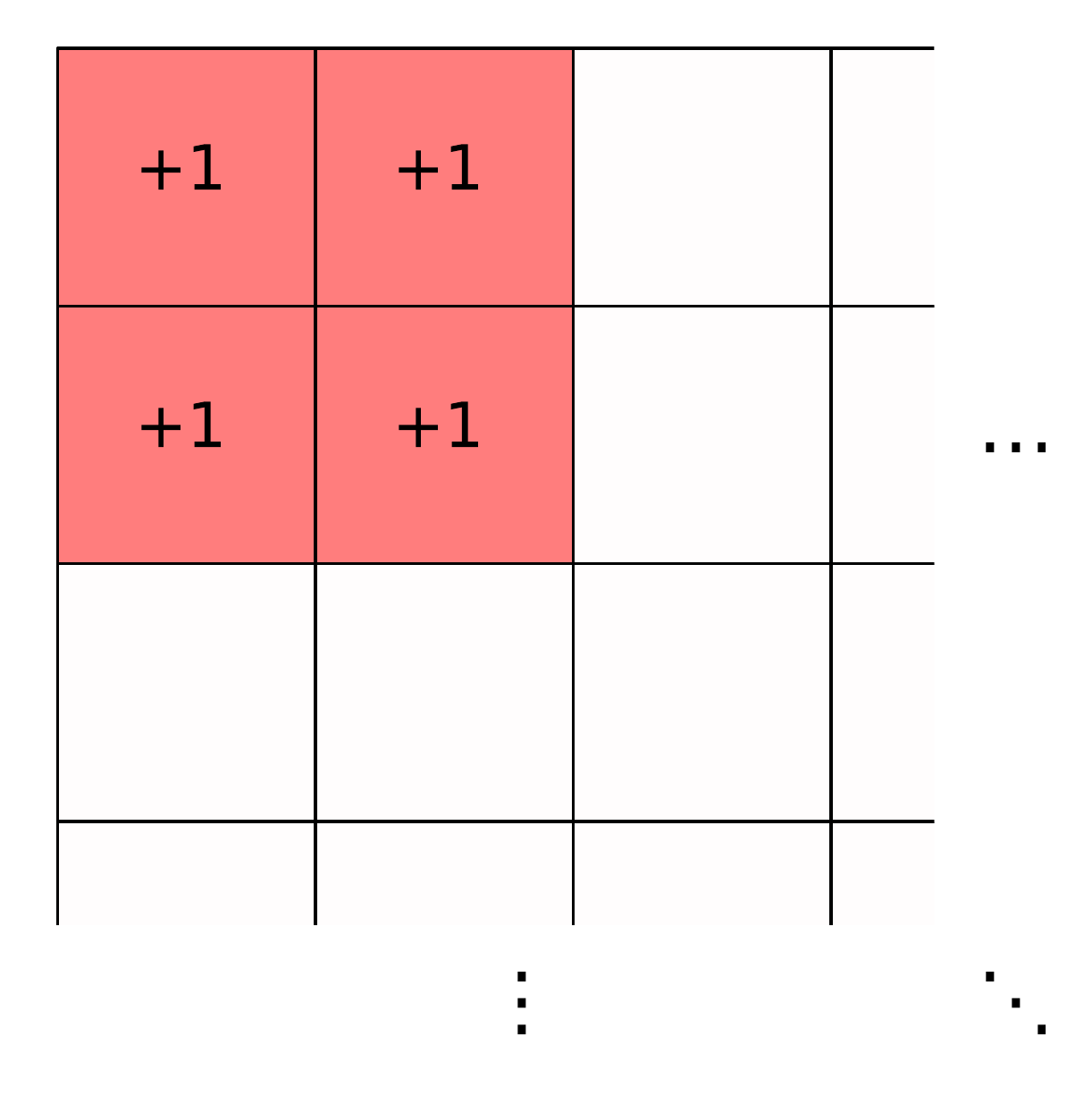}
    \includegraphics[width=0.24\columnwidth]{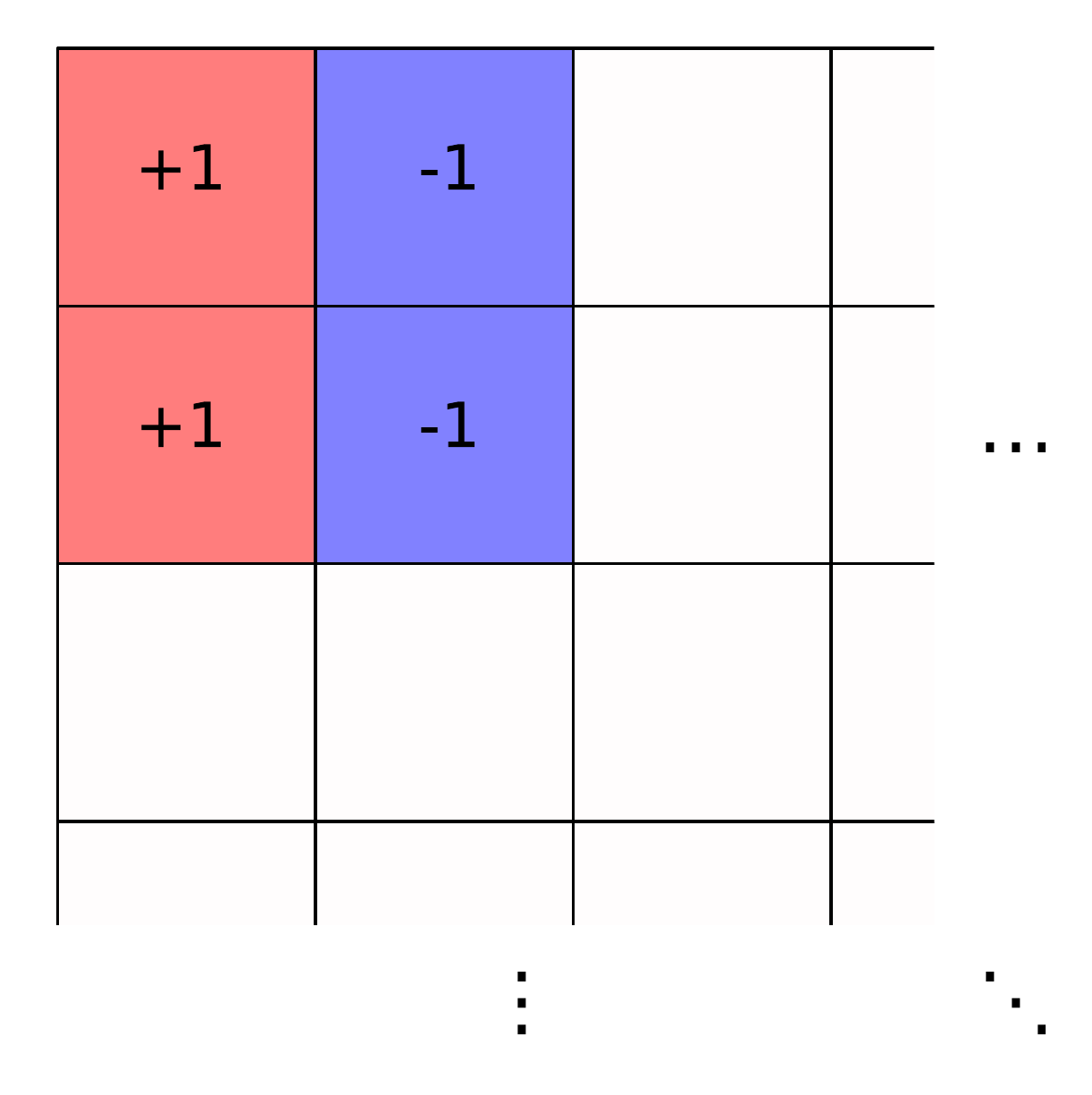}
    \includegraphics[width=0.24\columnwidth]{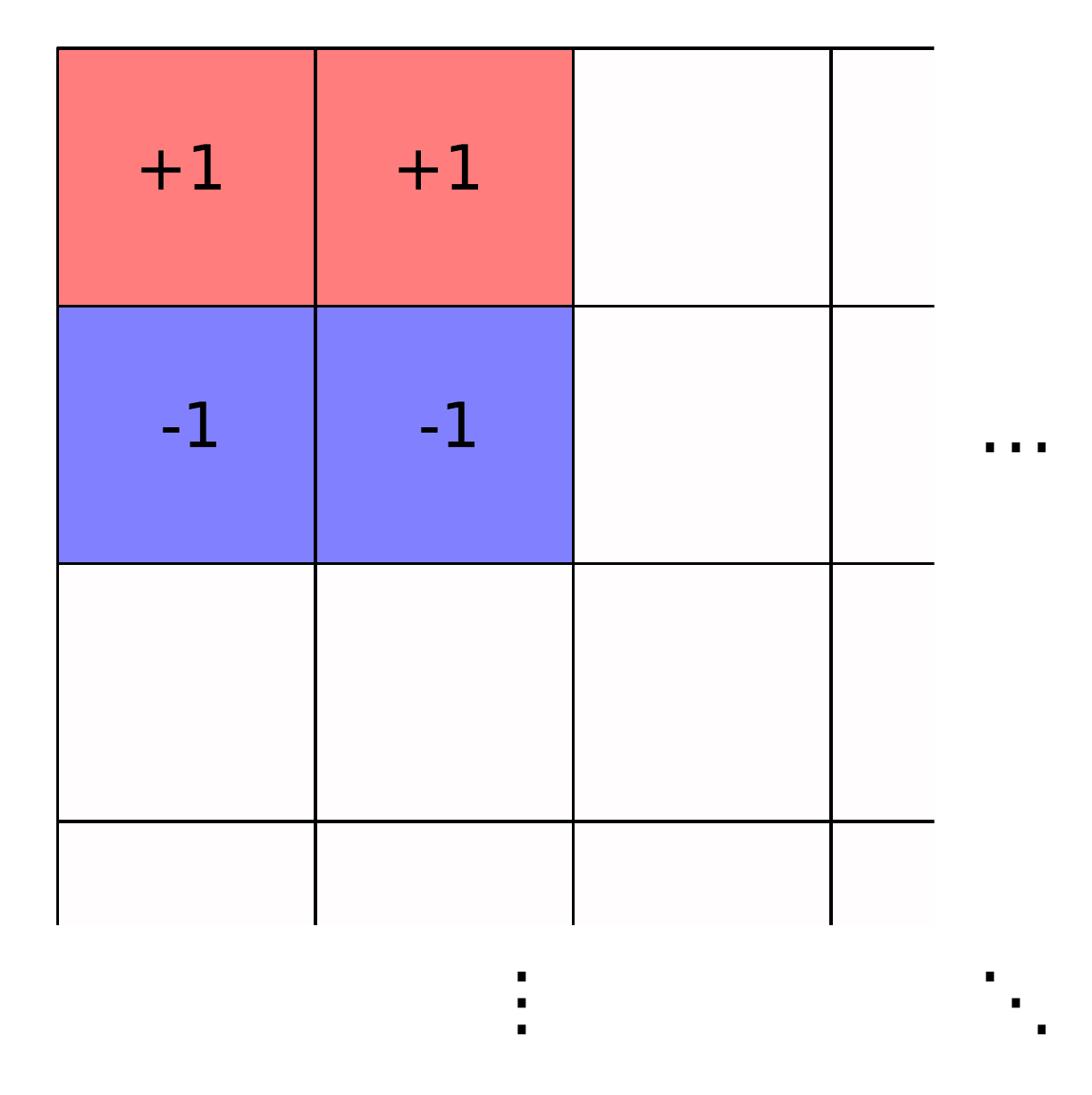}
    \includegraphics[width=0.24\columnwidth]{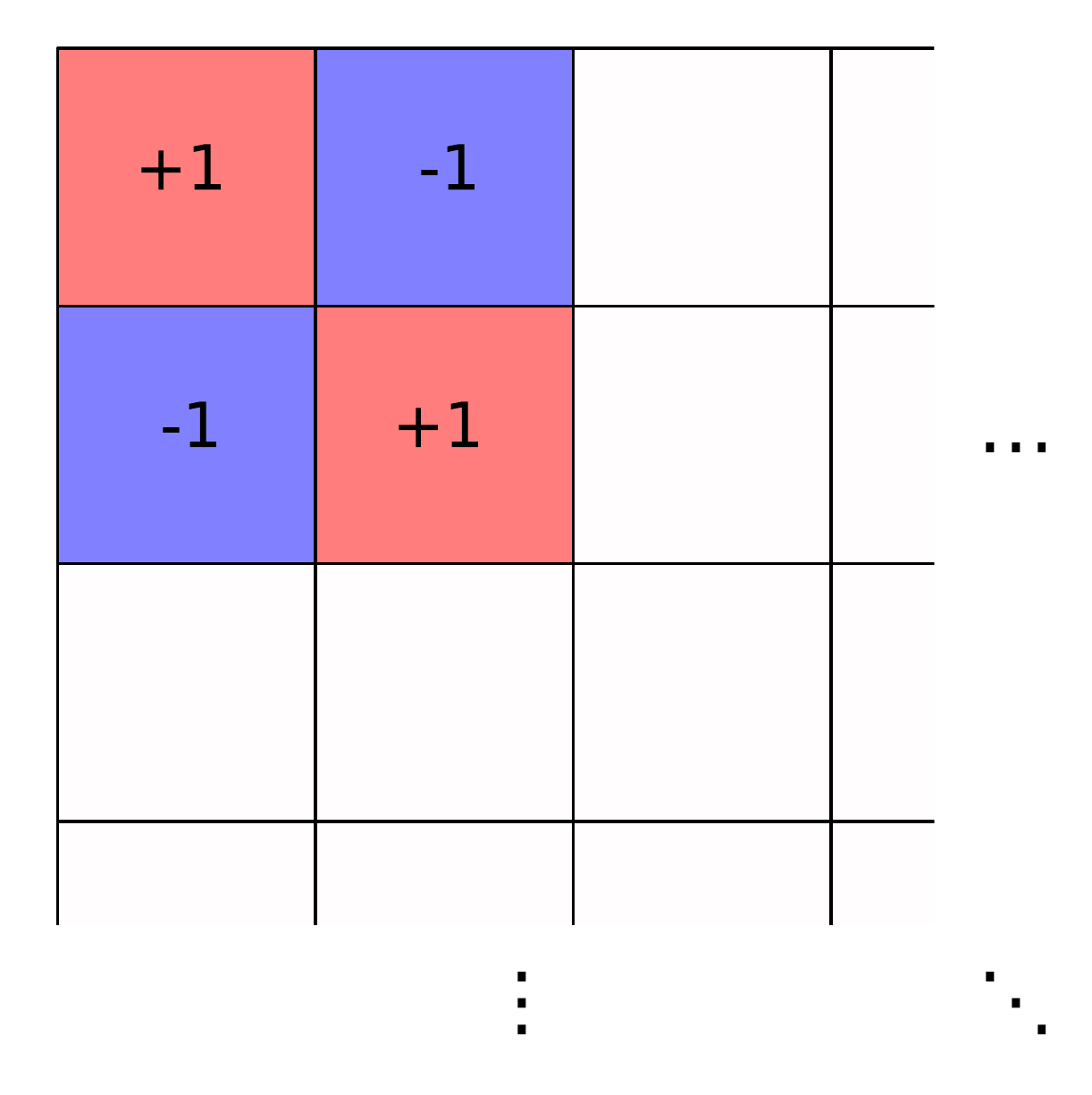}
    \vspace{-2mm}
    \caption{Left most: an example row of $\A_l$, plotted as a matrix; The rest: example rows of $\tilde{\A}_l$ correspond to the former row of $\A_l$. They (with some unplotted ones) form the Haar basis, and can be expressed as local convolution operation.}
    \label{fig: haar}
    \vspace{-4mm}
\end{figure}

Since $\A_l\in\R^{d_{l-1}\times d_l}$ is the average pooling operator, its rows, which give averages of each local patch, is a subset of the Haar basis, see Figure \ref{fig: haar}. We can collect the rest Haar basis as $\tilde{\A}_l\in\R^{(d_l-d_{l-1})\times d_l}$. Due to the orthogonality of the Haar basis, there exists a constant $\lambda_l>0$ such that
\begin{align*}
    \begin{bmatrix}
    \A_l\A_l^T &  \\
    & \tilde{\A}_l\tilde{\A}_l^T
    \end{bmatrix}&=
    \begin{bmatrix}
    \A_l \\
    \tilde{\A}_l
    \end{bmatrix}\begin{bmatrix}
    \A_l \\
    \tilde{\A}_l
    \end{bmatrix}^T=\lambda_l I_{d_l}\\
    &=\begin{bmatrix}
    \A_l \\
    \tilde{\A}_l
    \end{bmatrix}^T\begin{bmatrix}
    \A_l \\
    \tilde{\A}_l
    \end{bmatrix}=\A_l^T\A_l+\tilde{\A}_l^T\tilde{\A}_l\,.
\end{align*}
As a by-product we see $\A_l\A_l^T=\lambda_l I_{d_{l-1}}$ and $\tilde{\A}_l\tilde{\A}_l^T=\lambda_l I_{d_l-d_{l-1}}$. In our case, as $\A_l$ is the average pooling operator, we actually have $\lambda_l=1/4$.

Since we assume the covariance $\Sigma_l$ is a scalar matrix, we can find a scalar $c_l>0$ such that $\Sigma_l=c_lI_{d_l}$. Now following Theorem \ref{thm: gaussian_condition}, we can find an explicit form for $\Sigma_{l|l-1}$, $l\geq 2$, which is the $\Sigma^c$ at scale $l$:
\begin{align*}
    \Sigma_{l|l-1}&=\Sigma_l-\Sigma_l\A_l^T(\A_l\Sigma_l\A_l^T)^{-1}\A_l\Sigma_l  \\
    &=c_lI_{d_l} - c_l \A_l^T(\lambda_lI_{d_{l-1}})^{-1}\A_l\\
    &=\frac{c_l}{\lambda_l}(\A_l^T\A_l+\tilde{\A}_l^T\tilde{\A}_l) - \frac{c_l}{\lambda_l} \A_l^T\A_l\\
    &=\frac{c_l}{\lambda_l}\tilde{\A}_l^T\tilde{\A}_l\,.
\end{align*}
Therefore, we obtain the decomposition of $\Sigma_{l|l-1}=W_lW_l^T$ in Theorem \ref{thm: gaussian_condition} for free, where now $W_l$ is the original $W$ at scale $l$. One apparent choice is $W_l=\mu_l\tilde{\A}_l^T$ with $\mu_l=\sqrt{\frac{c_l}{\lambda_l}}$. Finally, as suggested by Theorem \ref{thm: gaussian_condition} we are now only left to estimate the scalar $\mu_l$ for each $l\geq 2$ to establish $\PC_l$.

The constant $\mu_l$ is estimated numerically on data sets. In fact, we have accessible to different resolutions of images from the data set when we perform pooling operation. We take $x_l$ to be the pooling of images from data set to its resolution, and estimate $\mu_l$ according to Theorem \ref{thm: gaussian_condition}: $$x_l=U_{l-1}x_{l-1}+W_lz_l=U_{l-1}x_{l-1}+\mu_l\tilde{\A}_l^Tz_l\,,$$
where $z_l\sim\N(0, I_{d_l-d_{l-1}})$ are the random noise at scale $l$, and $U_{l-1}$ by definition is
\begin{align*}
    U_{l-1}&=\Sigma_l\A_l^T(\A_l\Sigma_l\A_l^T)^{-1}\\
    &=c_l\A_l^T(c_l\A_l\A_l^T)^{-1}=\A_l^T(\A_l\A_l^T)^{-1}\\
    &=\A_l^T(\lambda_lI_{d_{l-1}})^{-1}=\frac{1}{\lambda_l}\A_l^T\,.
\end{align*}
Plug it back, we have
$$x_l=\frac{1}{\lambda_l}\A_l^Tx_{l-1}+\mu_l\tilde{A}_l^Tz_l\,.$$
Now multiply both sides with $\tilde{\A}_l$, noticing that $\tilde{\A}_l\tilde{A}_l^T=\lambda_lI_{d_l-d_{l-1}}$ and $\tilde{\A}_l\A_l^T=0$, we arrive at
$$\tilde{\A}_lx_l=\lambda_l\mu_lz_l\,.$$

Since $\mu_l$ is a scalar, it can be estimated by moment matching of both sides, as $\lambda_l$ and $\tilde{\A}_l$ is known. Here $x_l$ is the natural images at resolution $d_l$. For example, we use $10000$ randomly sampled images from each data set and estimate $\mu_l$ by matching the variance of both sides, we report our estimates of $\mu_l$ in Table \ref{tab: mu_estimate}.

\begin{table}[tbp]
\caption{Estimate of $\mu_l$ for different data sets and scale $l$.}
\label{tab: mu_estimate}
\begin{center}
\begin{small}
\begin{sc}
\begin{tabular}{l|ccc}
\toprule
Data Set & $\mu_2$ & $\mu_3$ \\
\midrule
MNIST & 0.67 & -- \\
CIFAR-10 & 0.48 & 0.46 \\
CelebA 64 & 0.22 & 0.30 \\
ImageNet 32 & 0.32 & 0.42 \\
ImageNet 64 & 0.28 & 0.36 \\
\bottomrule
\end{tabular}
\end{sc}
\end{small}
\end{center}
\end{table}

\begin{figure*}[bp]
    \centering
    \includegraphics[width=0.6\textwidth]{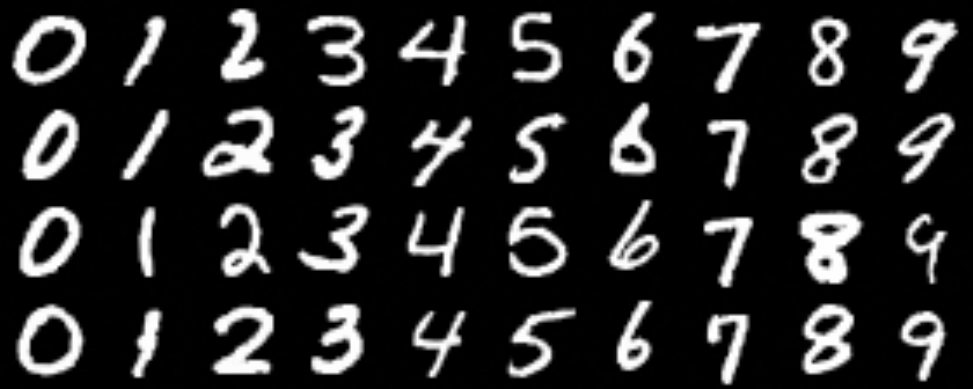}
    \caption{Synthesized $28\times 28$-resolution images from MsIGN on the MNIST data set, temperature $=1.0$. We show $4$ samples per digit.}
    \label{fig: mnist_synthesized}
\end{figure*}

\subsection{Additional Results of Image Synthesis}
\label{ap: exp_result_Image}

We attach more synthesized images by MsIGN from MNIST and CIFAR-10 in Figure \ref{fig: mnist_synthesized}, \ref{fig: cifar_synthesized}. For the CelebA data set, we made use of our multiscale design and trained our MsDGN for a higher resolution $128$. In this case, the number of scales $L=4$, and we set the hyperparameters for the first $3$ scales the same as we use for the $64*64$ resolution model. For the last scale $l=4$, due to memory limitation, we set $K=32$ and hidden channels $128$. We show our synthesized $128*128$ resolution results in Figure \ref{fig: celeba128_synthesized}.

We also use this $4$-scale model to show the interpret-ability of our internal neurons in Figure \ref{fig: internal_neuron_128}. We snapshot internal neurons for $4$ times every scale, resulting a snapshot chain of length $4*4=16$ for every generated image. We can see our MsIGN generates global features at the beginning scales and starts to add more local details at higher scales.

\begin{figure*}[tbp]
    \centering
    \includegraphics[width=0.6\textwidth]{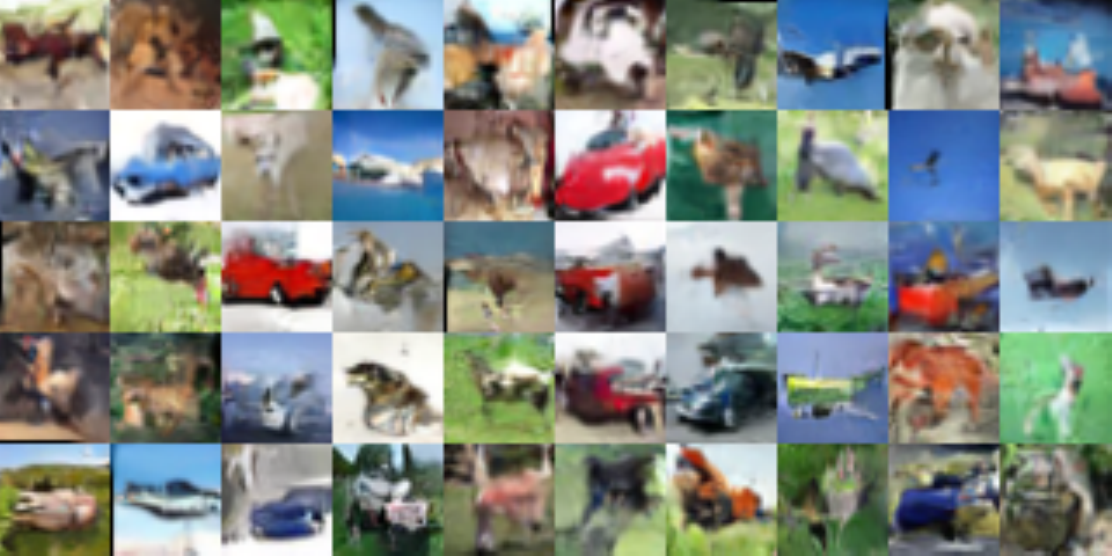}
    \caption{Synthesized images of resolution $32\times 32$ from MsIGN on the CIFAR-10 data set, temperature $=1.0$.}
    \label{fig: cifar_synthesized}
\end{figure*}

\begin{figure*}[tbp]
    \centering
    \includegraphics[width=0.9\textwidth]{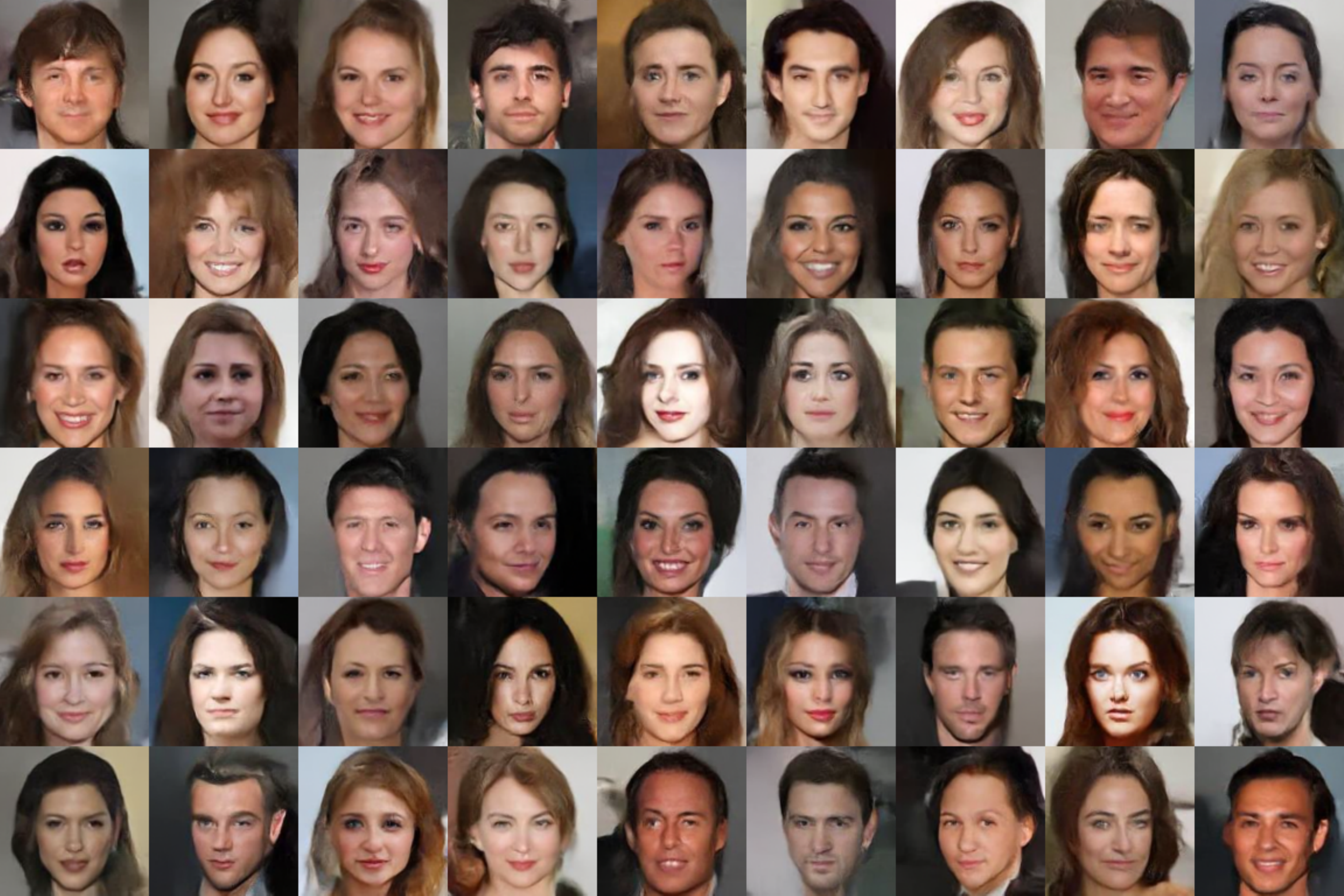}
    \caption{Synthesized images of resolution $128\times128$ from MsIGN on the CelebA data set, temperature $=0.8$.}
    \label{fig: celeba128_synthesized}
\end{figure*}

\begin{figure*}[tbp]
    \centering
    \includegraphics[width=0.78\textwidth]{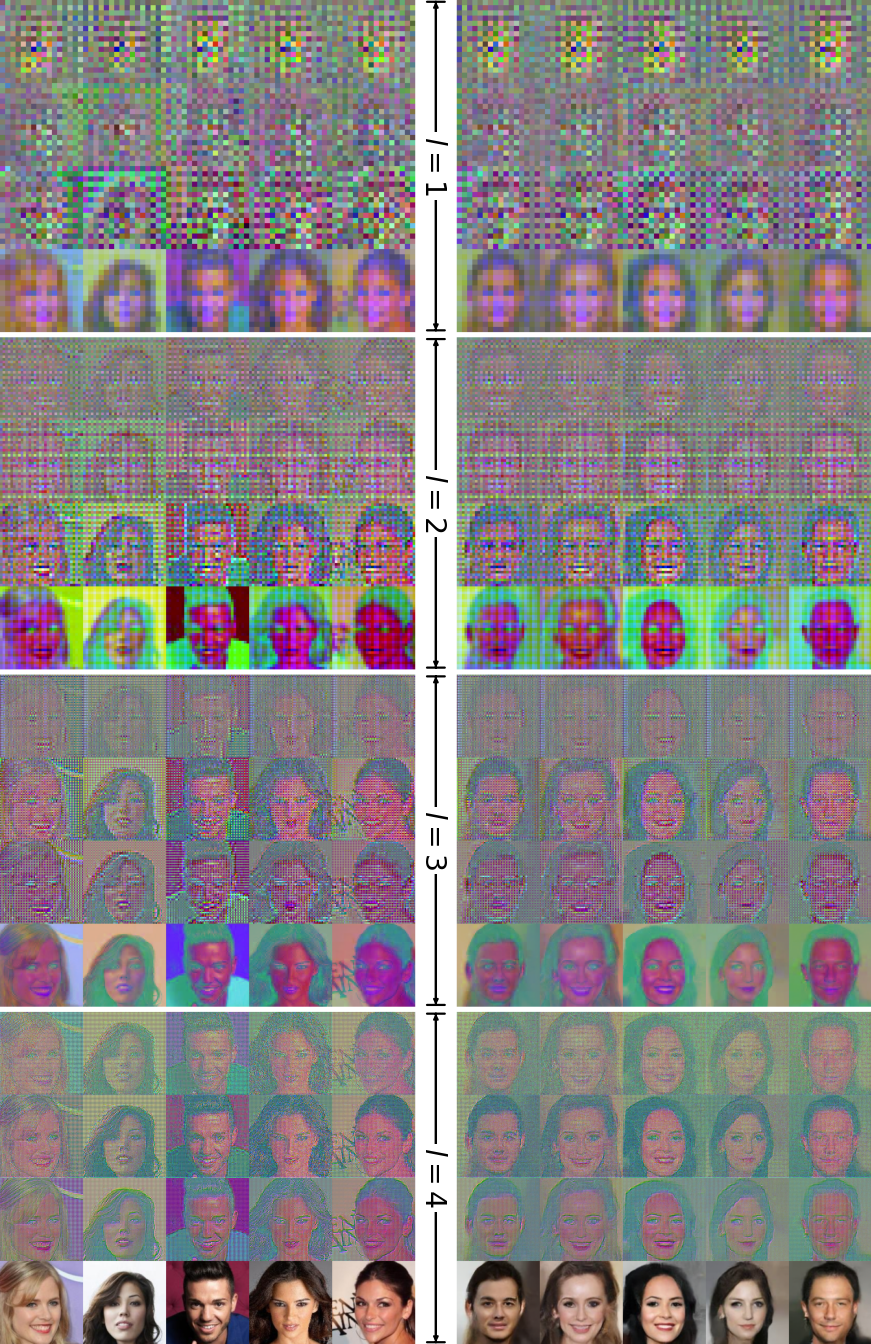}
    \caption{Visualization of internal neurons of MsIGN in synthesizing or recovering $128\times128$-resolution images on CelebA data set. Snapshots (from top to bottom) are taken $4$ times every scale, resulting $4*4=16$ checkpoints for every image generated. At scale $l$ ($1\leq l\leq 4$), where the resolution is $2^{3+l}\times2^{3+l}$, we take $4$ snapshots at the head, two trisection points and tail of the invertible flow $F_l$. Left: when recovering images from the data set; Right: when synthesizing new images from random noise.}
    \label{fig: internal_neuron_128}
\end{figure*}





\end{document}